\documentclass{article}
\usepackage{graphicx}
\usepackage{geometry}
\usepackage{latexsym}
\usepackage{amsfonts}
\usepackage{mathrsfs}
\usepackage{mathdots}
\usepackage{amssymb,amscd}
\usepackage[all]{xy}
\usepackage{amsmath}
\usepackage{theorem}
\usepackage{thmtools}
\usepackage{fancyhdr}
\usepackage{fancybox}
\usepackage{xcolor}
\usepackage{array}
\usepackage{makecell}
\usepackage{multirow}
\usepackage{multicol}
\usepackage{slashbox}
\usepackage{authblk}
\usepackage{url}

\author{Zhifeng Kong, Shuo Ding}
  \title{Generative Adversarial Networks \\with Inverse Transformation Unit}
\date{}

\begin{document}
\maketitle

\begin{abstract}
In this paper we introduce a new structure to Generative Adversarial Networks by adding an inverse transformation unit behind the generator. We present two theorems to claim the convergence of the model, and two conjectures to nonideal situations when the transformation is not bijection. A general survey on models with different transformations was done on the MNIST dataset and the Fashion-MNIST dataset, which shows the transformation does not necessarily need to be bijection. Also, with certain transformations that blurs an image, our model successfully learned to sharpen the images and recover blurred images, which was additionally verified by our measurement of sharpness.

\end{abstract}

\section{Introduction}
In recent two years generative adversarial networks (GAN) have been increasingly concerned \cite{GAN}. GAN introduce two perceptrons that behave against each other: the generator learns the probability distribution of training data, while the discriminator learns to tell the difference. The conciseness of GAN makes it possible to amend the structure in order to improve its performance, or make it able to achieve our additionally desired effects. While many works focused on the first point (see related works), this paper focuses on the second aspect. We add an inverse transformation unit behind the generator, and make it possible to generate data with the "inverse" effect of the input transformation function. Our architecture is quite useful when we want to generate samples with some additional effects which is hard to implement but the inverse is easy to achieve. This need is natural and common in certain situations. For instance, we want to generate clear images, but we only know the way to implement its inverse -- how to blur them.

In this paper, we make the following contributions:

$\bullet$
We presented a new architecture for generative adversarial networks by adding an inverse transformation unit behind the generator.

$\bullet$
We made rigorous theoretical analysis on our structure: we found the optimal discriminator for a fixed generator when the transformation is a continuous bijection. We also claimed the convergence of the algorithm in such situation.

$\bullet$
We made two conjectures for cases when the transformation is not bijection.

$\bullet$
We applied our method to MNIST dataset \cite{MNIST} and the Fashion-MNIST dataset \cite{fashion_MNIST} with different transformation functions. A general survey on various transformation functions was done; and with some special transformation functions, the model showed its ability to sharpen the images and recover blurred images.

\section{Related Works}

In recent two years a lot of works on generative adversarial networks (GAN) have appeared. They have researched various aspects of GAN, from theory to applications, and made great improvement to the original method. Many works put their emphasis on improving the performance of GAN, by introducing new loss functions \cite{bGAN,fGAN,InfoGAN}, integrating it with other deep learning architectures \cite{DCGAN,GRAN,LAPGAN,seqGAN}, or making amendments to the original GAN with strong theoretical analysis \cite{EnergyGAN,WGAN,LSGAN}. A number of works also apply GAN to practical issues and solved problems in those domains \cite{seqGAN,cross-domain}. The purpose of our paper is to survey a new architecture of GAN which makes it possible to learn the samples with certain desired effect.


Graph generation has been a popular topic for years, and people have tried different methods to generate graphs with their desired effects. Convolutional Neural Networks (CNN) and GAN are two popular methods in this domain. In \cite{transfer_CNN}, researchers successfully train a model to transfer an image's texture style to another image using CNN. In \cite{DCGAN}, an integration of CNN and GAN is made, and the model turns out to have a better performance in generating images than the original GAN. In \cite{letter_CNN,letter_GAN,letter_VAE}, three methods including CNN, GAN and Variational Auto Encoders (VAE) are used to learn the typographical style and generate images of letters with new styles.


In order to learn the graphical samples with certain desired effect, we add an inverse transformation unit $T$ to the generator, based on the intuition that the generator will learn some additional effect, such as $T^{-1}$ if $T$ is invertible, to offset the effect of $T$. This intuition also appears in \cite{realNVP} and \cite{cycleGAN}.
In \cite{realNVP}, the mapping $f$ from the data distribution to the latent distribution is learned. The function $f$ needs to be invertible and stable, and its inverse $f^{-1}$ maps samples from the latent distribution to the data distribution. With $f$, an unsupervised learning algorithm with exact log-likelihood computation, sampling, inference of latent variables, and an interpretable latent space, is developed to model natural images.
In \cite{cycleGAN}, a pair of transformation functions, $F$ and $G$, are introduced to be the bridges between the source domain $X$ and the target domain $Y$. Both $F$ and $G$ are unknown and learned to satisfy that $F(X)$ is indistinguishable from $Y$ and $F(G(X))\approx X$. This pair (cycle), $F$ and $G$, demonstrates great ability to transfer and enhance the photo style.
In our paper, the inverse transformation $T$ is not required to be invertible though theoretical analysis only apply for invertible $T$'s. Also, when we learn the inverse effect of $T$, $T$ is given explicitly.

\section{GAN with Inverse Transformation Unit}
\begin{figure}[!h]
  \centering
  \includegraphics[width=0.55\textwidth]{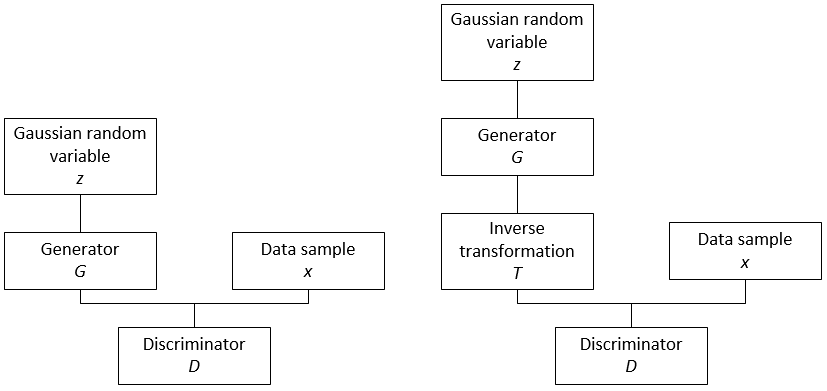}\\
  \caption{Architecture of GAN (left) and GAN with Inverse Transformation Unit (right).}\label{compare}
\end{figure}

GAN \cite{GAN} is an excellent architecture for training generative models. It includes two networks, each "fighting with" the other, and both of them are improved during the process. Specifically, the generator $G$
captures the distribution of training data while the discriminator $D$ distinguishes between samples from $G$ and the training data. In our approach, we add an Inverse Transformation Unit $T$, or a "filter" after $G$ generates a sample distribution. Figure \ref{compare} demonstrates our model compared to the original GAN \cite{GAN} architecture. $V(D,G)$ in equation (1) is our value function; we maximize it over $D$ and minimize it over $G$:

\begin{equation}\min_G\max_D V(D,G)=\mathbb{E}_{x\sim p_{data}(x)}\log D(x)+\mathbb{E}_{z\sim p_z(z)}\log\left(1-D(T(G(z)))\right).\end{equation}

Here is an intuitive explanation for the name of $T$, the inverse transformation unit. If we train on $\widetilde{G}=T\circ G$, this is exactly the original GAN, and $\widetilde{G}$ will learn the probability distribution of training data. In this sense, the generator $G$ is creating samples that contain information of "inverse of $T$", if it exists. For example, the generator $G$ learns how to generate dogs, while the discriminator $D$ learns to judge if it's a true image of dog. Suppose $T$ makes the image blurred. Since $T\circ G$ learns the distribution of true dogs, $G$ will generate samples that are clear enough to eliminate the blurring effect.

However, things are complicated when $T^{-1}$ doesn't exist. It might be the case that $G$ learns information of $\hat{T}$ where $\hat{T}\circ T$ is almost identity mapping; however, $G$ may also fail to learn it. In the rest part of the paper, both theocratical analysis and experiments are made to investigate such situations.

\section{Theoretical Results}
 In this section, we show that when $T$ is bijection with invertible Jacobian matrix, then the generator $G$ does create samples similar to $T^{-1}$ of data. The optimal discriminator $D_G^*$ is given explicitly, and the convergence is analyzed. However, when $T$ is not bijection, the optimal discriminator either does not exist or cannot be written explicitly. Two conjectures are posted about the optimal discriminator when $G$ fixed, with respect to two situations when $T$ is not surjection/injection.

\textbf{Theorem 1.} \textit{Suppose the transformation function $T$ is a bijection from $\mathbb{R}^n$ to $\mathbb{R}^n$. If $T$ has an invertible Jacobian matrix $J$, then for $G$ fixed, the optimal discriminator $D$ is given by}
\begin{equation}D_G^*(x)=\frac{p_{data}(x)}{p_{data}(x)+p_g(T^{-1}(x))|J^{-1}(x)|},\ a.e.\end{equation}

\textit{Proof. } For $G$ fixed, the discriminator $D$ is trained to maximize
\begin{equation}\begin{array}{ll}
V(D,G)&=\displaystyle\int_{\mathbb{R}^n}p_{data}(x)\log D(x)\mathrm{d} x+\int_{\mathbb{R}^n}p_z(z)\log\left(1-D(T(G(z)))\right)\mathrm{d} z\\
&\\
&\displaystyle=\int_{\mathbb{R}^n}\Big(p_{data}(x)\log D(x)+p_g(x)\log\left(1-D(T(x))\right)\Big)\mathrm{d} x.
\end{array}\end{equation}
The variational method is used to solve the problem. For any $H:\mathbb{R}^n\rightarrow\mathbb{R}^n$ with compact support, function $h$ is defined by $h(t)=V(D+tH,G)$, $t\in\mathbb{R}$. Then the optimal discriminator can be found by solving the equation $h'(0)=0$ with constraint $h''(0)<0$.

The function $h(t)$ can be expanded as
\begin{equation}h(t)=\int_{\mathbb{R}^n}\Big(p_{data}(x)\log \left(D(x)+tH(x)\right)+p_g(x)\log\left(1-D(T(x))-tH(T(x))\right)\Big)\mathrm{d} x.\end{equation}
Thus, $h'(t)$ can be calculated as
\begin{equation}h'(t)=\int_{\mathbb{R}^n}\left(p_{data}(x)\cdot\frac{H(x)}{D(x)+tH(x)}+p_g(x)\cdot\frac{-H(T(x))}{1-D(T(x))-tH(T(x))}\right)\mathrm{d} x.\end{equation}
Let $h'(0)=0$, we have
\begin{equation}0=h'(0)=\int_{\mathbb{R}^n}\left(p_{data}(x)\cdot\frac{H(x)}{D(x)}+p_g(x)\cdot\frac{-H(T(x))}{1-D(T(x))}\right)\mathrm{d} x.\end{equation}
Through substitutions $y=T(x)$ and $x=T^{-1}(y)$, we have
\begin{equation}\int_{\mathbb{R}^n}p_g(x)\cdot\frac{-H(T(x))}{1-D(T(x))}\mathrm{d} x
=\int_{\mathbb{R}^n}p_g(T^{-1}(y))\cdot\frac{-H(y)}{1-D(y)}\cdot|J^{-1}(y)|\mathrm{d} y.
\end{equation}
As a result,
\begin{equation}0=h'(0)=\int_{\mathbb{R}^n}\left(p_{data}(x)\cdot\frac{H(x)}{D(x)}-p_g(T^{-1}(x))\cdot\frac{H(x)}{1-D(x)}\cdot|J^{-1}(x)|\right)\mathrm{d} x.\end{equation}
Since $H$ is arbitrary, it follows that
\begin{equation}\frac{p_{data}(x)}{D(x)}-\frac{p_g(T^{-1}(x))}{1-D(x)}\cdot|J^{-1}(x)|=0,\ a.e.,\end{equation}
which leads to the result that the optimal discriminator $D$ is given by
\begin{equation}D_G^*(x)=\frac{p_{data}(x)}{p_{data}(x)+p_g(T^{-1}(x))|J^{-1}(x)|},\ a.e.\end{equation}
Additionally, $h''(0)<0$ is trivial. $\Box$

\textbf{Theorem 2.} \textit{Let $C(G)=\max_{D}V(D,G)=V(D_G^*,G)$. The global minimum of $C(G)$ is achieved if and only if $p_{data}=|J^{-1}|p_g\circ T^{-1}$; at that point, $C(G)$ achieves the value $-\log 4$.}

\textit{Proof. } Let $\widetilde{G}=T\circ G$ be the transformed generator, and we see that $p_g$ and $p_{\tilde{g}}$ has the following relationship according to Theorem 1:
\begin{equation}p_{\tilde{g}}=|J^{-1}|p_g\circ T^{-1}.\end{equation}
Then, the rest of the proof is exactly the same as the proof of Theorem 1 (not Theorem 1 in this paper) by Goodfellow, et al. \cite{GAN} $\Box$

Furthermore, if we change the view using equation 11, the convergence is guaranteed under the same conditions of Proposition 2 by Goodfellow, et al\cite{GAN}.

When $T$ is not bijection, things are much more complicated. Usually, we can't find the optimal discriminator $D_G^*$ for $G$ fixed. Following is some analysis as well as two conjectures.
When $T$ is not surjection, let the range of $T$ is $\mathbb{R}^n\setminus A$. Then, the right side of equation 7 will be the integration on $\mathbb{R}^n\setminus A$, which leads to the result that equation 8 is changed to
\begin{equation}\begin{array}{ll}
0=h'(0)=&\int_{\mathbb{R}^n\setminus A}\left(p_{data}(x)\cdot\frac{H(x)}{D(x)}-p_g(T^{-1}(x))\cdot\frac{H(x)}{1-D(x)}\cdot|J^{-1}(x)|\right)\mathrm{d} x\\&+\int_{A}p_{data}(x)\cdot\frac{H(x)}{D(x)}\mathrm{d} x.\end{array}
\end{equation}
However, we usually can't make $\frac{p_{data}(x)}{D(x)}=0$ for $x\in A$.

\textbf{Conjecture 1. } \emph{When $T$ is not surjection, there is no explicit optimal discriminator for $G$ fixed. However, the following condition should be satisfied for a good discriminator (i.e. if a discriminator doesn't satisfy the following condition, it's always better to change it into this condition):}
\begin{equation}D_G^*(x)=\frac{p_{data}(x)}{p_{data}(x)+p_g(T^{-1}(x))|J^{-1}(x)|},x\in \mathbb{R}^n\setminus A,\ a.e.\end{equation}

When $T$ is not injection, define the set $T^{-1}(x)=\{y\in\mathbb{R}^n:T(y)=x\}$. Then, in equation 7 the expression $p_g(T^{-1}(y))$ doesn't exist anymore. Instead, it is $p_g(t_y)$, where $t_y\in T^{-1}(y)$ is a point in the set but we don't know what it is. This makes the system hard to analysis. Another point of view is that suppose $T(x_1)=T(x_2)=y_0$, then
the following two discriminators perform exactly the same:
\begin{equation}
D_G^1(y_0)=\frac{p_{data}(y_0)}{p_{data}(y_0)+p_g(x_1)|J^{-1}(y_0)|};\
D_G^2(y_0)=\frac{p_{data}(y_0)}{p_{data}(y_0)+p_g(x_2)|J^{-1}(y_0)|}.
\end{equation}

\textbf{Conjecture 2.} \emph{When $T$ is not injection, there is no explicit optimal discriminator for $G$ fixed. However, the following discriminator is good enough (i.e. there might be better solutions in different situations, but it's the best one that can be written explicitly):
\begin{equation}D_G^*(x)=\frac{p_{data}(x)}{p_{data}(x)+\overline{p_g(T^{-1}(x))}|J^{-1}(x)|},\ a.e.\end{equation}
where $\overline{p_g(T^{-1}(x))}$ refers to the average value of $p_g(y)$ for $y\in T^{-1}(x)$.}

The existence of the inverse transformation unit $T$ makes generative adversarial networks possible to generate a wider range of samples with our additionally desired effects. The bijection case is proved to work well, but other cases still need deeper theoretical analysis so that we could figure out when our architecture is effective.

\section{Experiments}
In this section we made several experiments to show in which conditions our architecture is able to work and some possible effects our architecture is able to produce. First, in order to know if our architecture works (i.e., being successfully trained) for transformation functions with different properties, we made a general survey on $T$'s with different properties. Then, we showed that our architecture is able to learn the opposite effect of "blurring" with certain transformations. Specifically, it is able to generate sharpened images and generate recovered images that were initially blurred. Additionally, we introduced a measurement of sharpness $\chi_s$ and verified the effect of our architecture using $\chi_s$.
Our architecture is realized by adding an inverse transformation unit based on DCGAN \cite{DCGAN}, and all experiments are done on the MNIST dataset \cite{MNIST} and the Fashion-MNIST dataset \cite{fashion_MNIST}.

\subsection{A General Survey on $T$'s with Different Properties}
\begin{figure}[!h]
  \centering
  \includegraphics[width=0.4\textwidth]{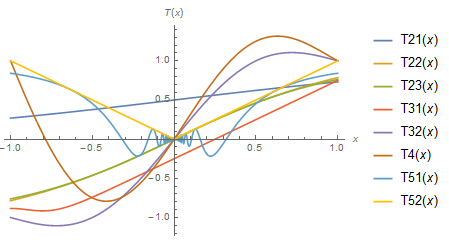}\\
  \caption{Plots of various $T$'s surveyed in Table
  \ref{all_T_table}.}\label{all_T}
\end{figure}

The intuition is that if $T$ is bijection, the training is likely to be successful. This intuition is also supported by Theorem 2 in that the global minimum is achieved with $p_{data}$ given explicitly. But what if $T$ is not bijection? This question leads to a survey on the effect through various $T$'s with a set of properties shown in Table \ref{all_T_table}, where  $\hat{I}_{n\times n}(i,j)=1$ if $i+j=n$ and otherwise 0,
and $\sigma(x)=1/(1+\exp(-x))$. Since the images are gray and each pixel can be compressed into interval $[-1,1]$, all $T$'s are mappings from $[-1,1]$ to $[-1,1]$. The plots of these functions are demonstrated in Fig \ref{all_T}.

\begin{figure}[!h]
  \centering
\fbox{\includegraphics[width=0.18\textwidth]{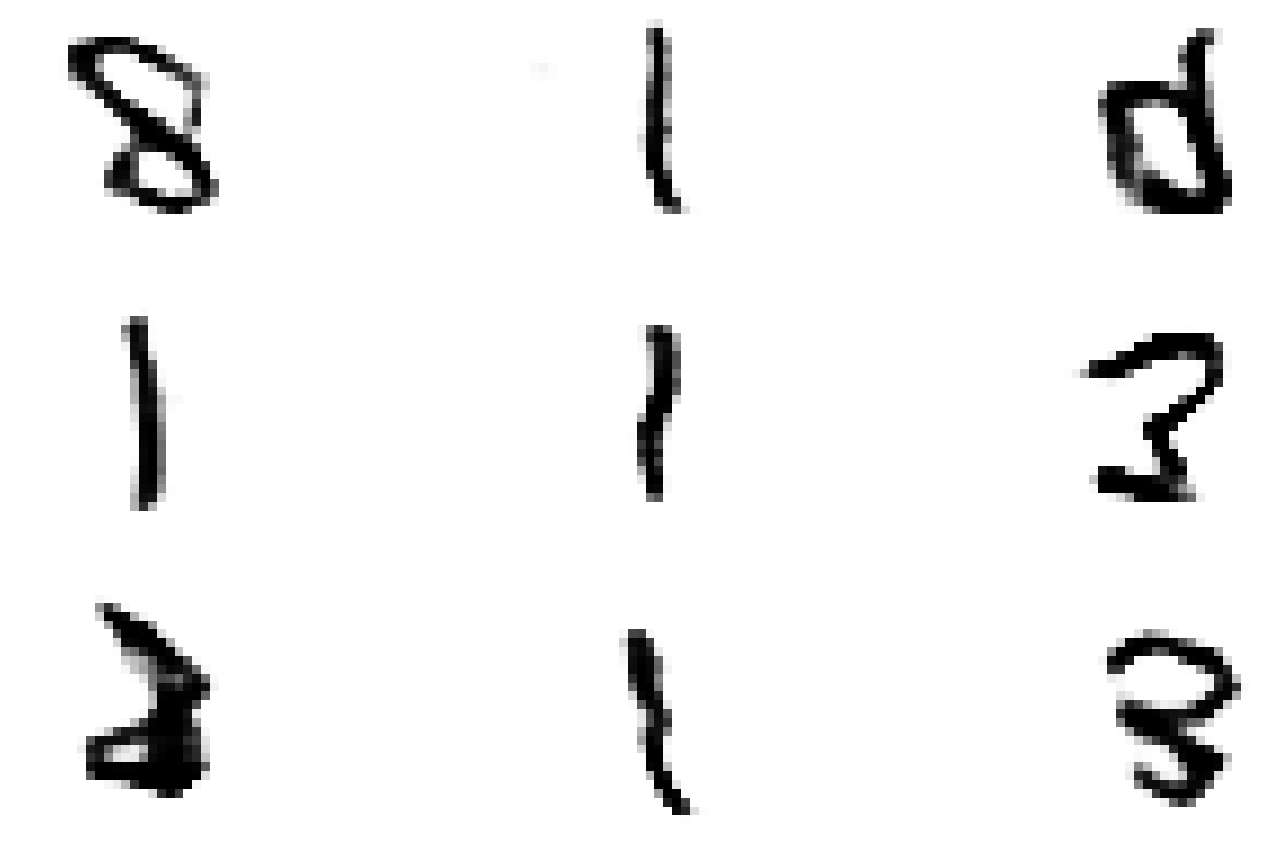}}
\fbox{\includegraphics[width=0.18\textwidth]{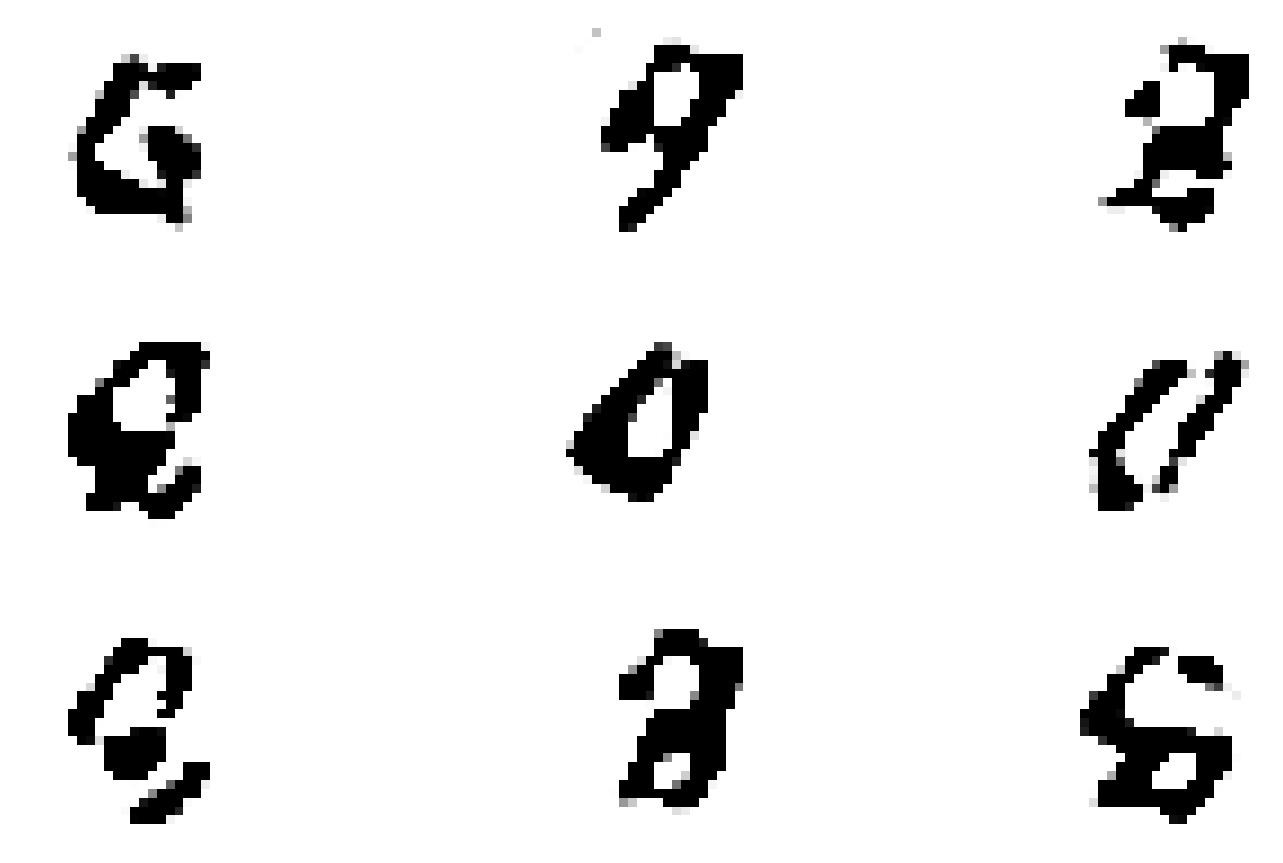}}
\fbox{\includegraphics[width=0.18\textwidth]{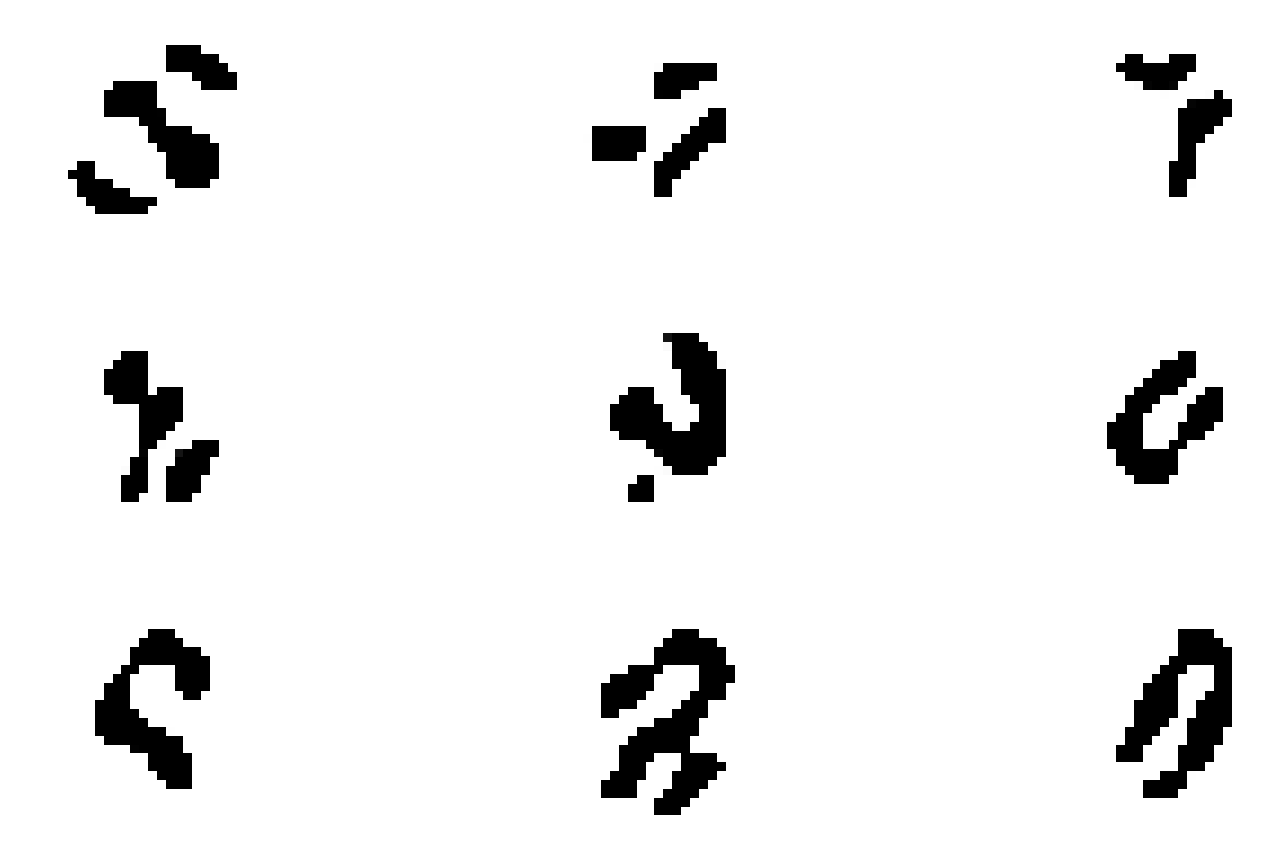}}
\fbox{\includegraphics[width=0.18\textwidth]{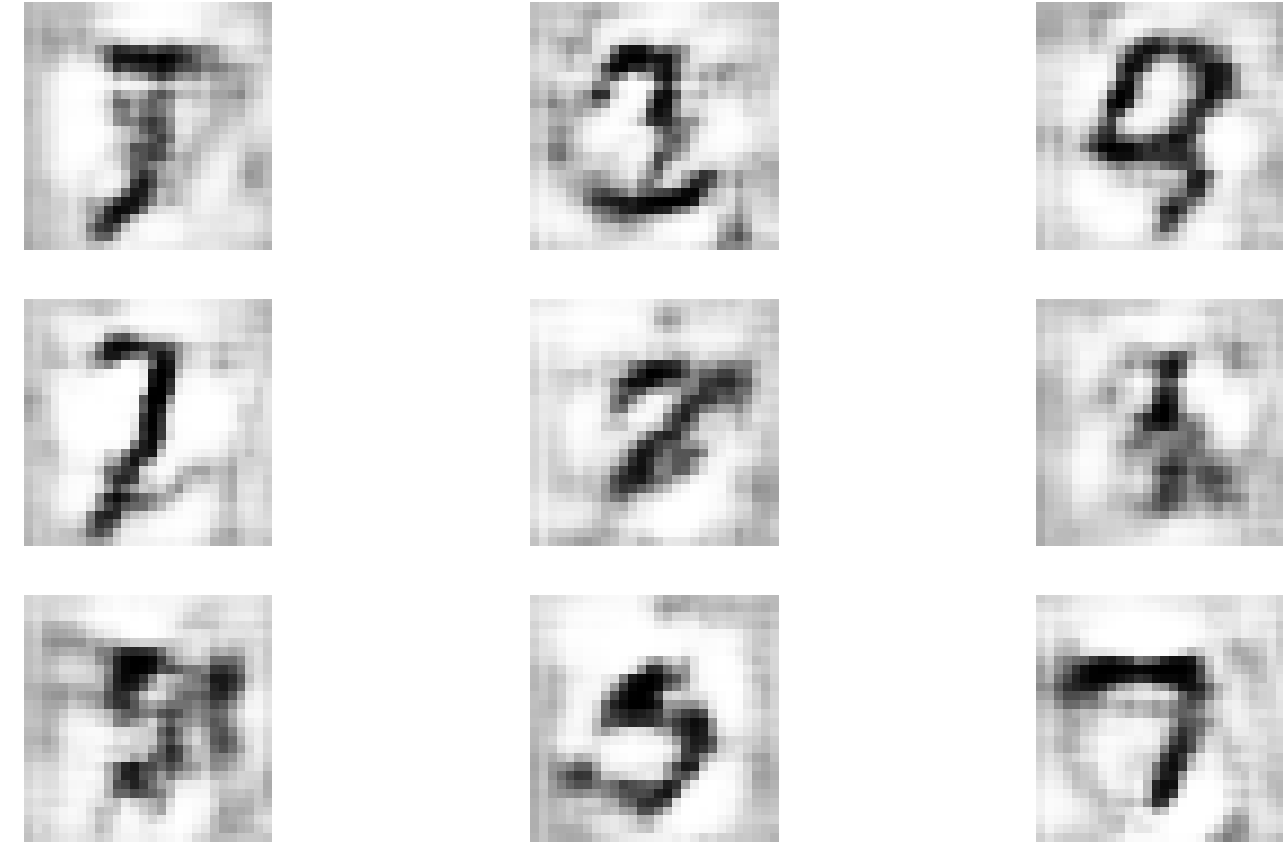}}
  \caption{Sample images generated by the models with $T_1$, $T_{22}$, $T_{23}$, $T_{32}$. The models are trained with 5000 images from the MNIST dataset.}
  \label{effect}
\end{figure}

Among the nine transformation functions selected in Table \ref{all_T_table}, four demonstrate good effects (i.e., the corresponding models are able to generate the numbers with desire effect). With $T_1$, a bijection that maps an image to its mirror image, the model indeed generates the mirror image of the original numbers. With $T_{22}$ and $T_{23}$, two injections that compress the interval $[0,1]$, the models are able to generate numbers with stronger contrast: more white and black but less gray pixels. With $T_{32}$, a surjection that is not one-to-one, the model can also generate images of numbers. Fig \ref{effect} shows the images these four models generate and images transformed by the transformation functions. As we can see, easier transformation functions are more likely to take effect, while complicated ones will have more problems during the training process, leading to bad local optimum or misconvergence. Specifically, $T_{21}$ fails because it cannot reach negative values, and for $T_4, T_{51}, T_{52}$, there is a large range of $y\in[0,1]$ that is reached by at least two $x$'s through the transformation, which confuses the model.

\subsection{Sharpening and Recovery of Blurred images}
\begin{figure}[!h]
\centering
\includegraphics[width=0.6\textwidth]{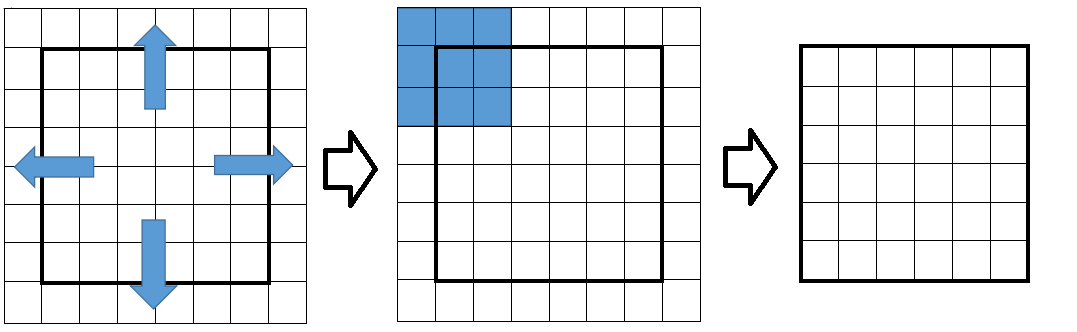}
\caption{Architecture of the blurring method using a convolution kernel.}
\label{process}
\end{figure}

One easy way to blur an image is weighted averaging each pixel and its neighbor pixels. This can be achieved by using the convolutional kernel. Consider an image as an $n\times n$ matrix $X$. To deal with the edges well, we first design a method of extension. The first step is extending a row on the top and the bottom respectively, with values of the rows next to the original edges. This yields an $(n+2)\times n$ matrix. The second step is extending a column on the left and the right, using the values of the columns next to them. This yields an $(n+2)\times (n+2)$ matrix. After the extension, we perform a convolution using a $3\times 3$ convolution kernel $K$, which yields an $n\times n$ matrix, representing the blurred image. The whole process is demonstrated in Fig \ref{process}.

\begin{figure}[!h]
\centering
{
\includegraphics[width=0.18\textwidth]{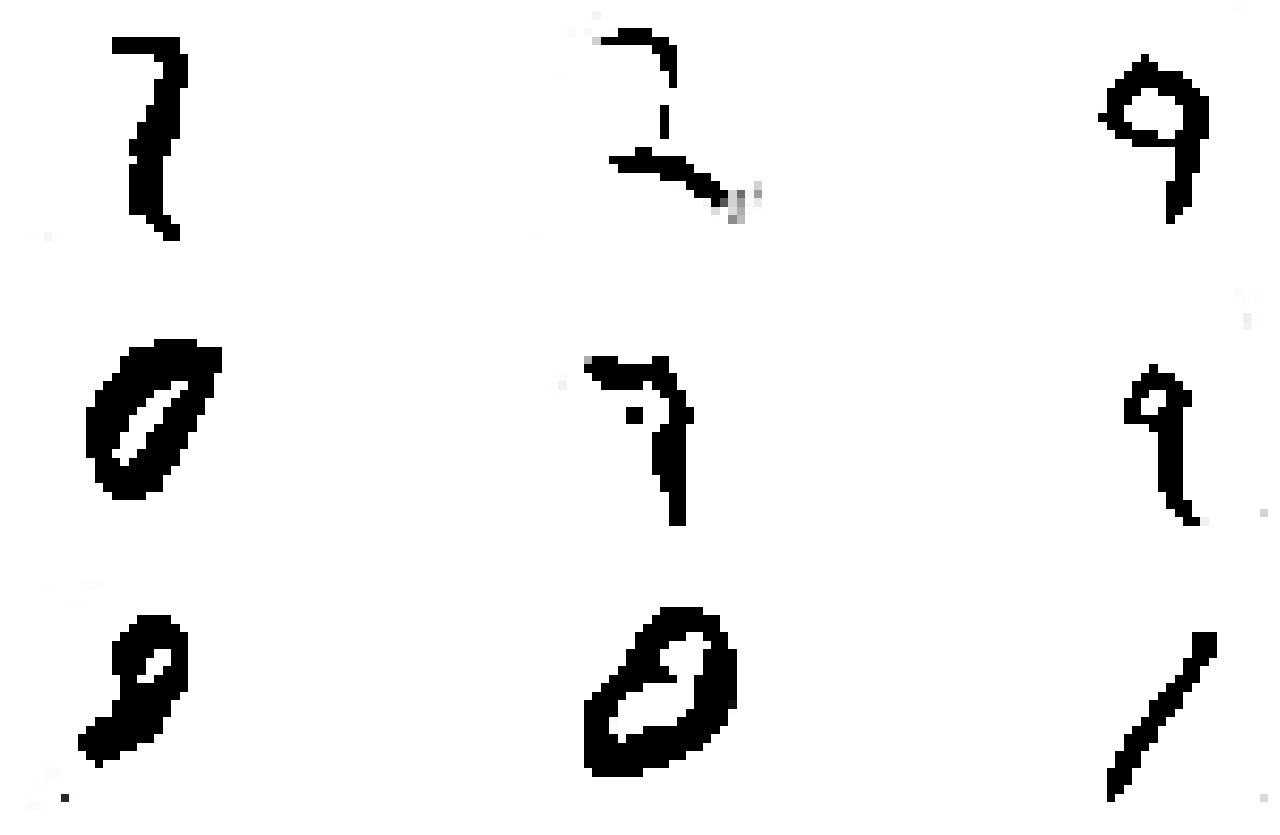}
}
{
\includegraphics[width=0.18\textwidth]{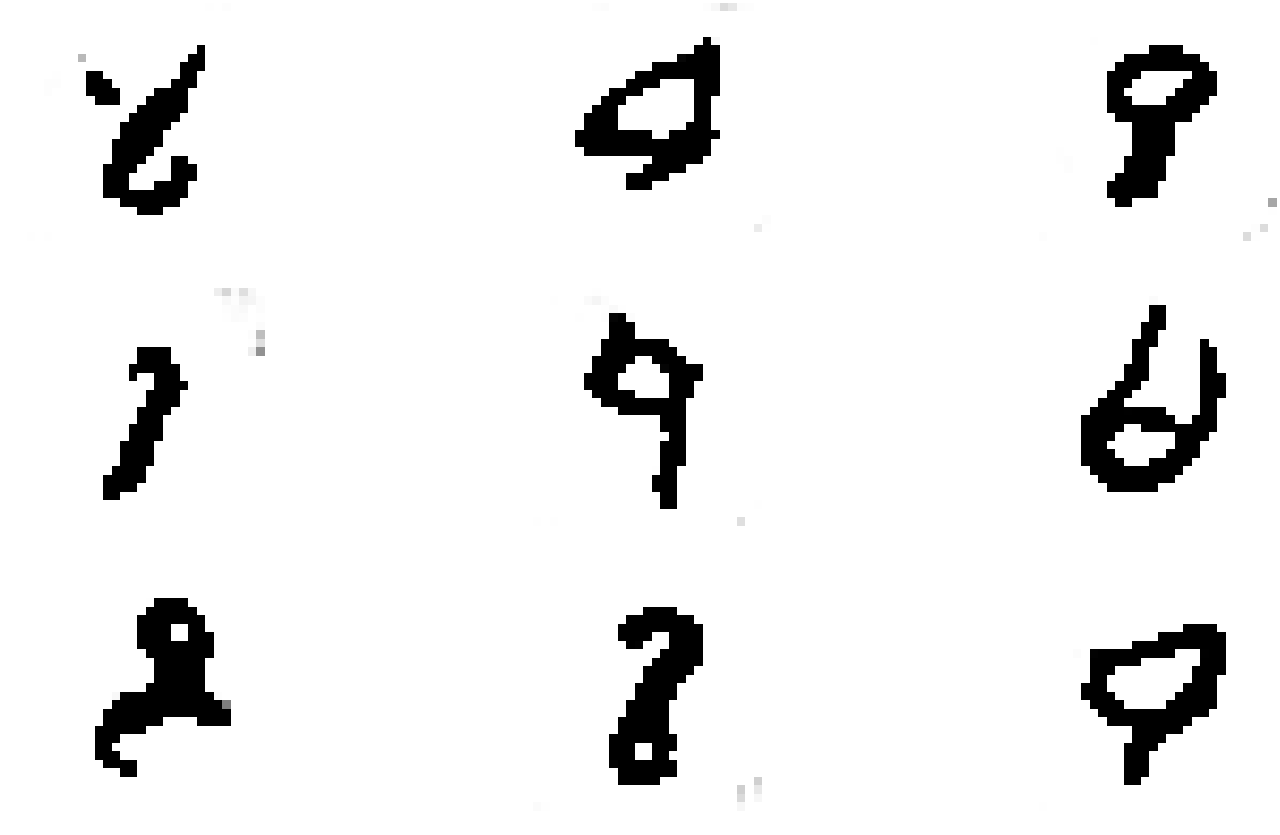}
}
{
\includegraphics[width=0.18\textwidth]{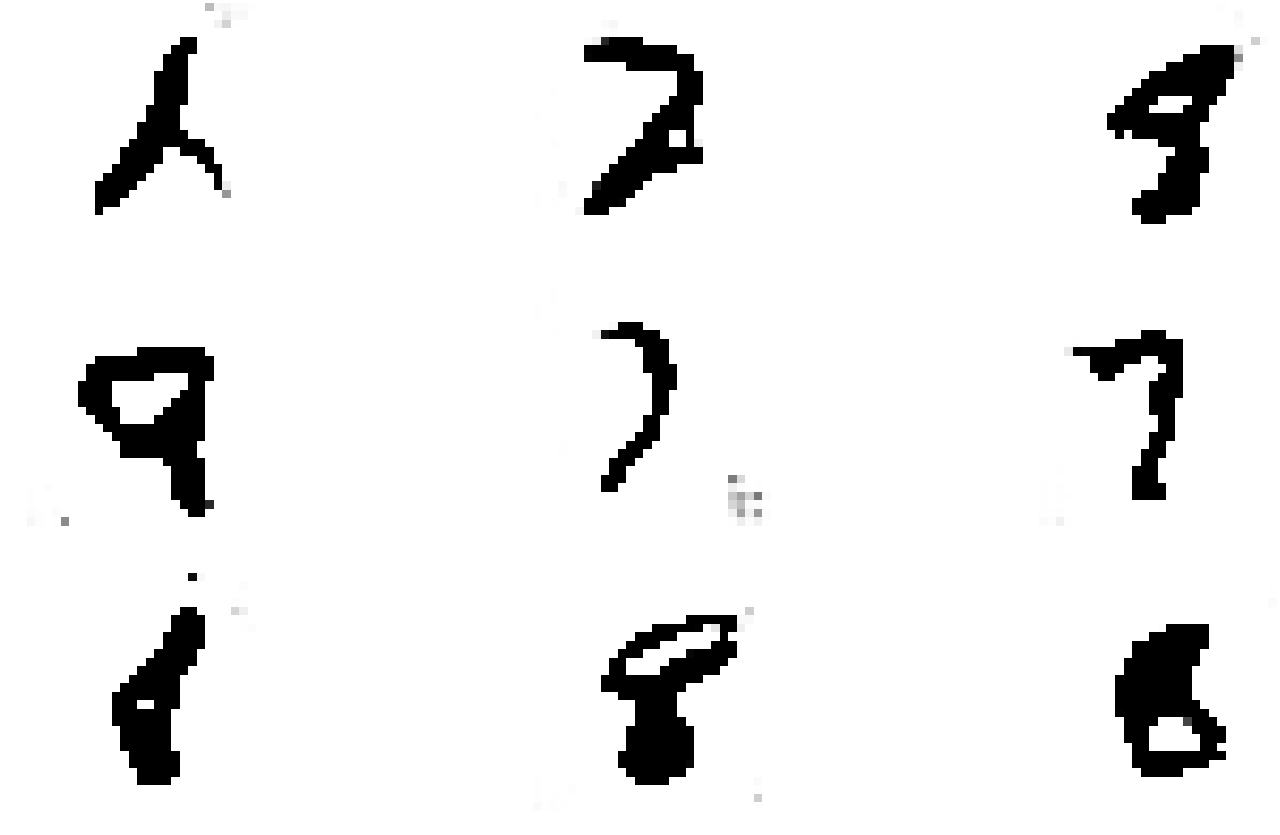}
}
{
\includegraphics[width=0.18\textwidth]{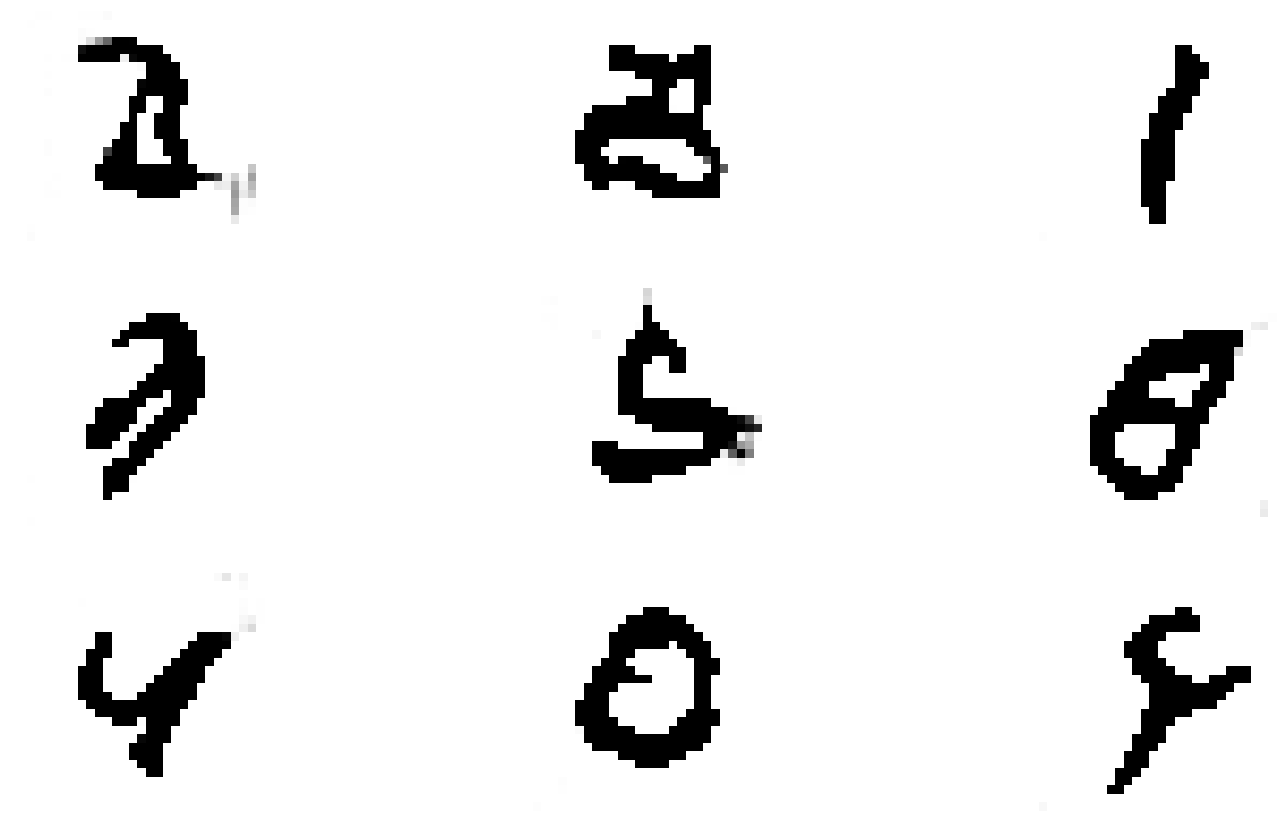}
}
\caption{Sharpened images generated by the model with convolution kernel $K_{sharpen}$. The model is trained with 10000 images from the MNIST dataset.}
\label{sharpen}
\end{figure}

\begin{figure}[!h]
\centering
{
\includegraphics[width=0.18\textwidth]{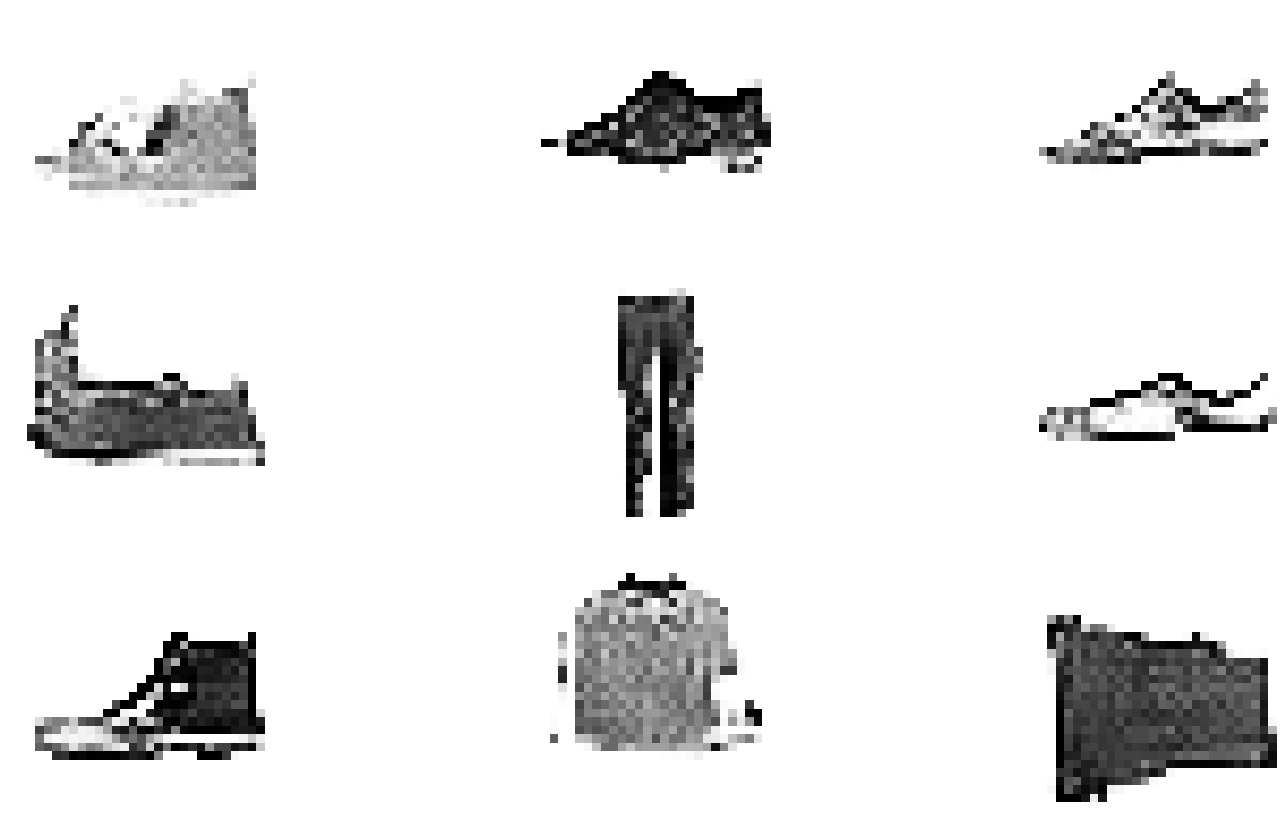}
}
{
\includegraphics[width=0.18\textwidth]{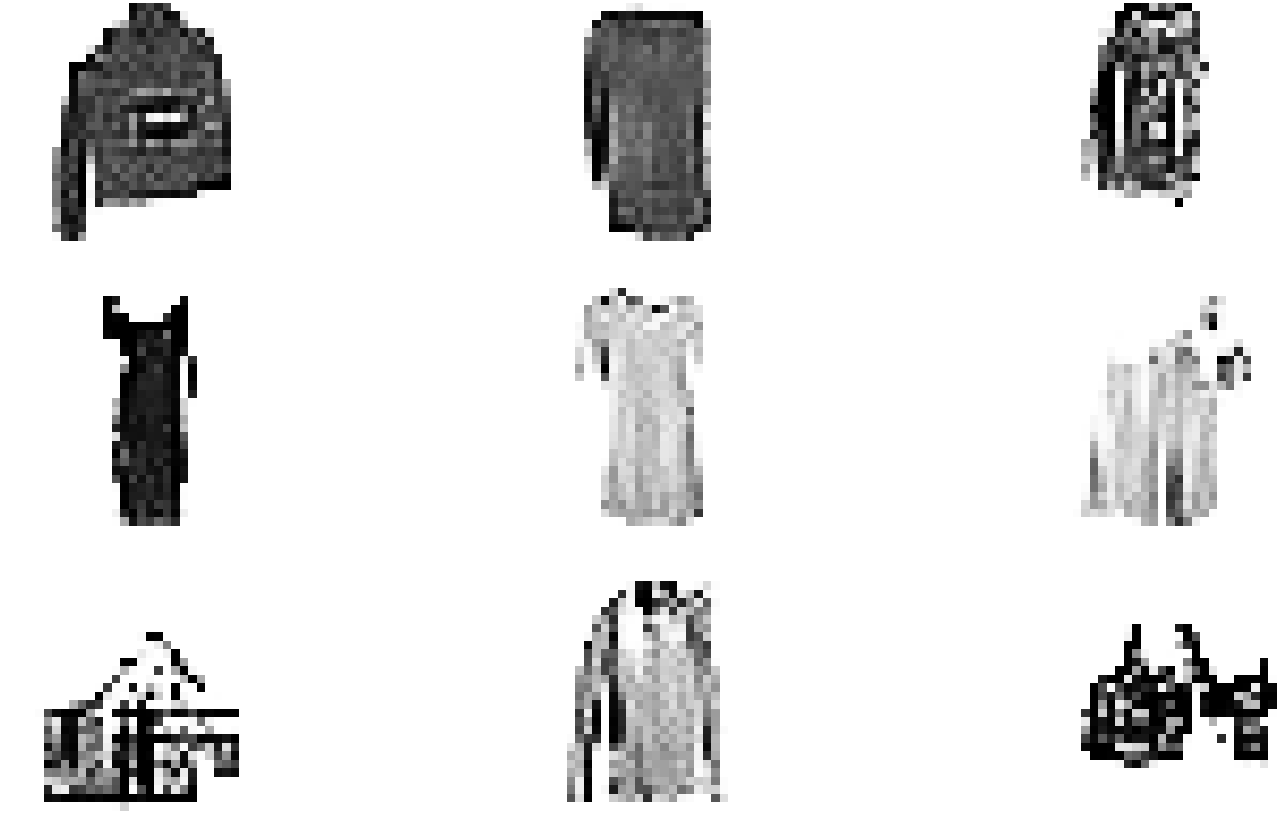}
}
{
\includegraphics[width=0.18\textwidth]{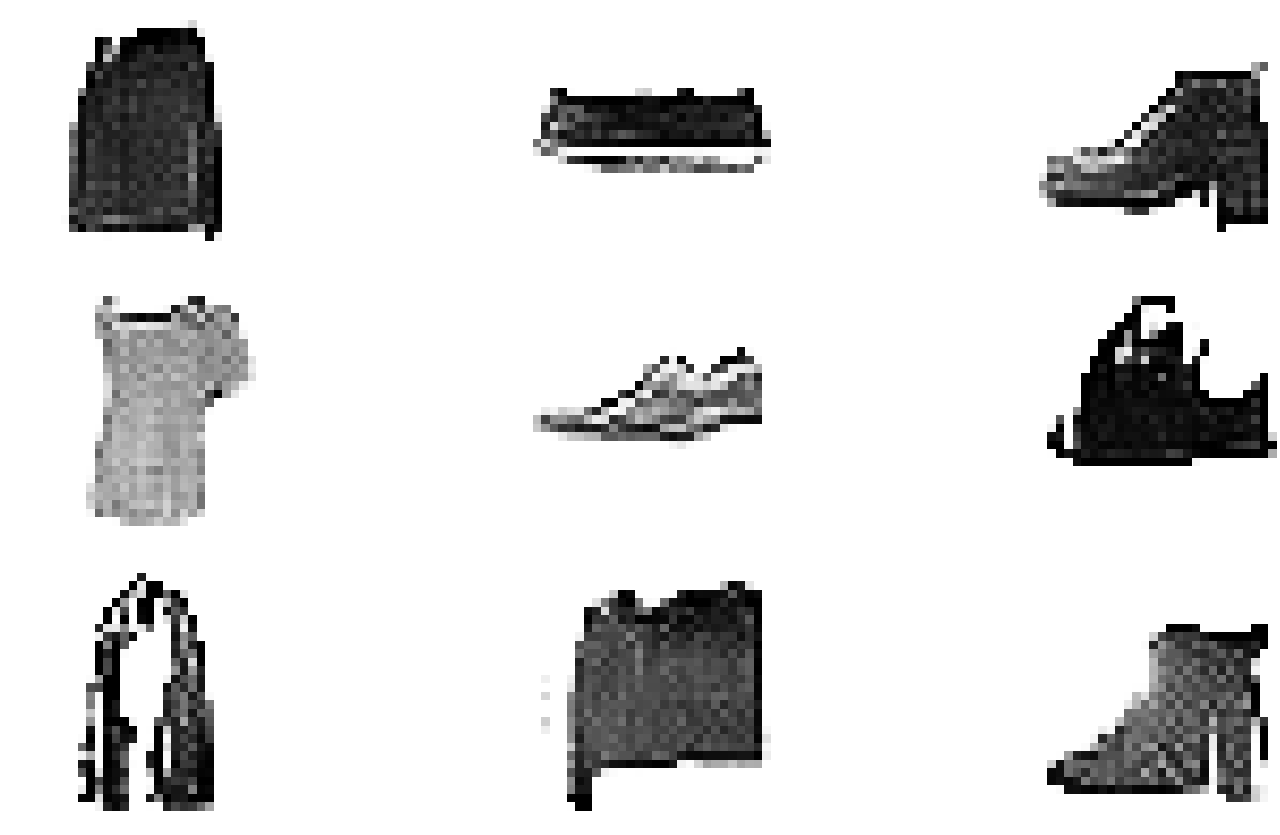}
}
{
\includegraphics[width=0.18\textwidth]{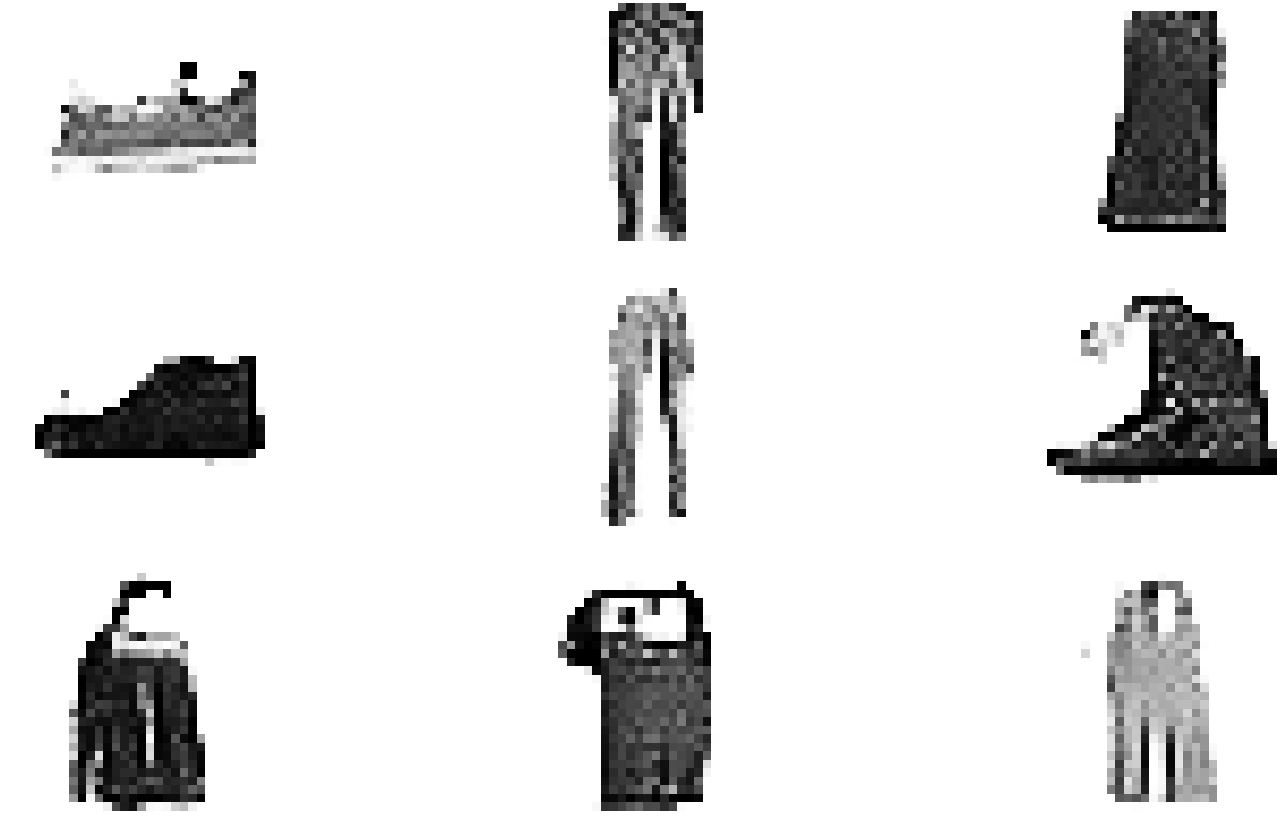}
}
\caption{Sharpened images generated by the model with convolution kernel $K_{sharpen}$. The model is trained with 20000 images from the Fashion-MNIST dataset.}
\label{fashion_sharpen}
\end{figure}

If we take this function as $T$, the generator is expected to learn the inverse effect of this blurring. In other words, the generator may learn to sharpen the images so that after $T$, the generated images are similar to the images from the dataset.
Fig \ref{sharpen} and Fig \ref{fashion_sharpen} show several samples generated by the model with convolution kernel $K_{sharpen}=\left(
\begin{array}{ccc}
 0.01 & 0.08 & 0.01 \\
 0.08 & 0.64 & 0.08 \\
 0.01 & 0.08 & 0.01 \\
\end{array}
\right)$. As expected, there are much fewer gray pixels in the images which make the image smooth. Although the effect of our model is not as obvious as that of standard techniques in image processing, the results show the ability of our architecture to deal with such tasks.

\begin{figure}[!h]
\centering
\fbox{
\includegraphics[width=0.18\textwidth]{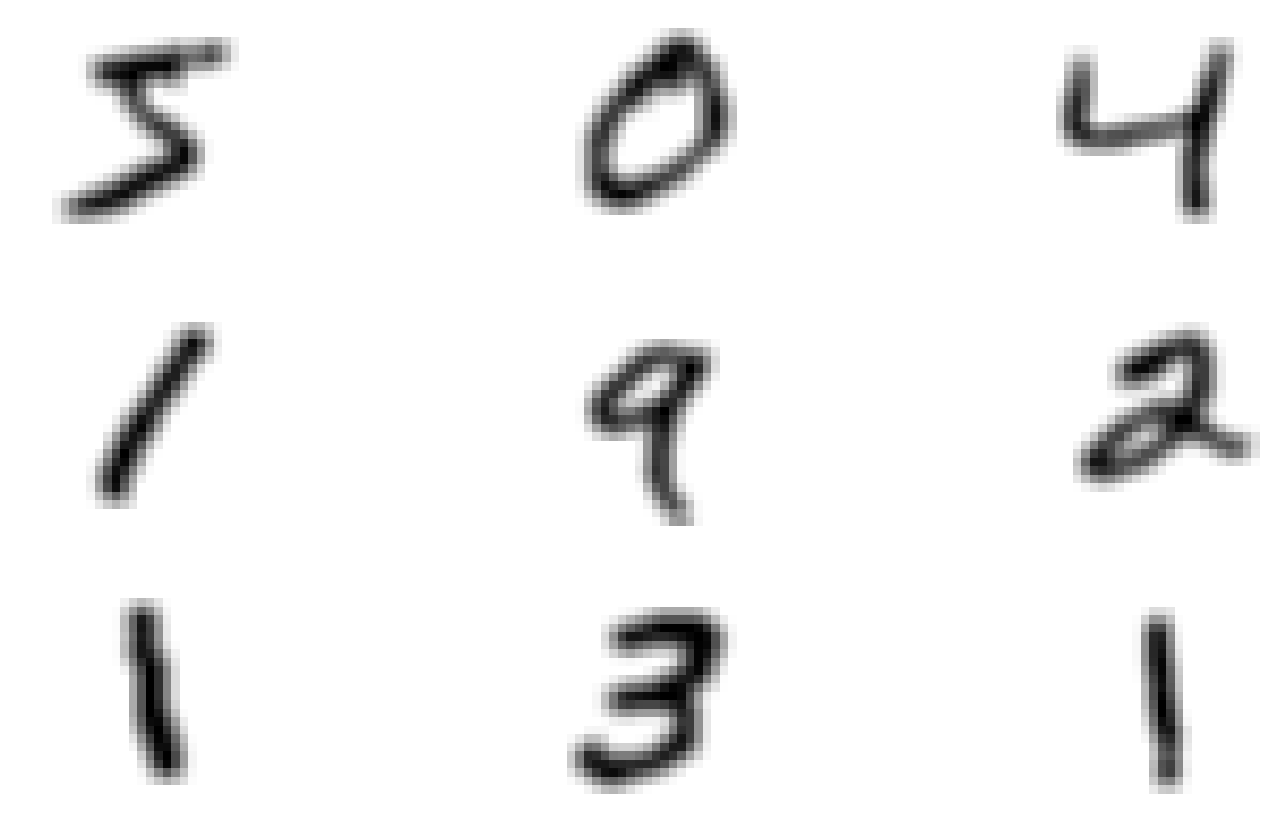}
}
{
\includegraphics[width=0.18\textwidth]{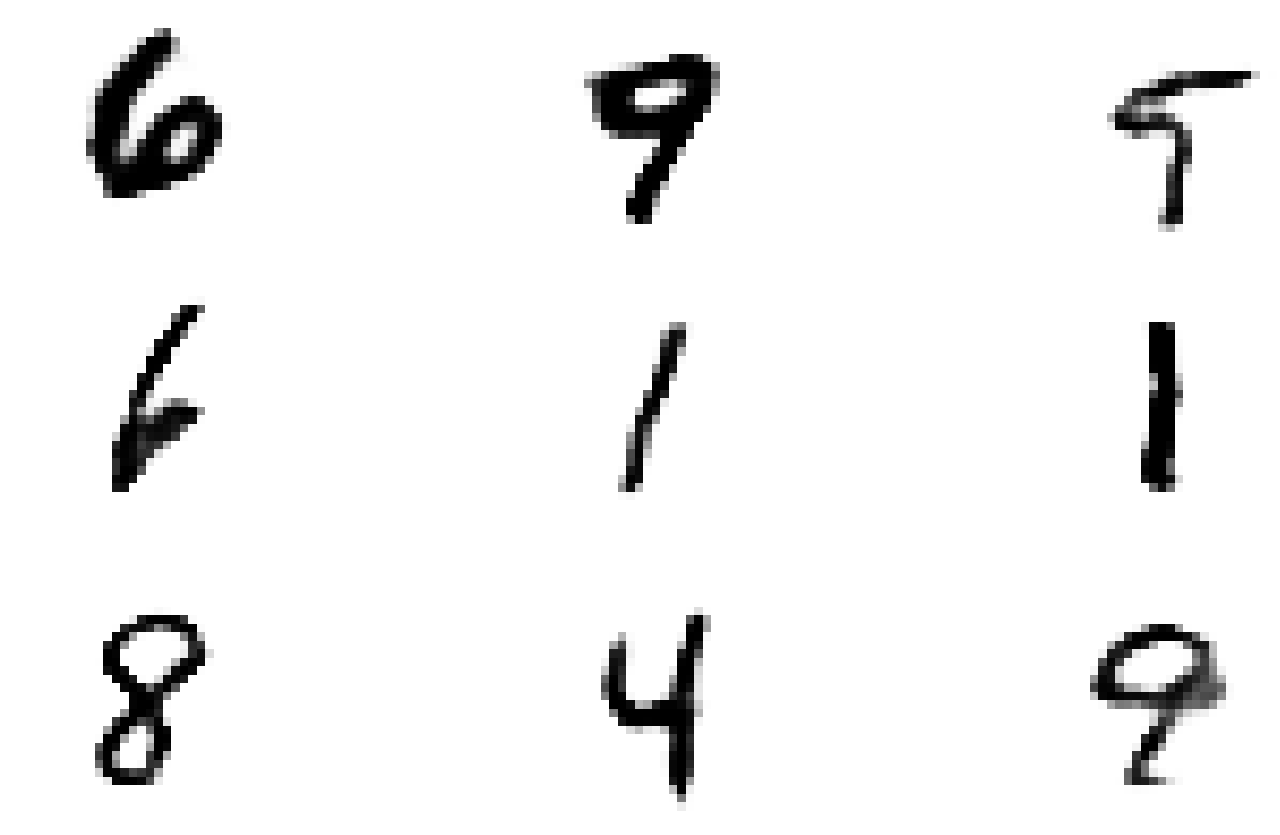}
}
{
\includegraphics[width=0.18\textwidth]{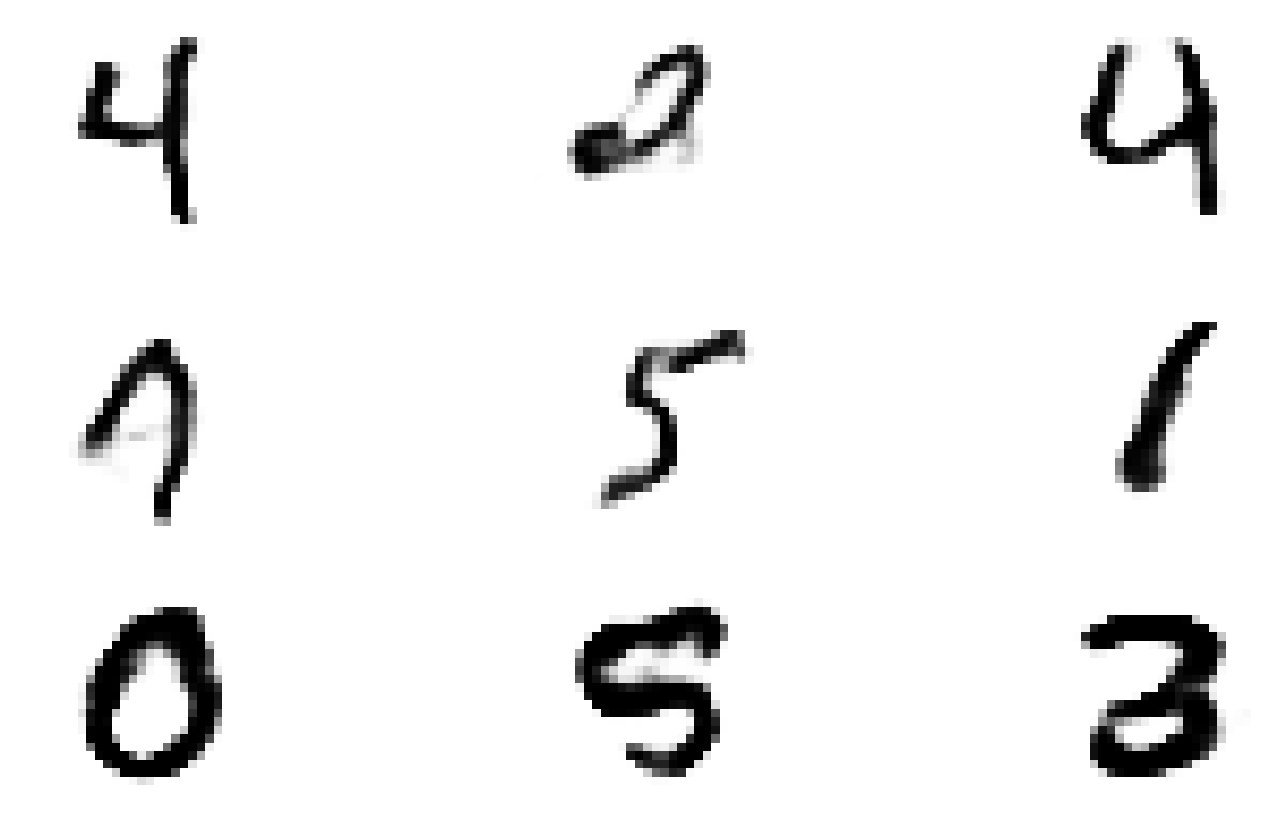}
}
{
\includegraphics[width=0.18\textwidth]{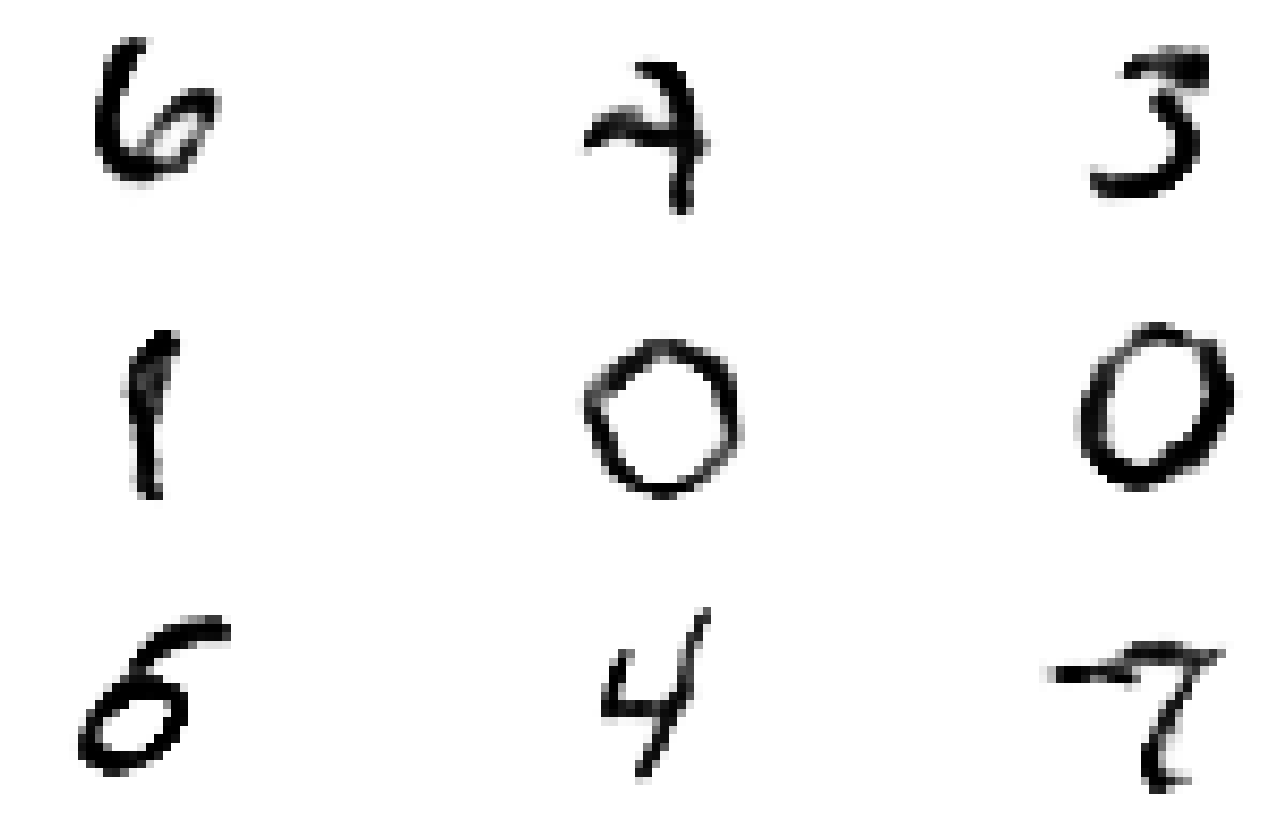}
}
\\
{
\includegraphics[width=0.18\textwidth]{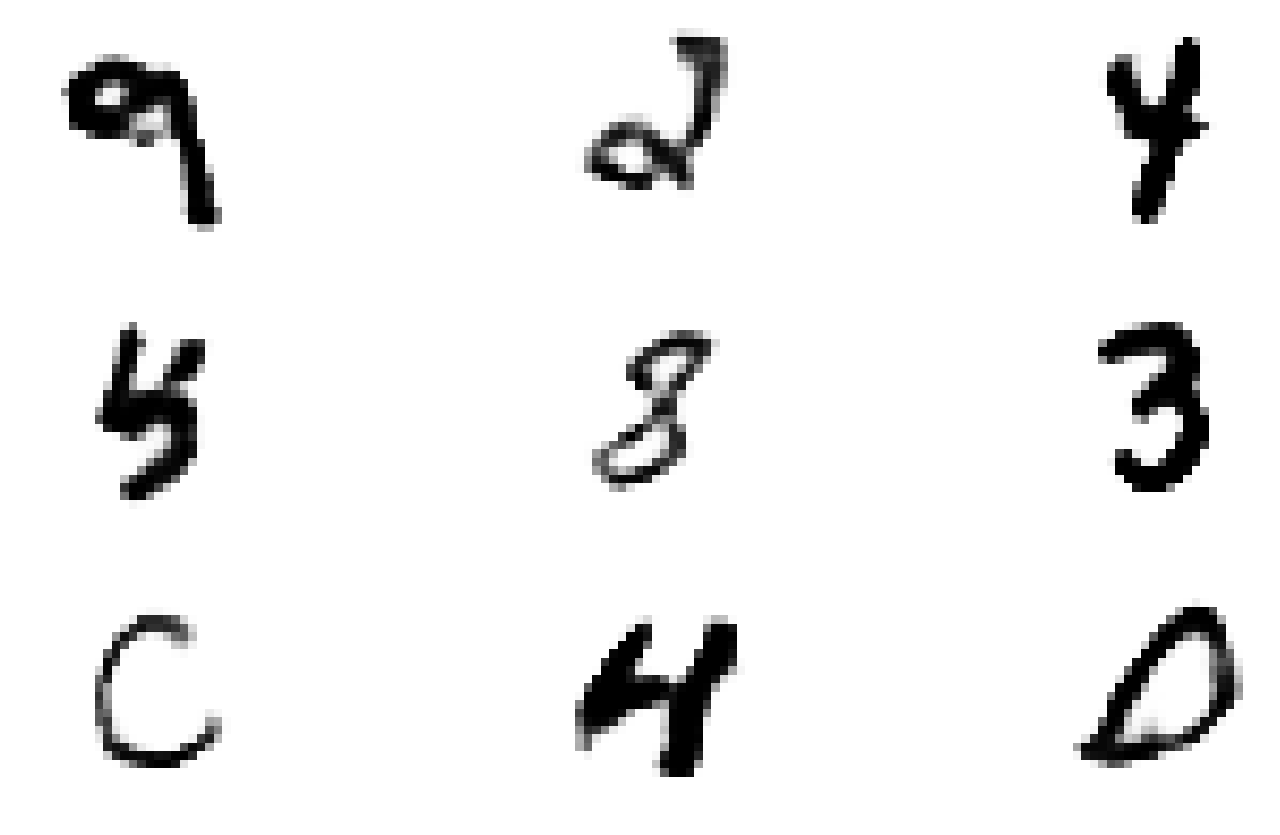}
}
{
\includegraphics[width=0.18\textwidth]{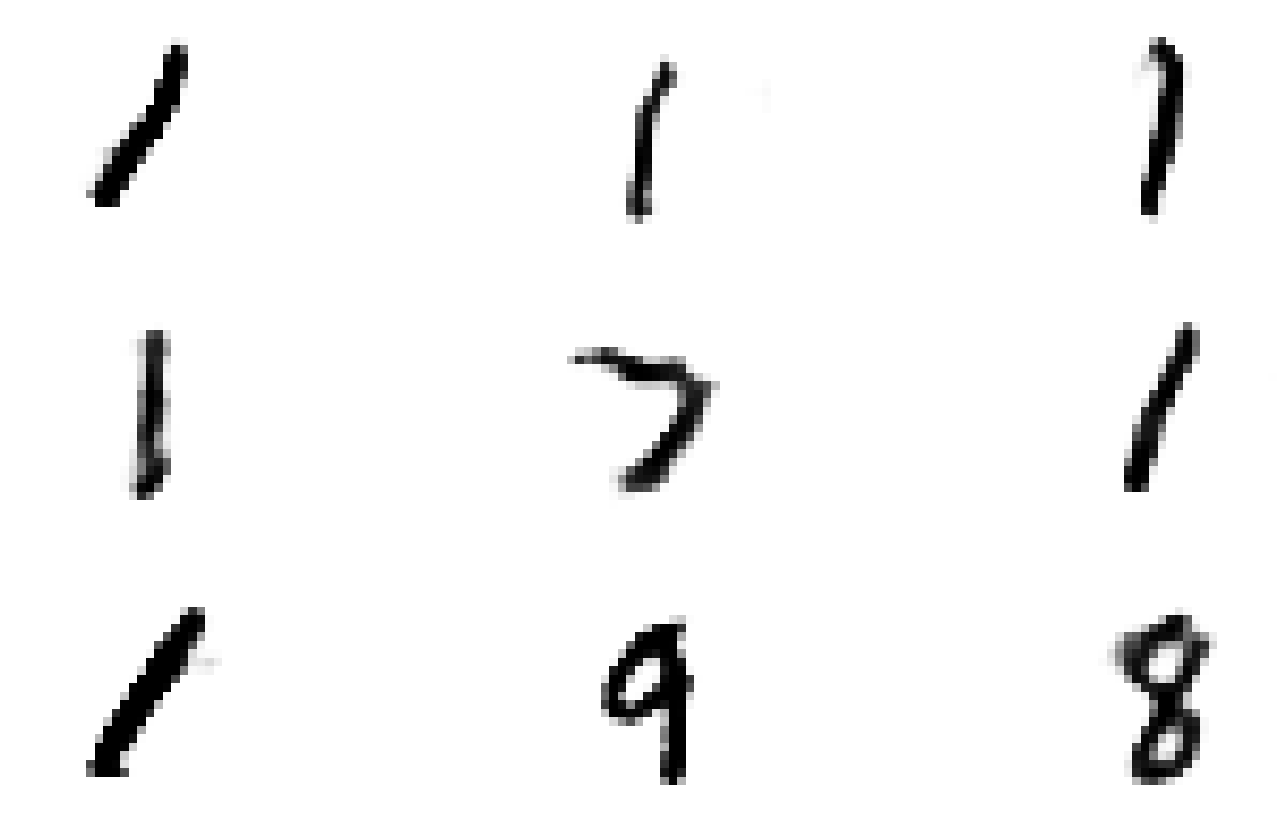}
}
{
\includegraphics[width=0.18\textwidth]{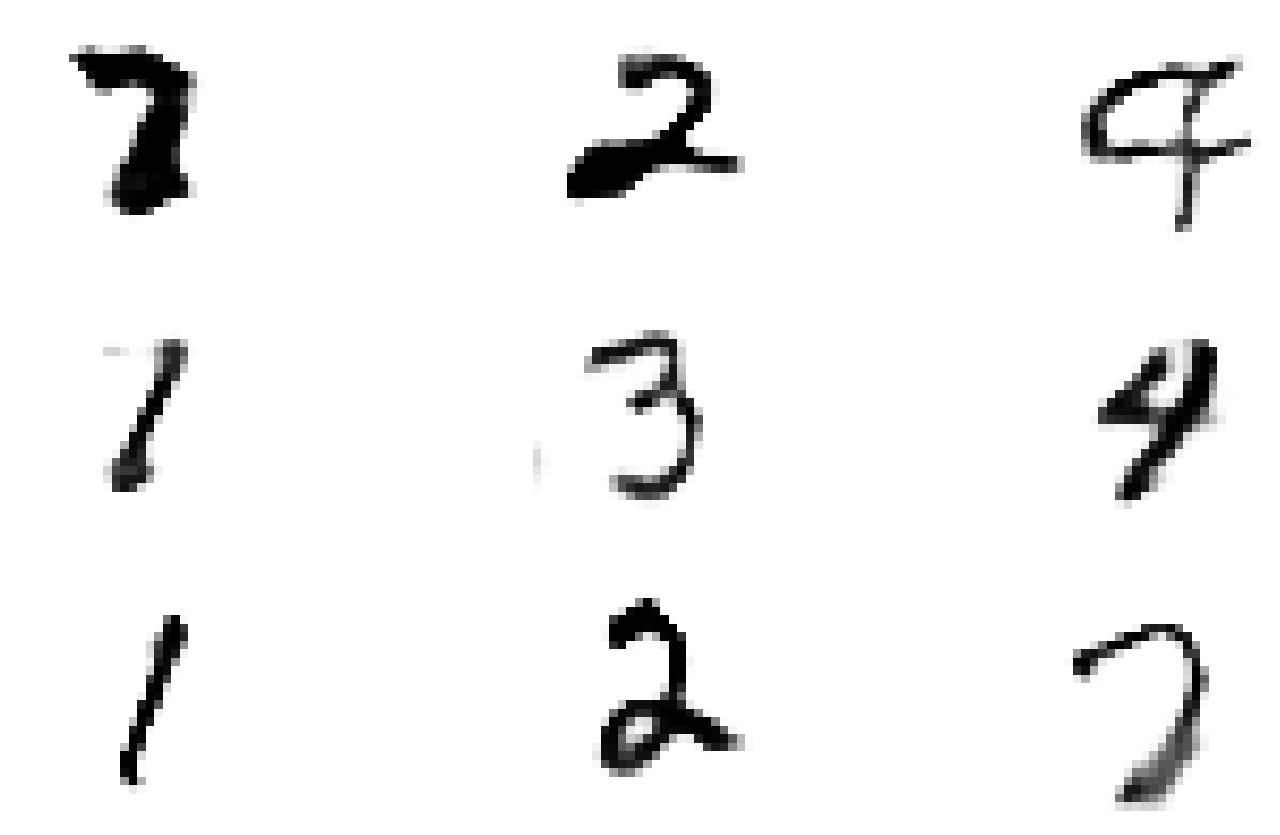}
}
{
\includegraphics[width=0.18\textwidth]{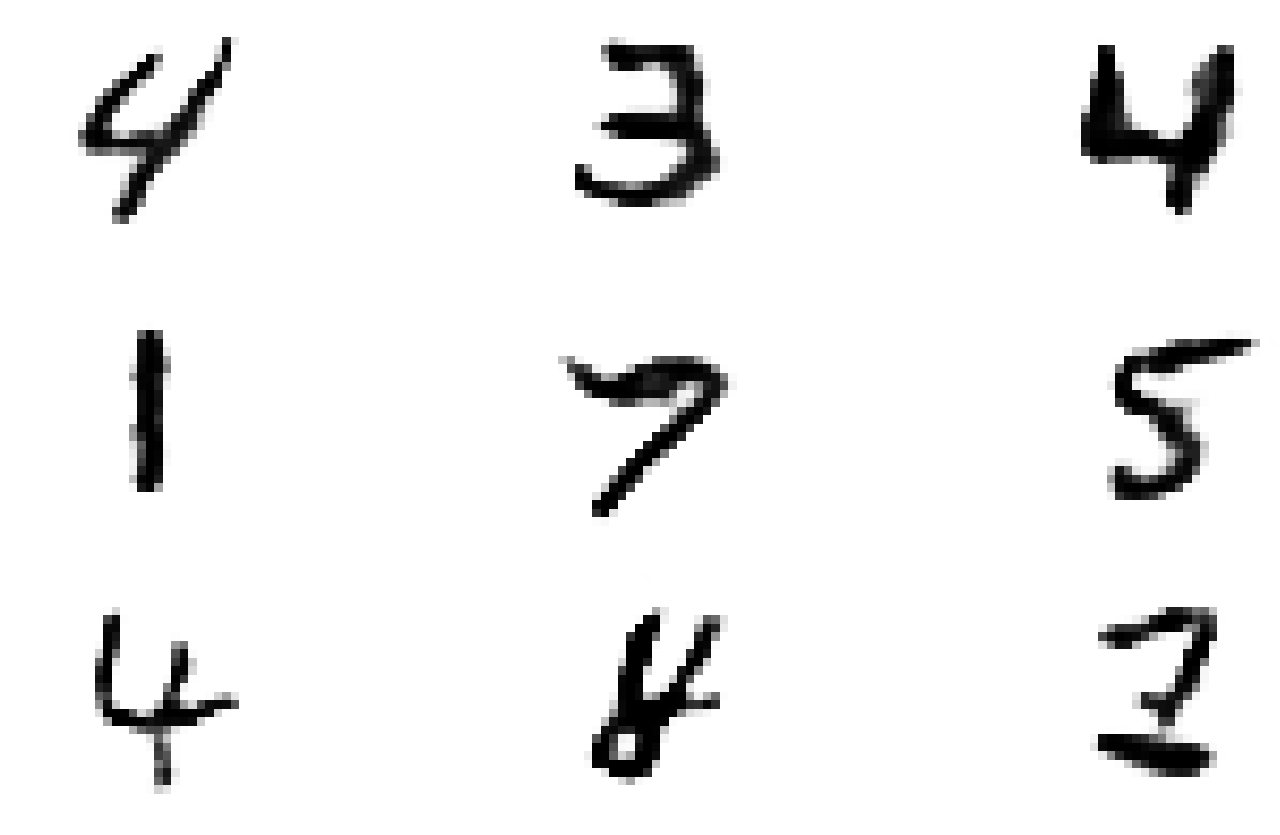}
}

\rule{0.8\textwidth}{0.3mm}\\
{
\includegraphics[width=0.18\textwidth]{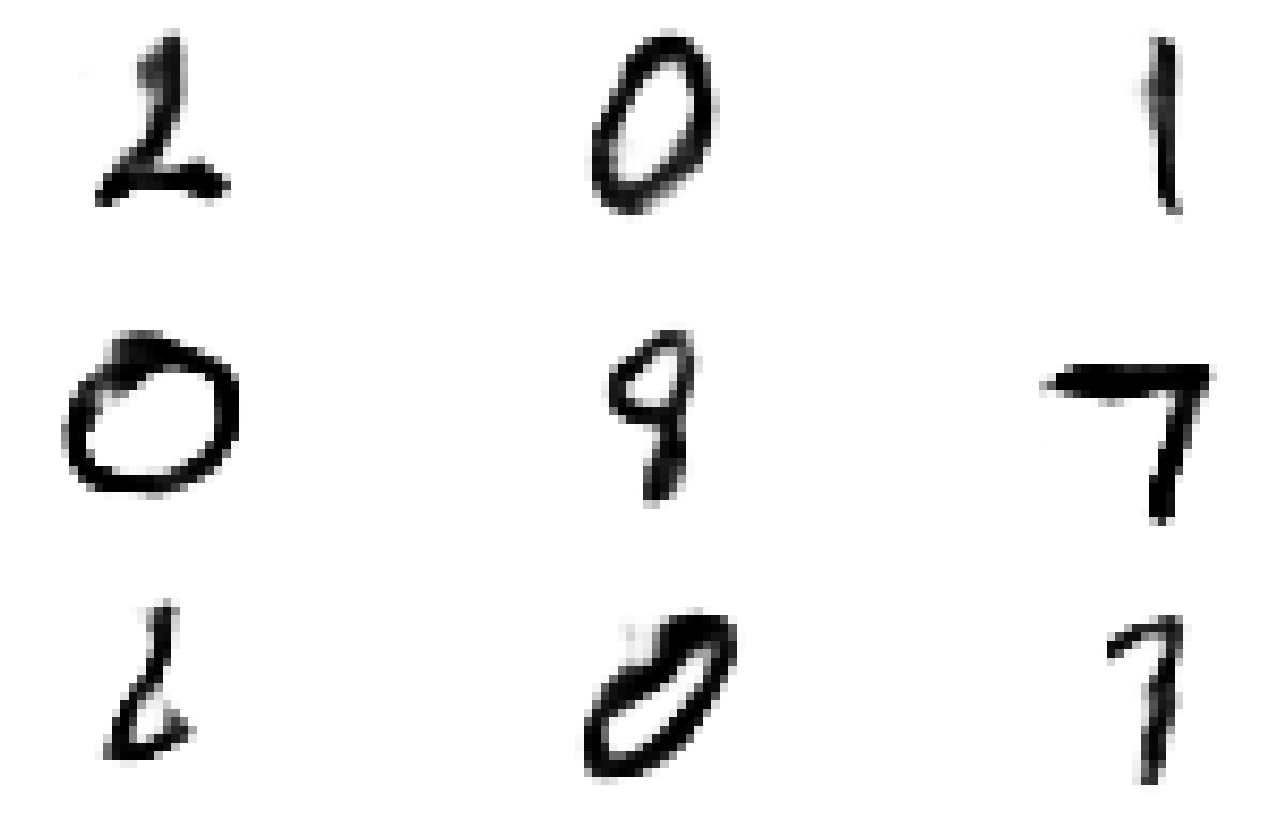}
}
{
\includegraphics[width=0.18\textwidth]{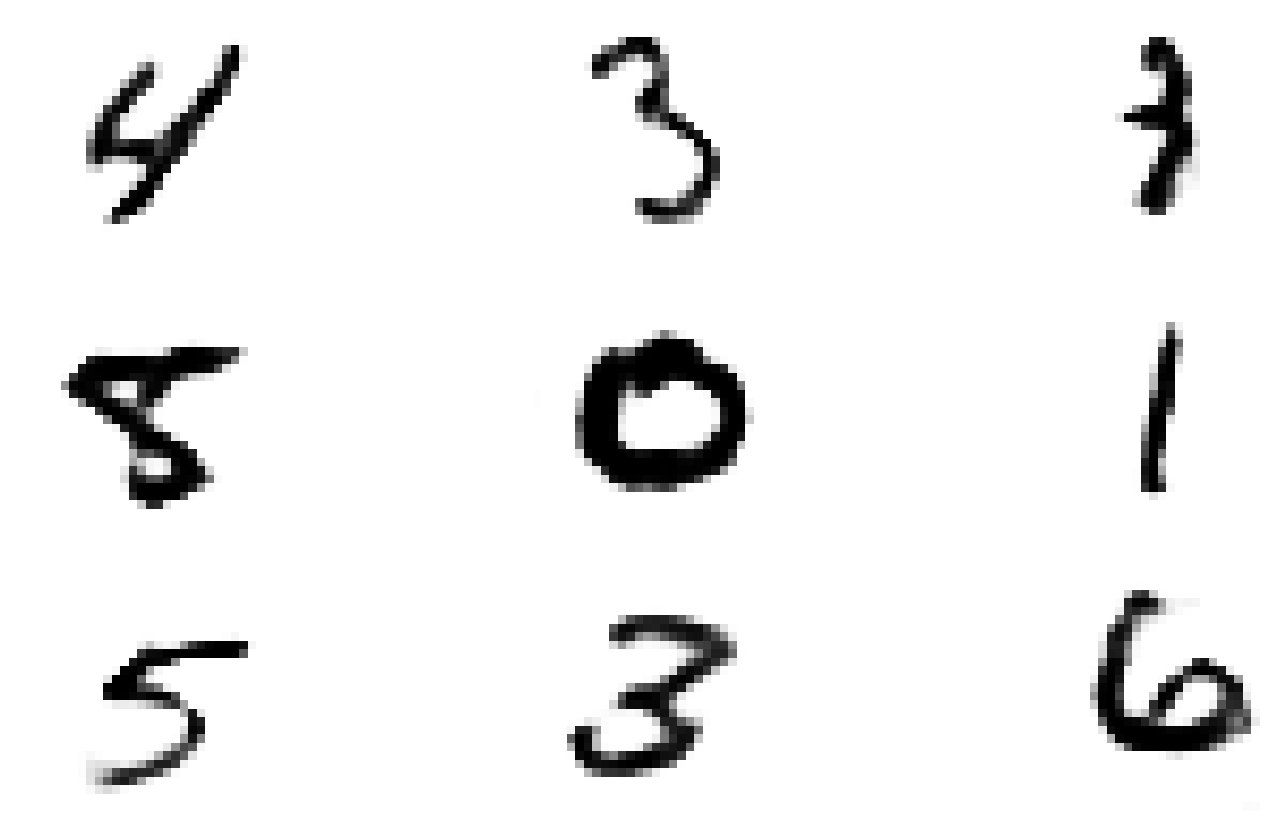}
}
{
\includegraphics[width=0.18\textwidth]{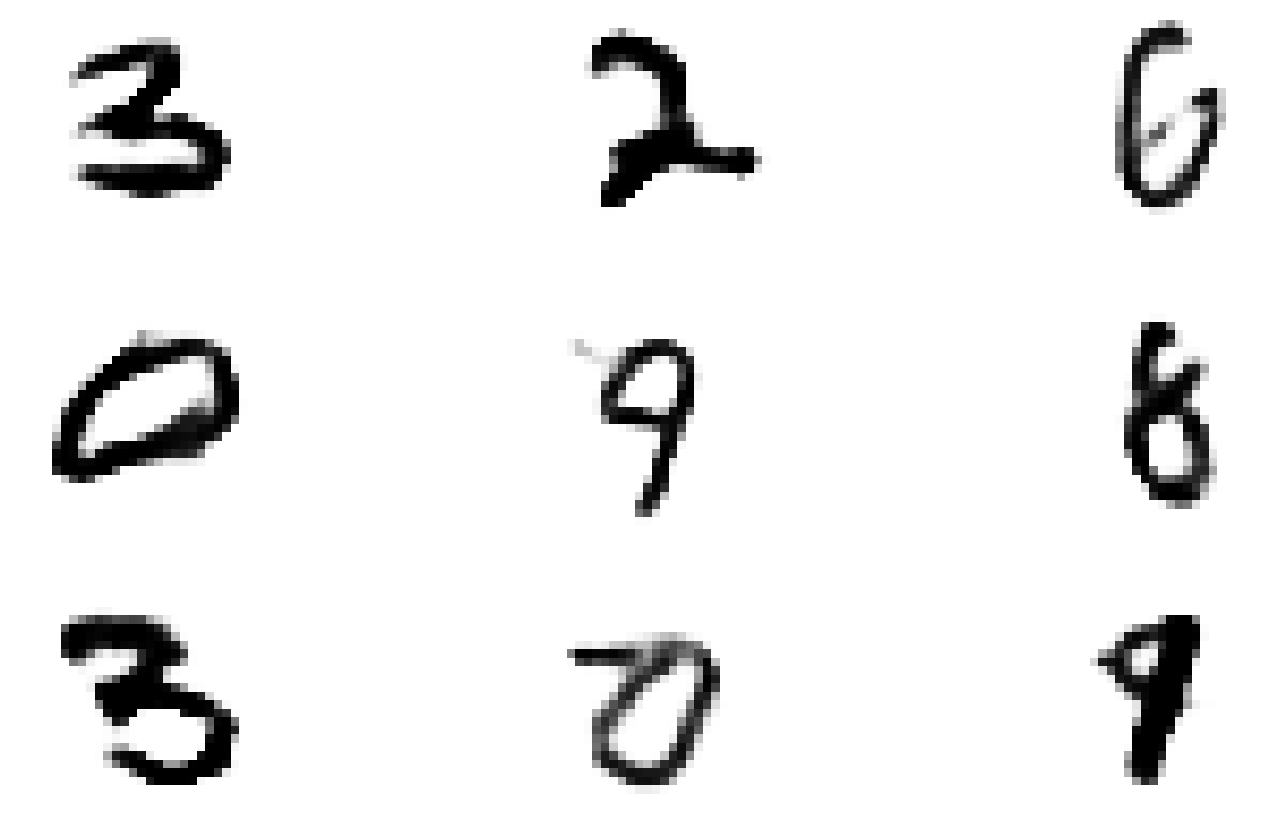}
}
{
\includegraphics[width=0.18\textwidth]{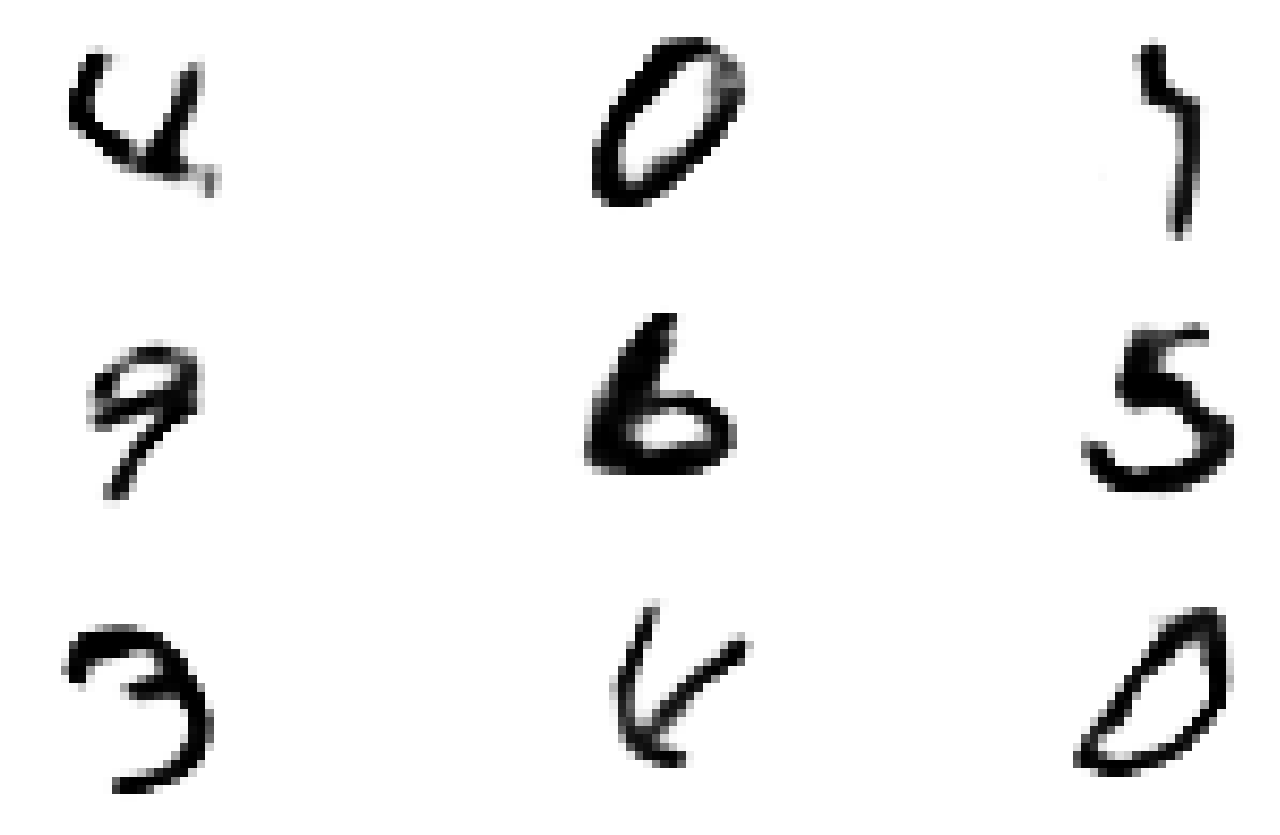}
}
\\
{
\includegraphics[width=0.18\textwidth]{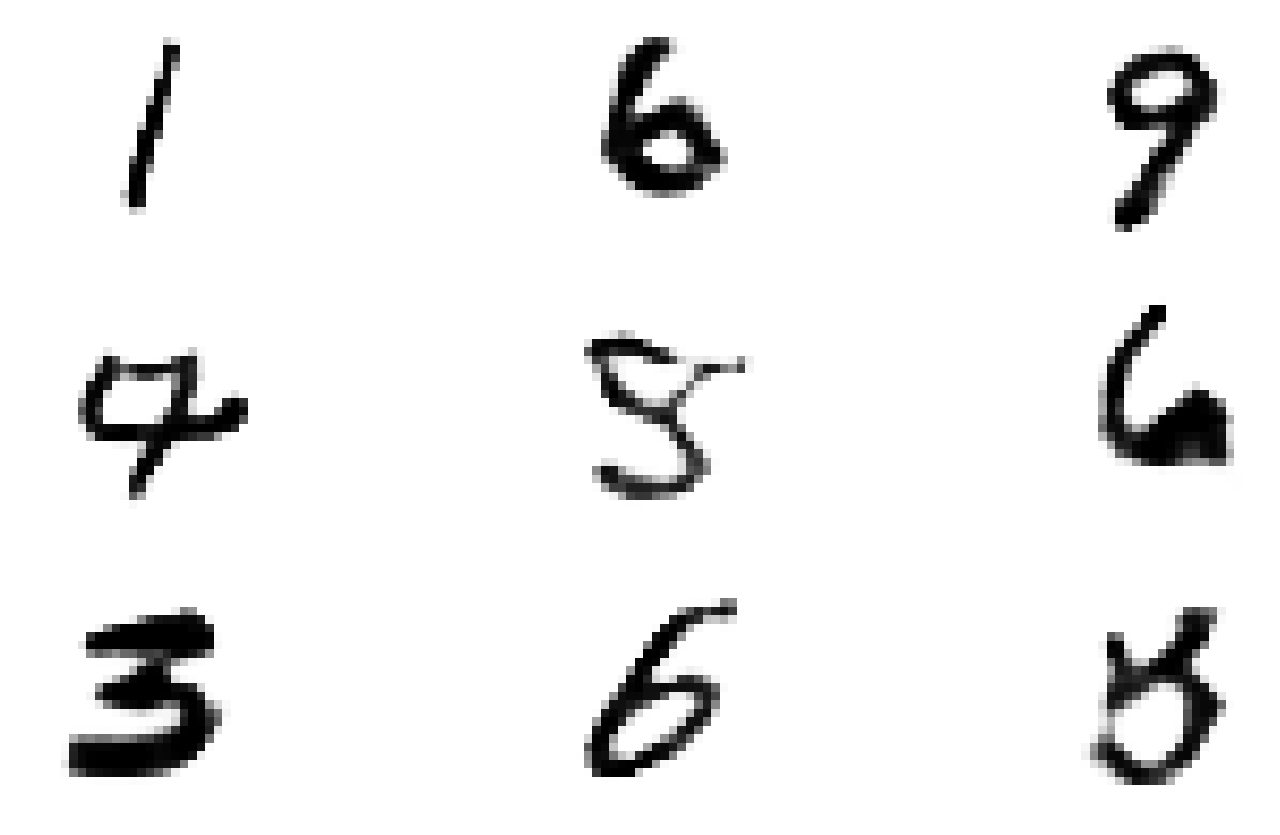}
}
{
\includegraphics[width=0.18\textwidth]{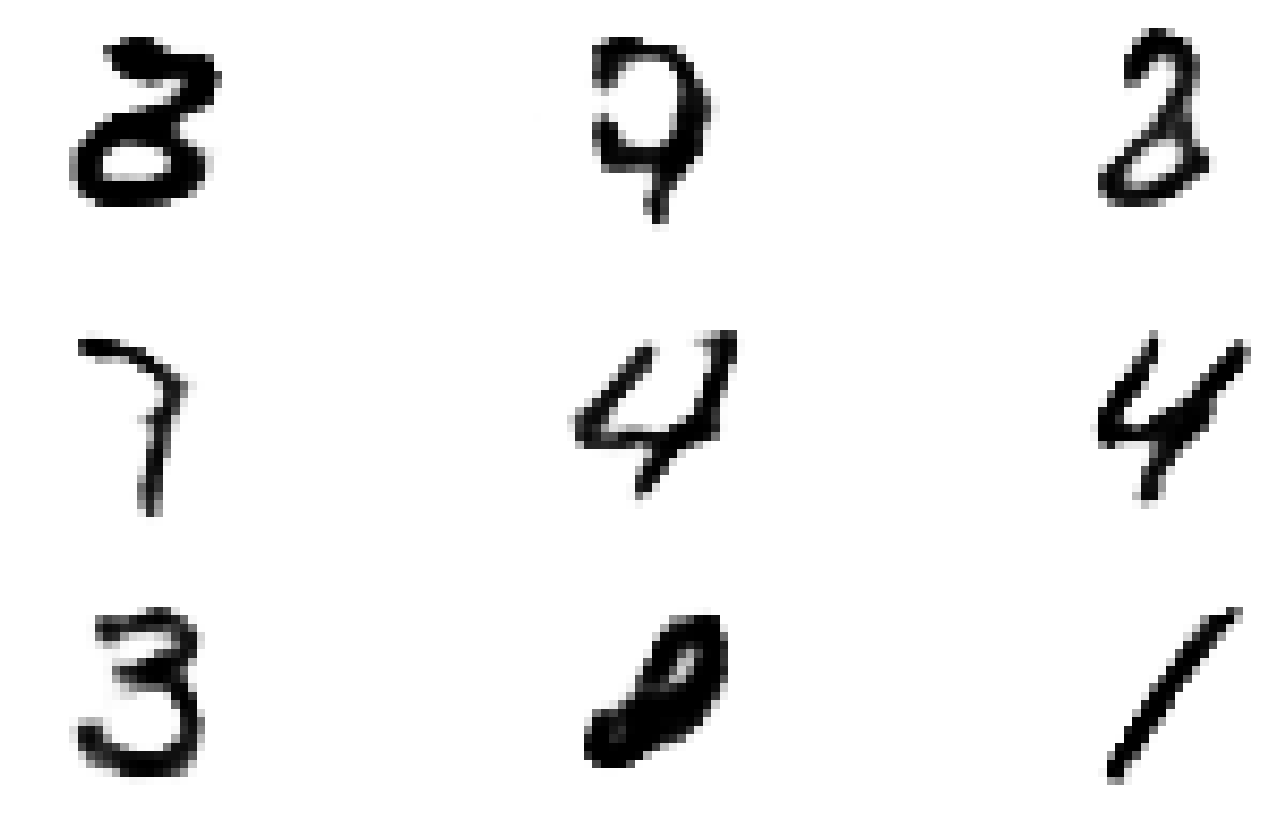}
}
{
\includegraphics[width=0.18\textwidth]{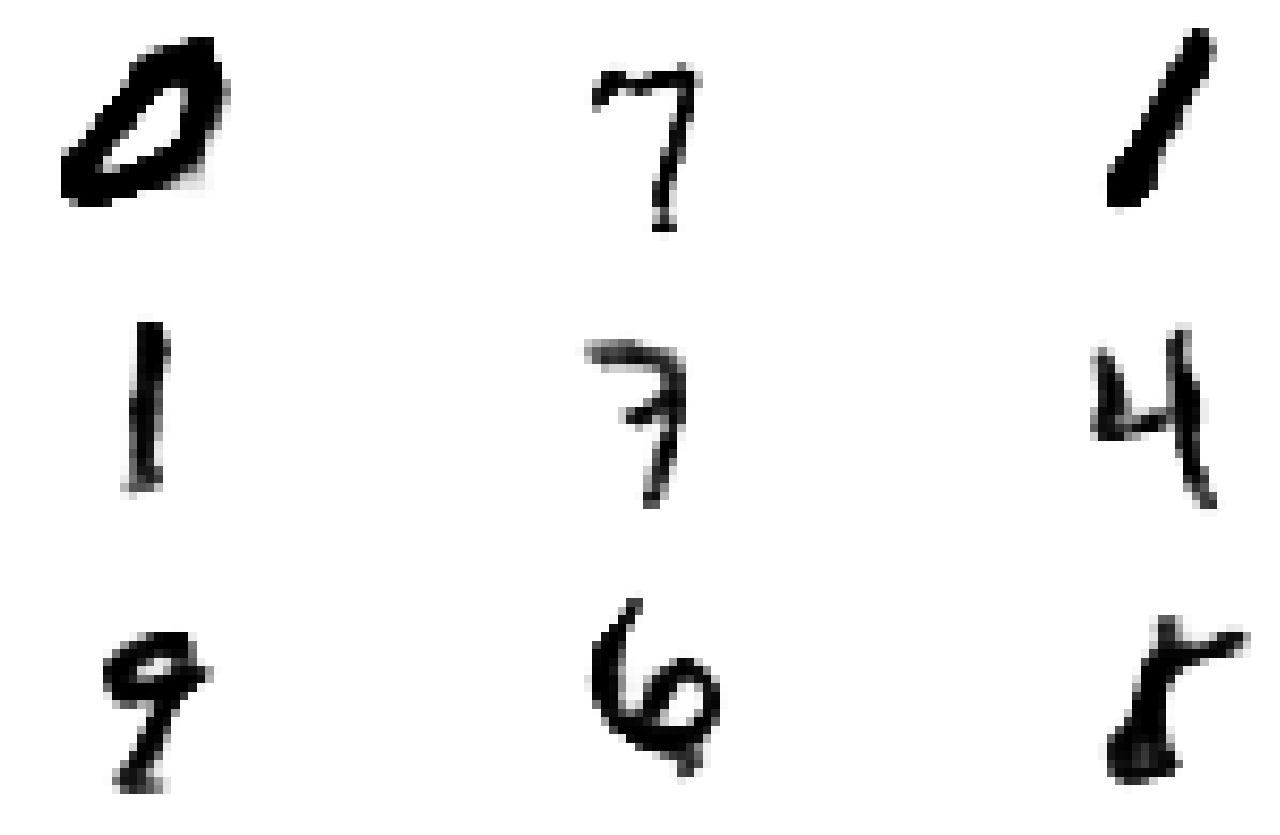}
}
{
\includegraphics[width=0.18\textwidth]{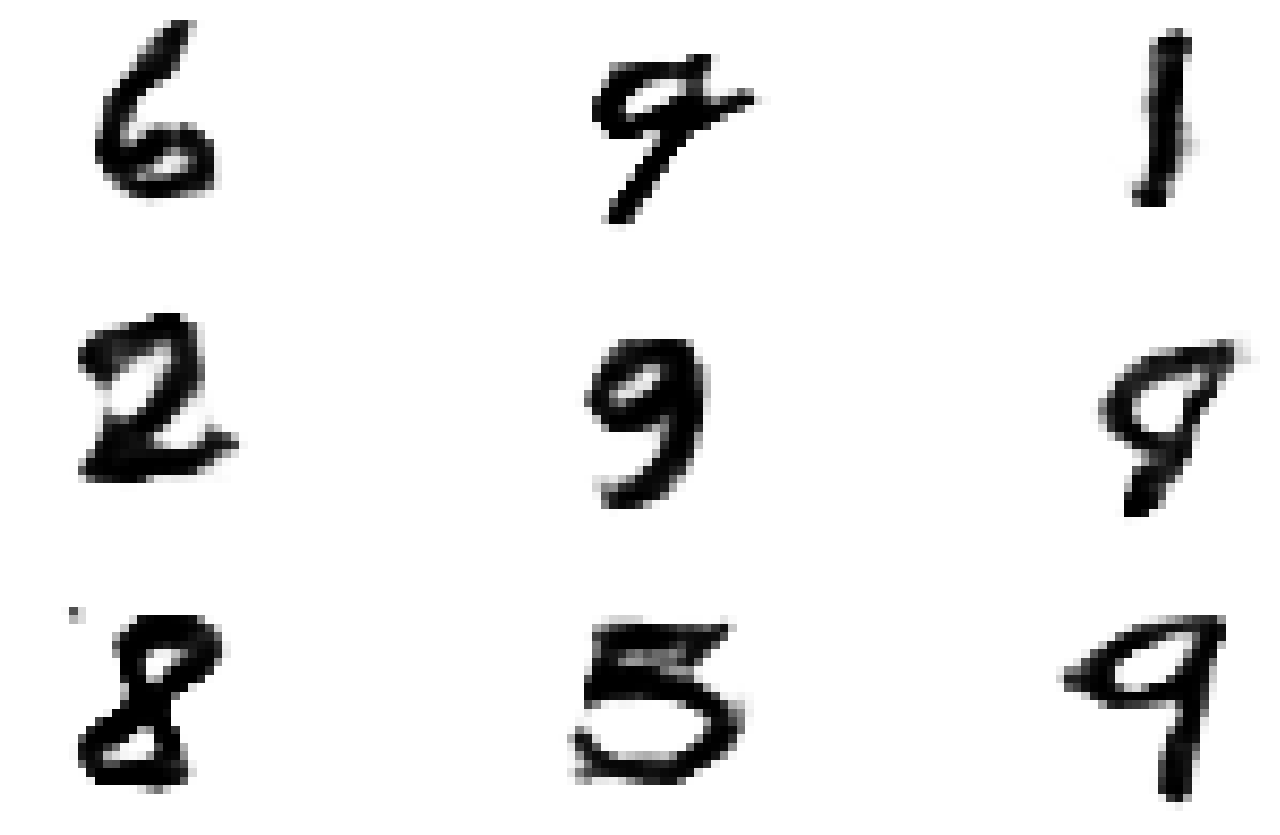}
}

\rule{0.8\textwidth}{0.3mm}\\
{
\includegraphics[width=0.18\textwidth]{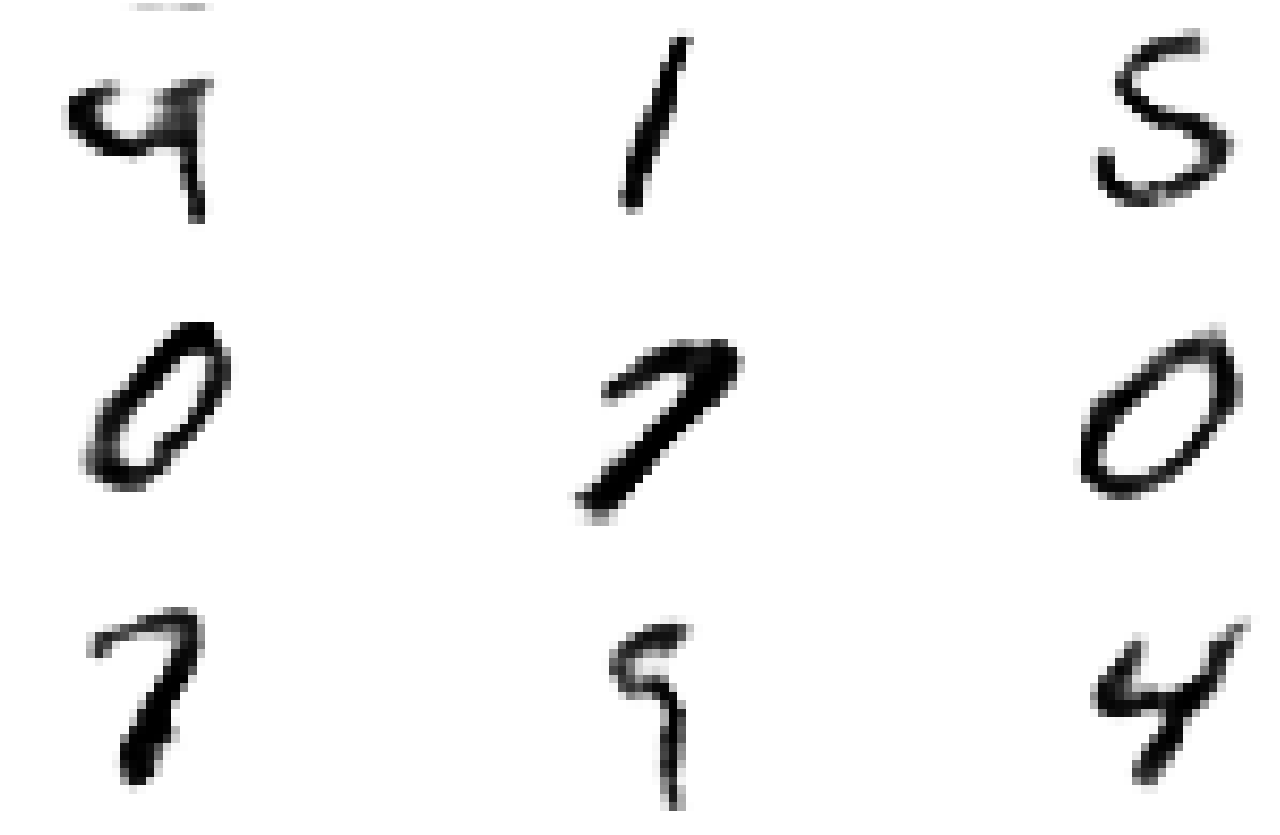}
}
{
\includegraphics[width=0.18\textwidth]{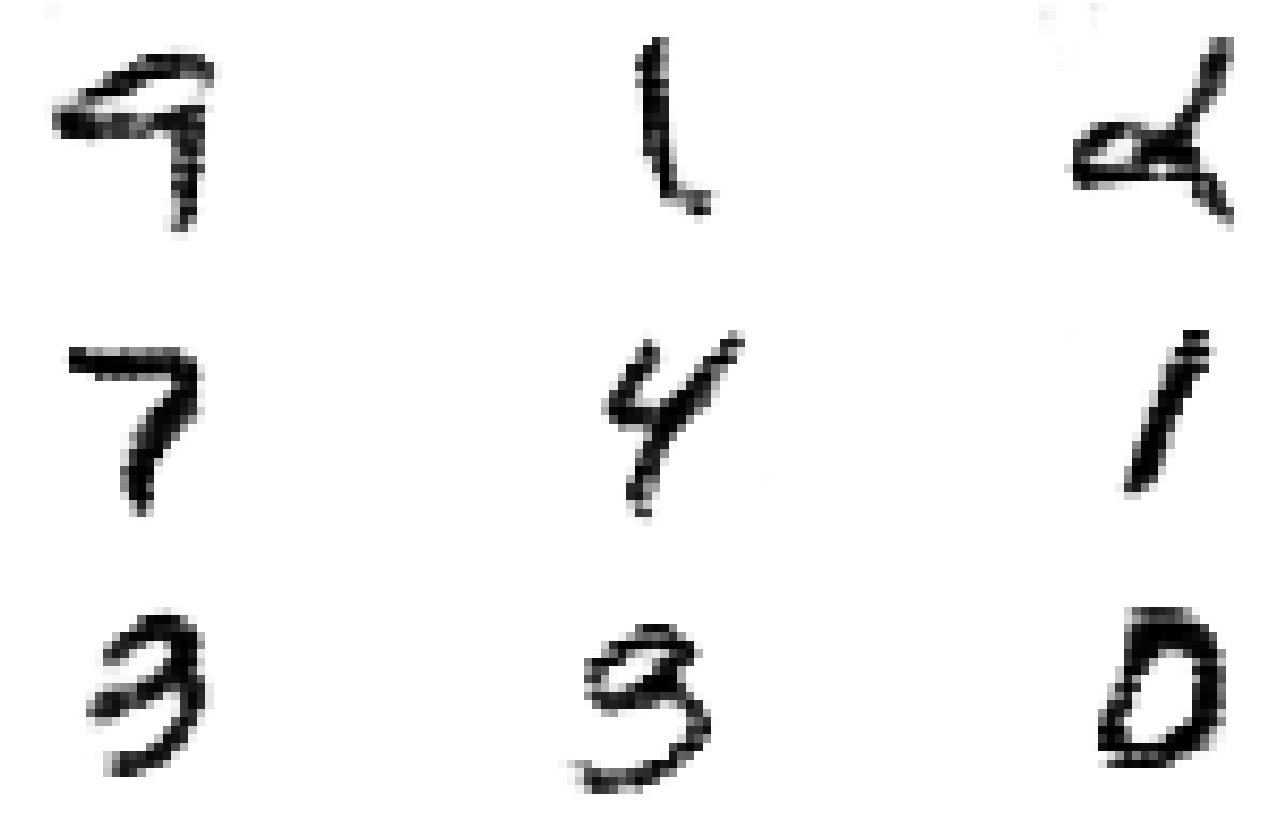}
}
{
\includegraphics[width=0.18\textwidth]{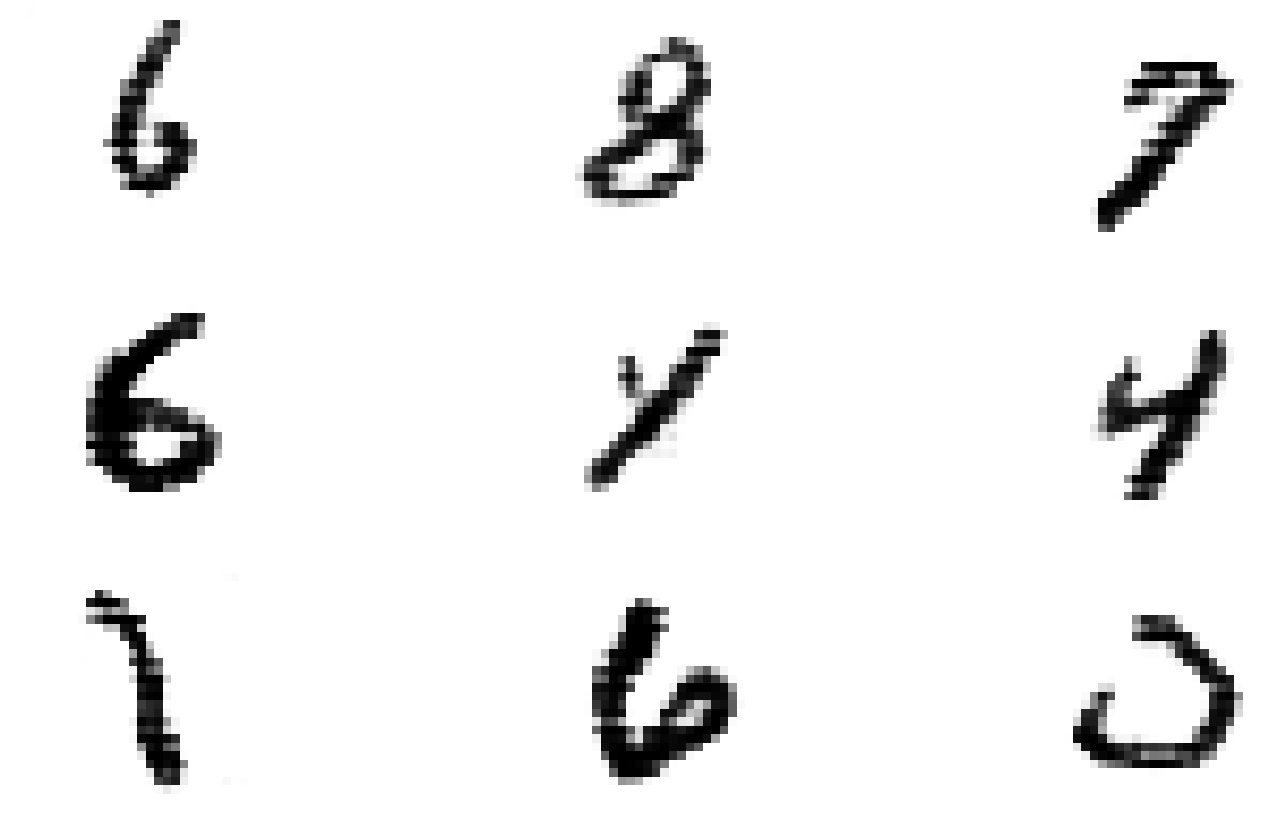}
}
{
\includegraphics[width=0.18\textwidth]{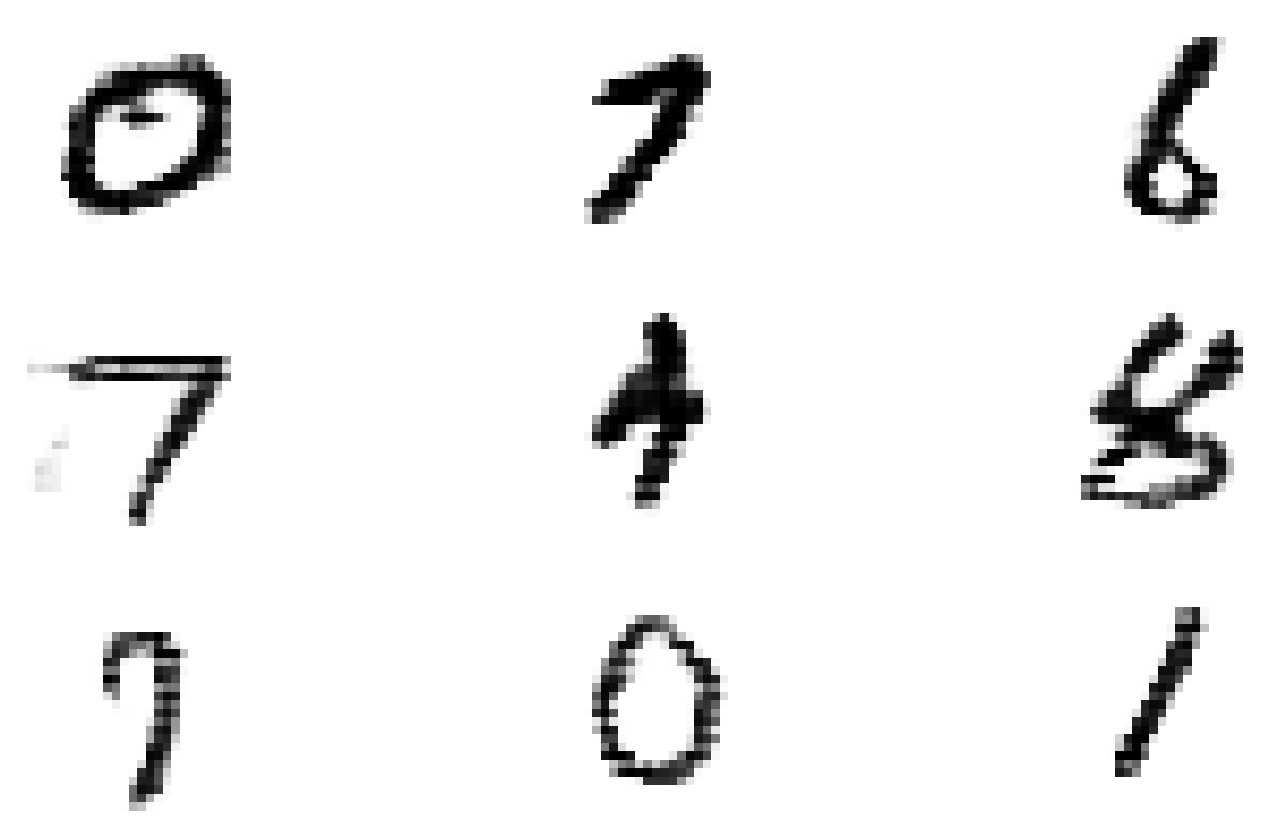}
}
\\
{
\includegraphics[width=0.18\textwidth]{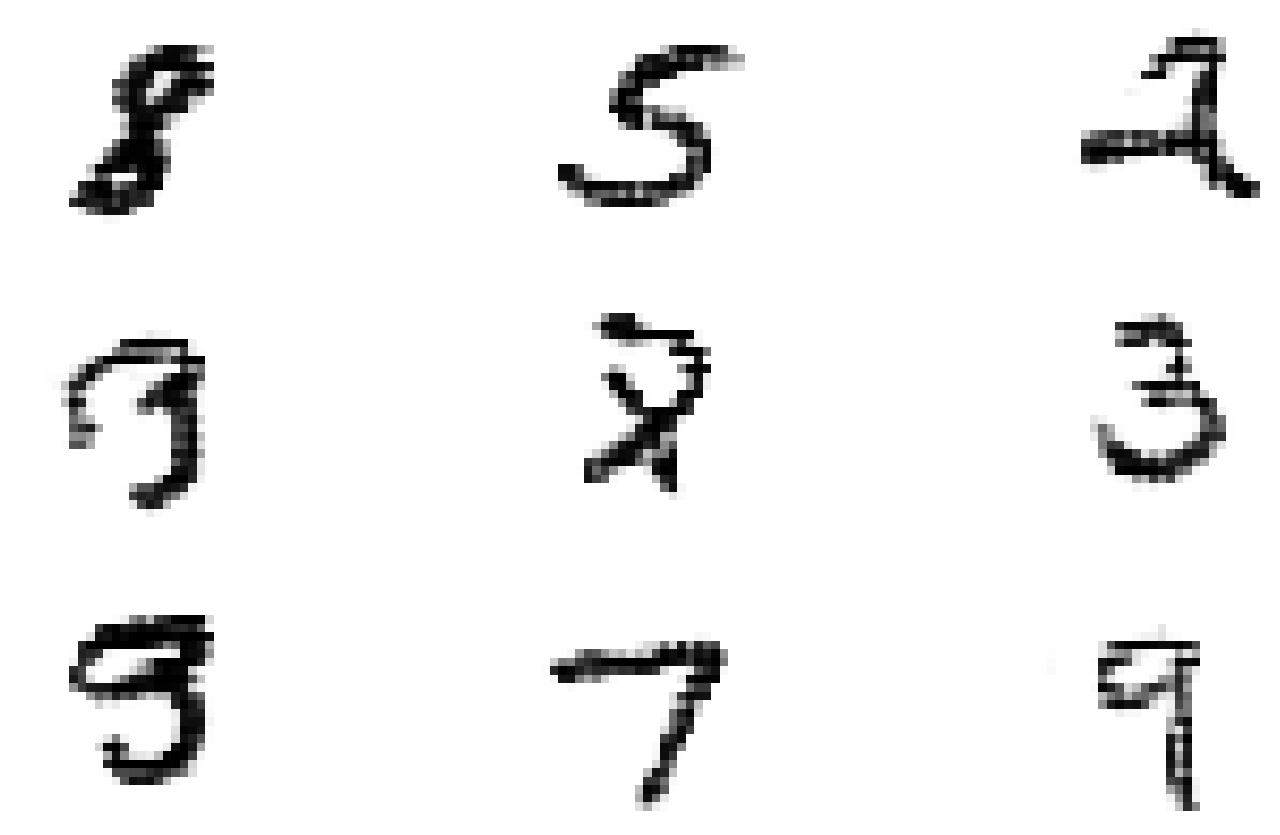}
}
{
\includegraphics[width=0.18\textwidth]{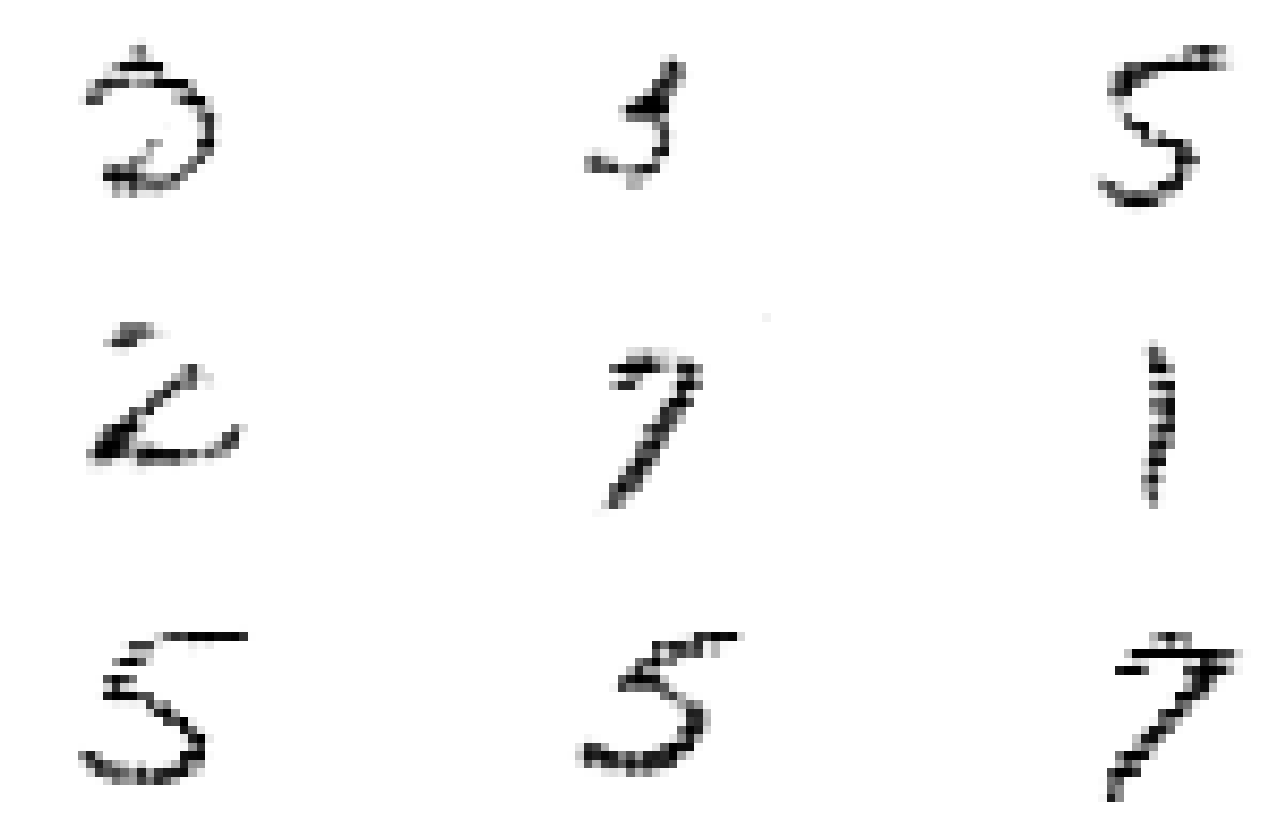}
}
{
\includegraphics[width=0.18\textwidth]{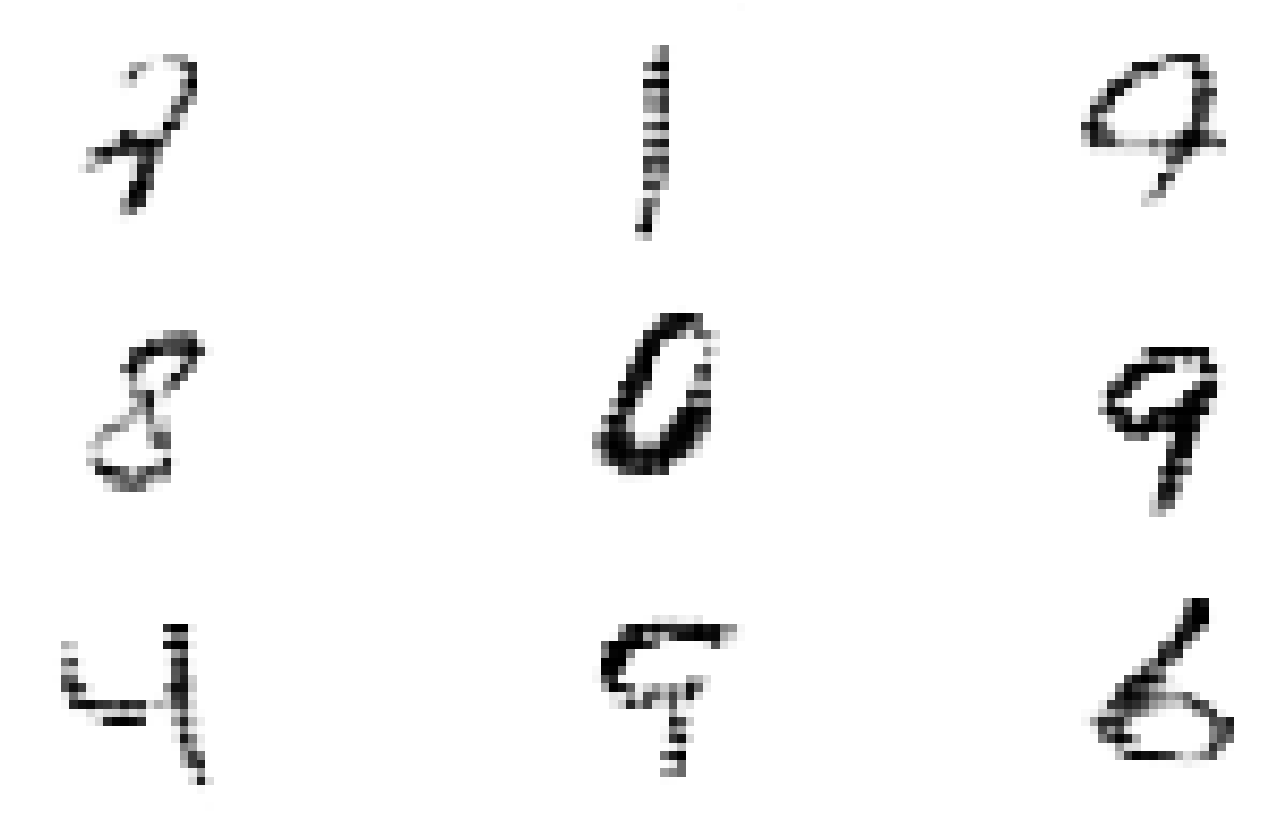}
}
{
\includegraphics[width=0.18\textwidth]{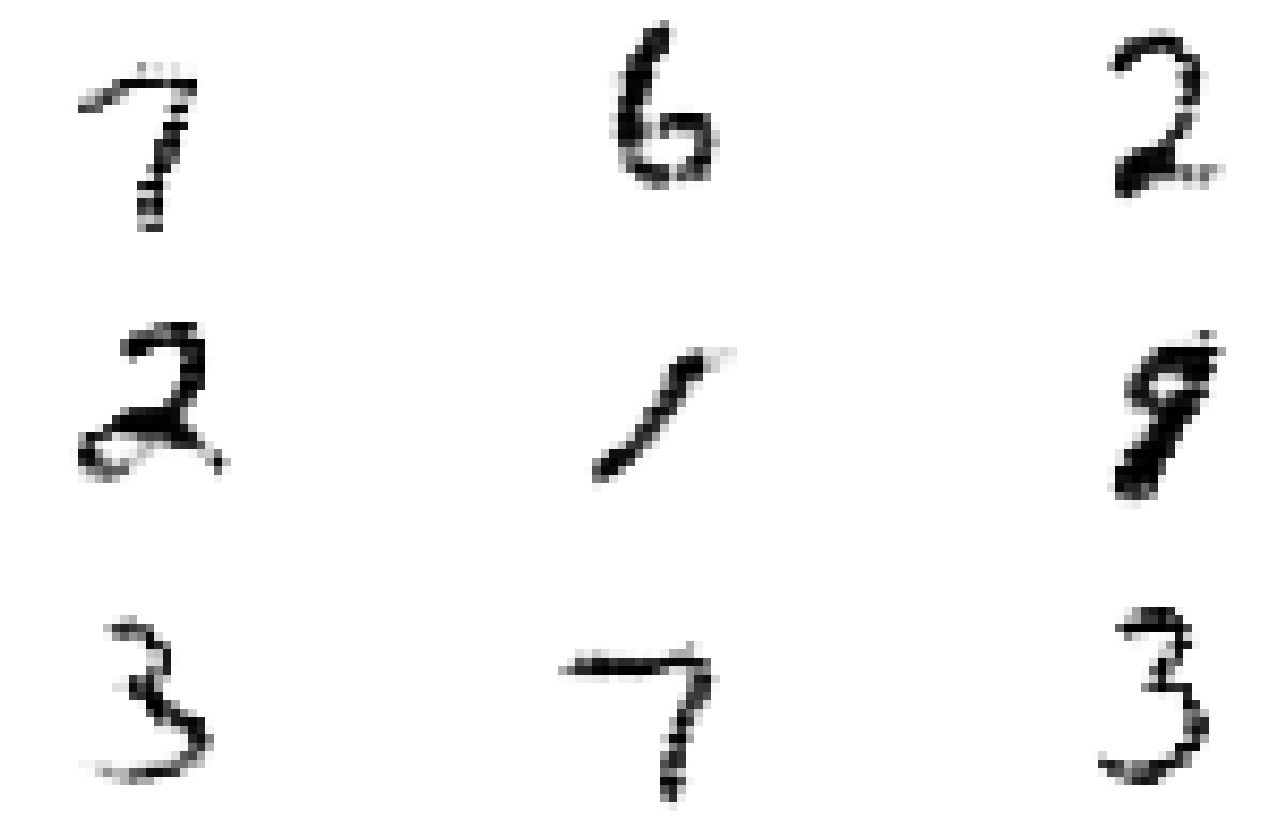}
}

\caption{Blurred and recovered images. The image in a box refers to the blurred image with convolution kernel $K_{blur}$. The rest images, divided into 3 parts by two lines, refer to the recovered images generated by the models with convolution kernels $K_{rec}^1\ (top),\ K_{rec}^2\ (medium),\ K_{rec}^3\ (bottom)$. The models are trained with 10000 images from the MNIST dataset.}
\label{recovery}
\end{figure}

\begin{figure}[!h]
\centering
\fbox{
\includegraphics[width=0.18\textwidth]{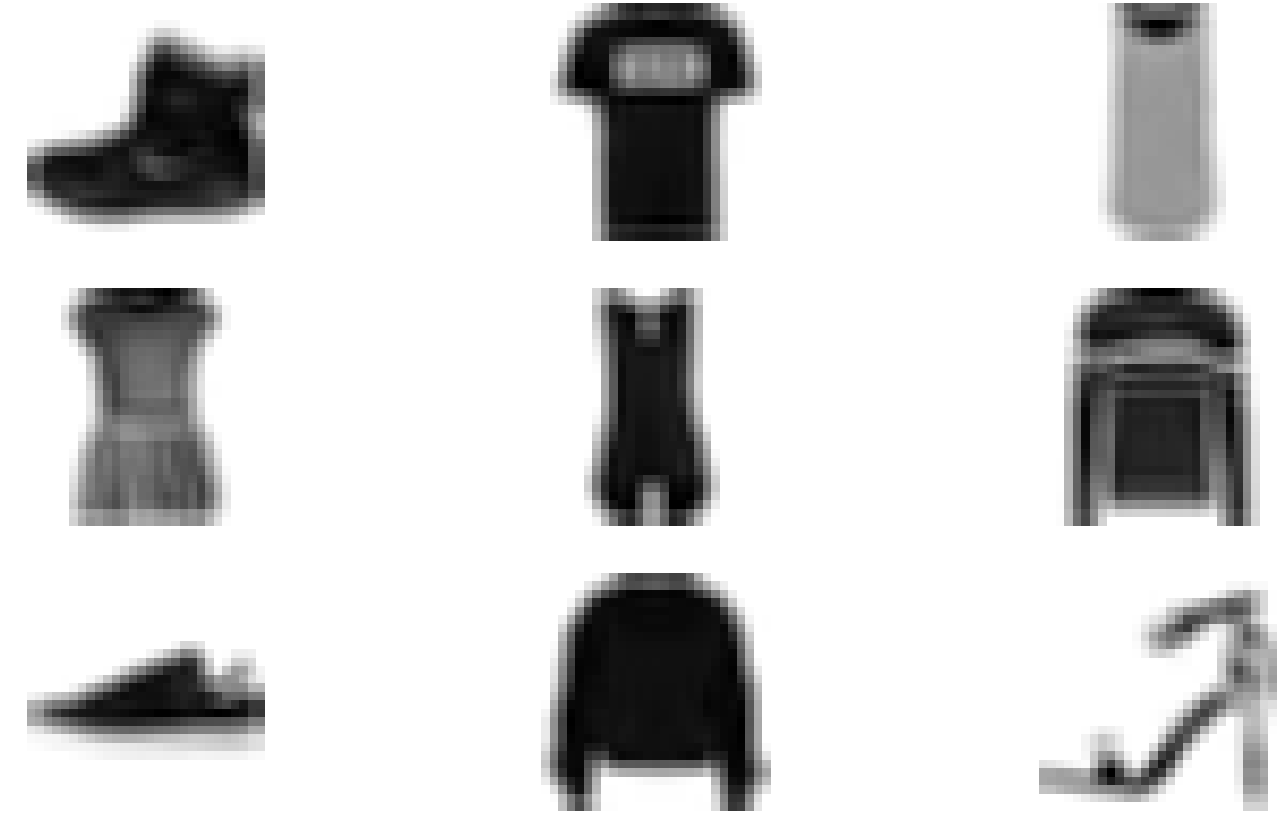}
}
{
\includegraphics[width=0.18\textwidth]{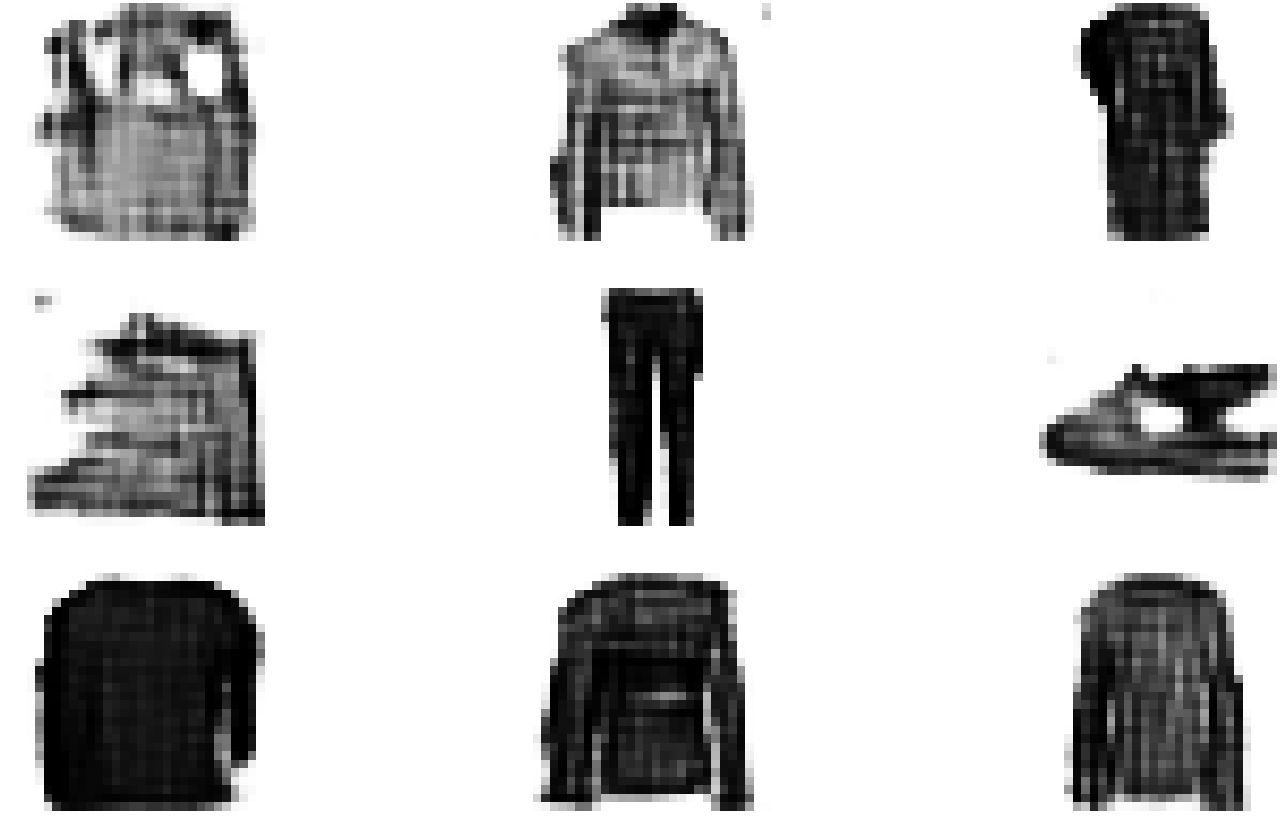}
}
{
\includegraphics[width=0.18\textwidth]{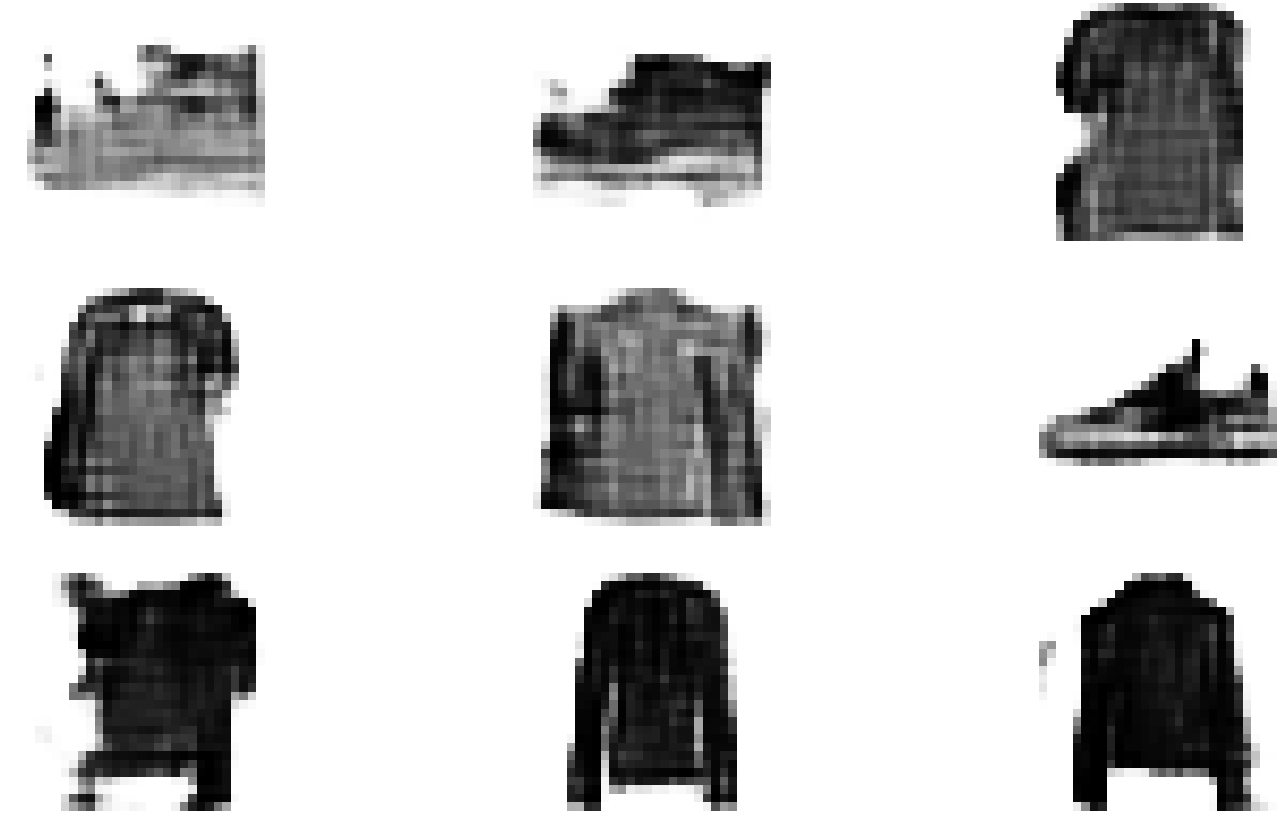}
}
{
\includegraphics[width=0.18\textwidth]{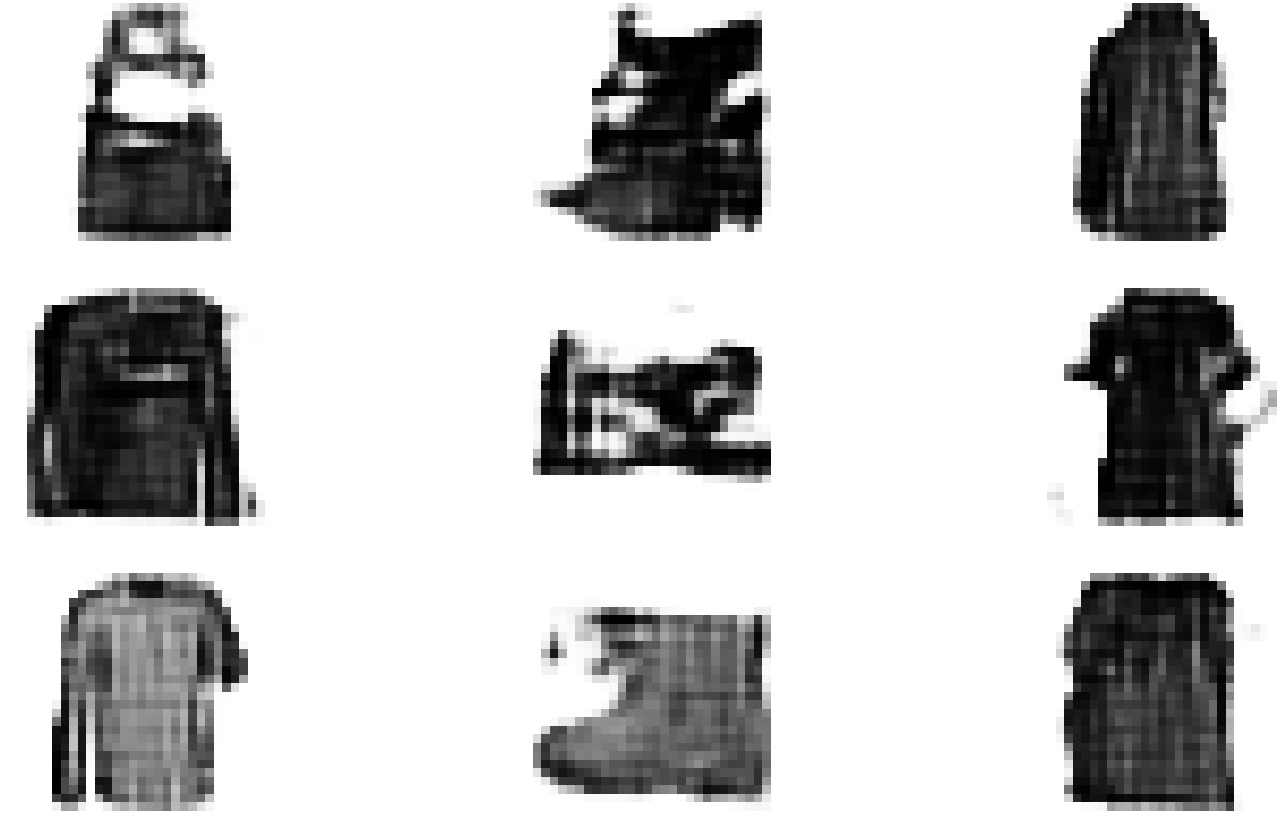}
}
\\
{
\includegraphics[width=0.18\textwidth]{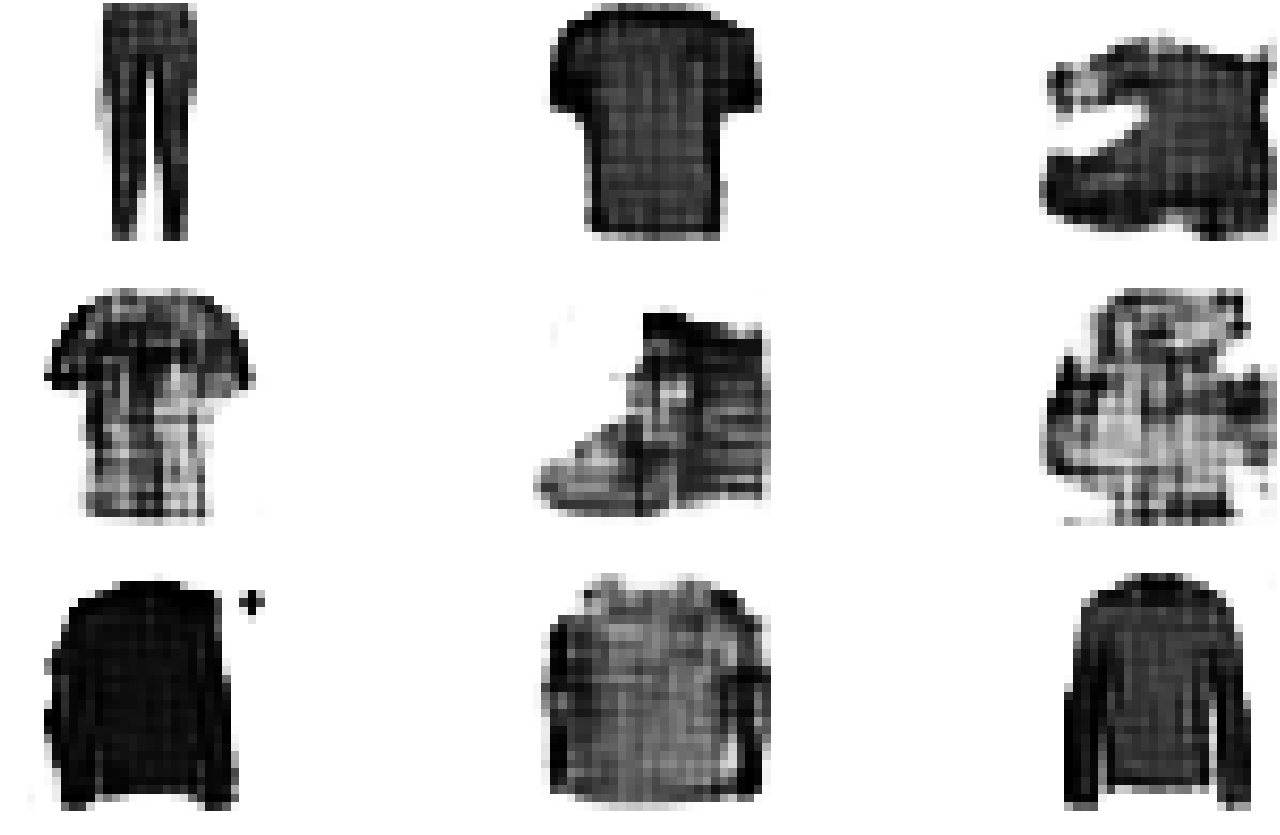}
}
{
\includegraphics[width=0.18\textwidth]{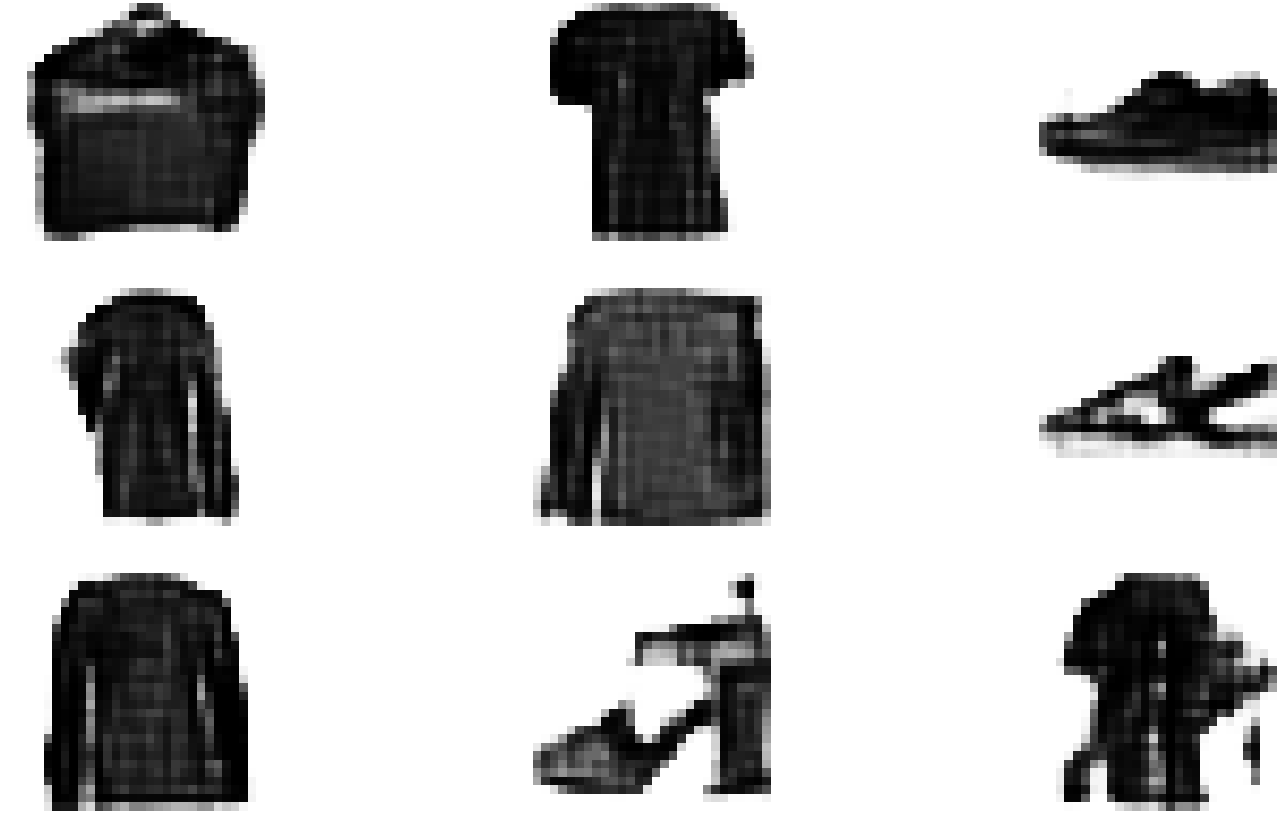}
}
{
\includegraphics[width=0.18\textwidth]{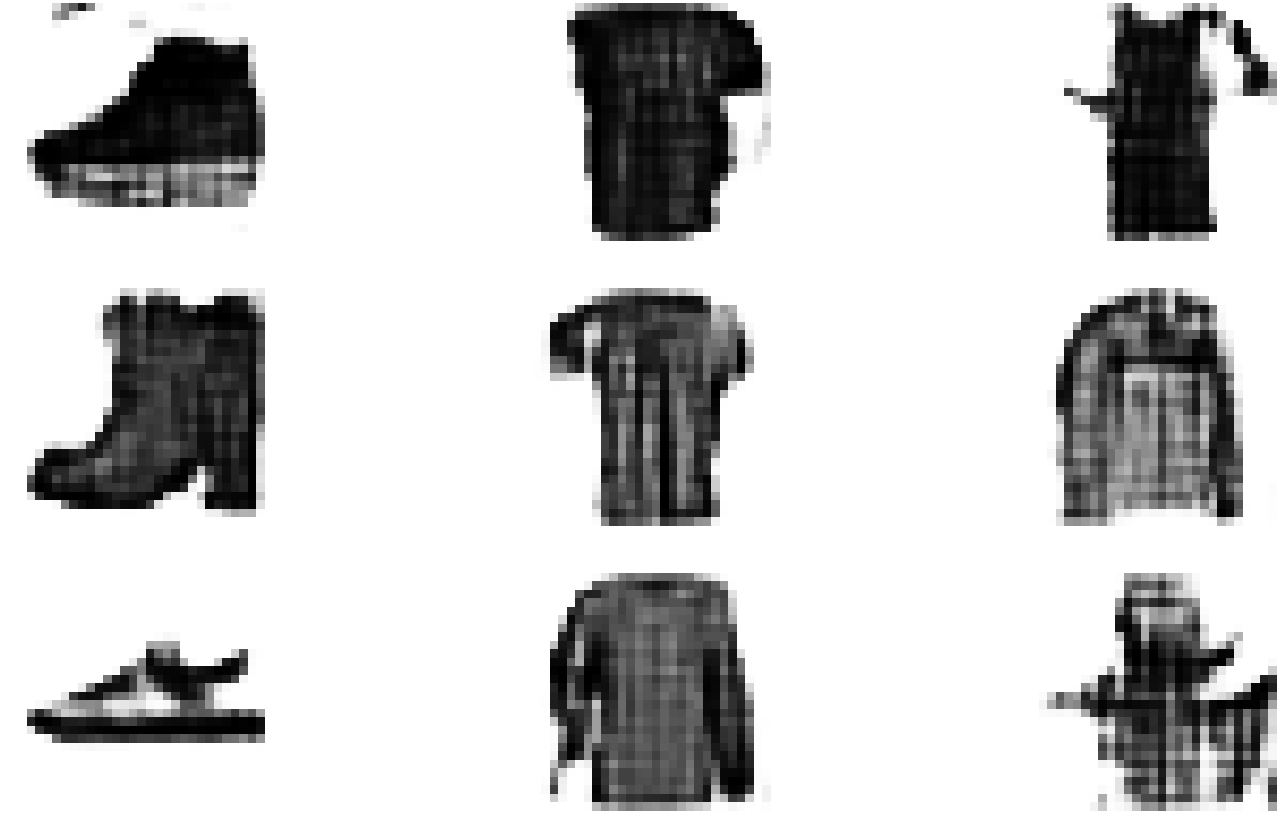}
}
{
\includegraphics[width=0.18\textwidth]{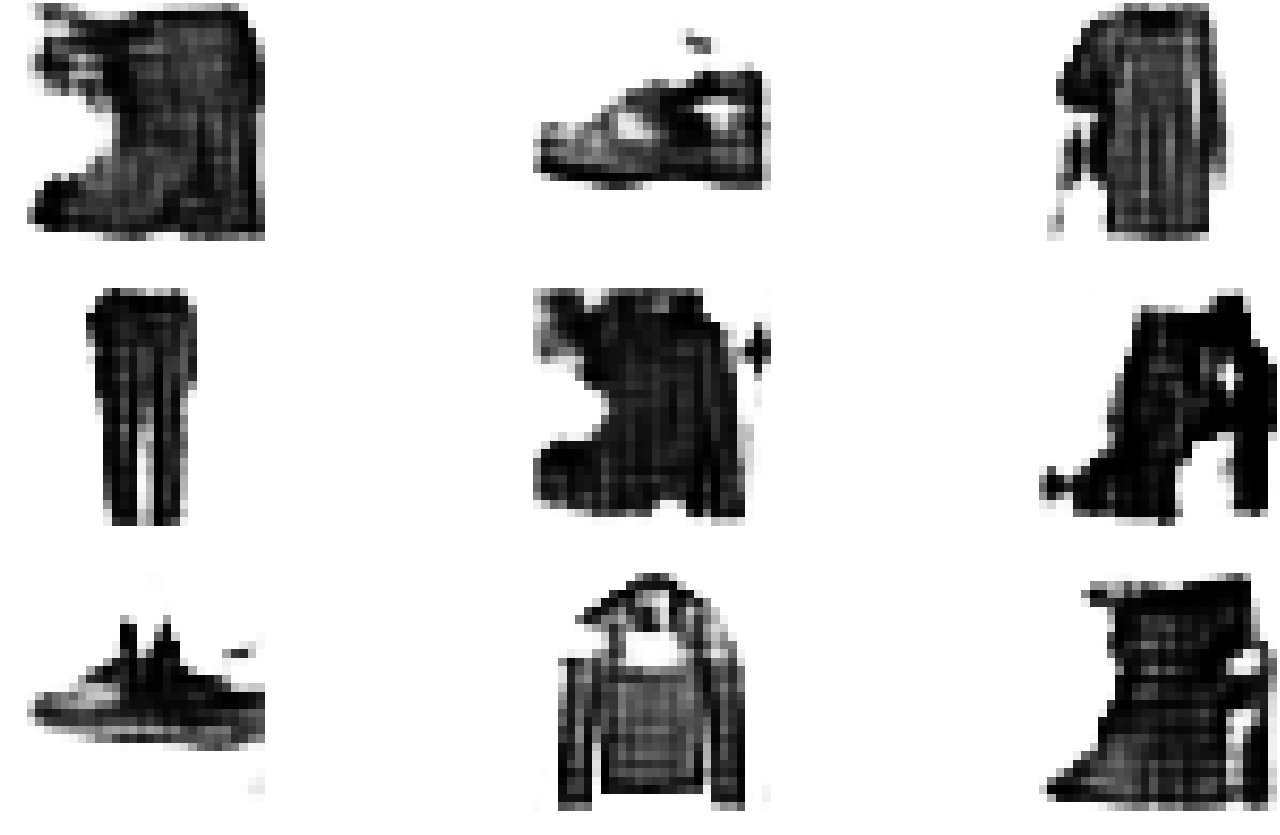}
}

\rule{0.8\textwidth}{0.3mm}\\
{
\includegraphics[width=0.18\textwidth]{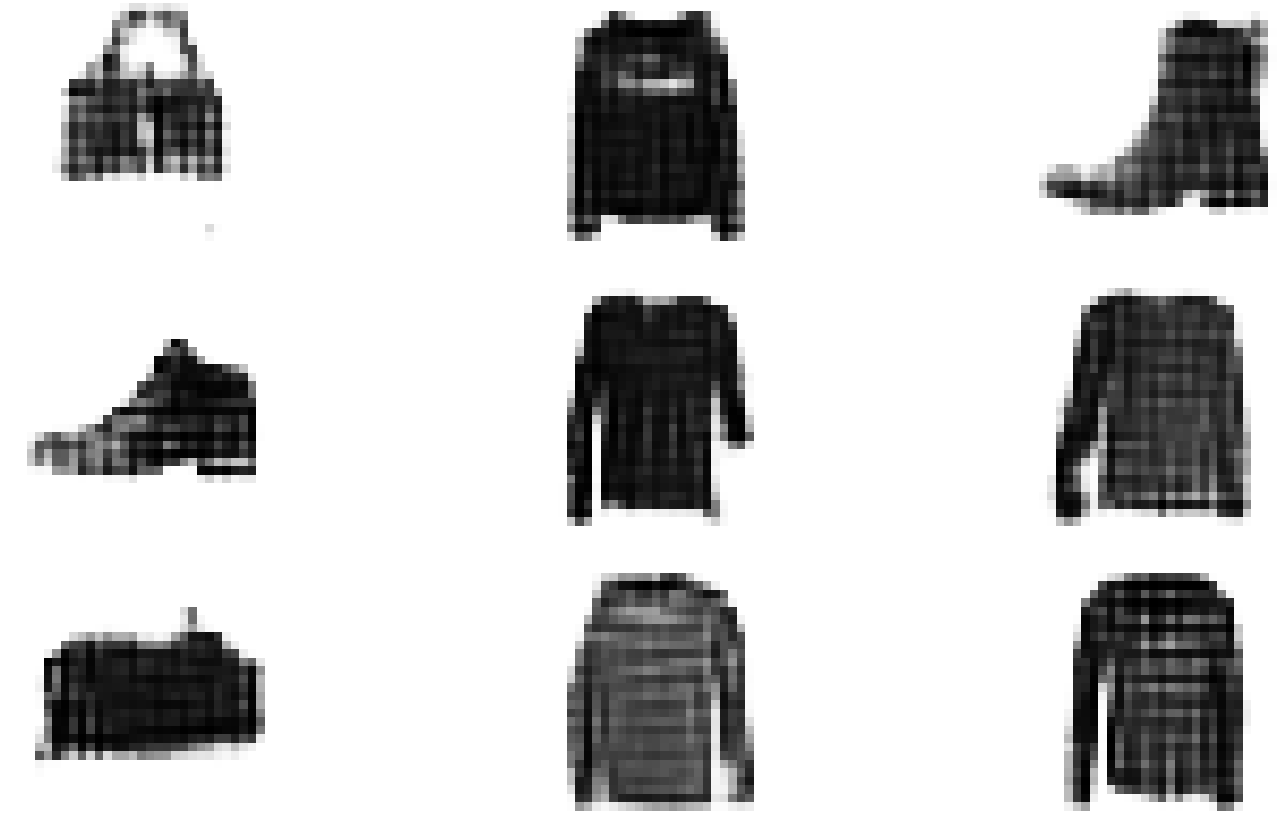}
}
{
\includegraphics[width=0.18\textwidth]{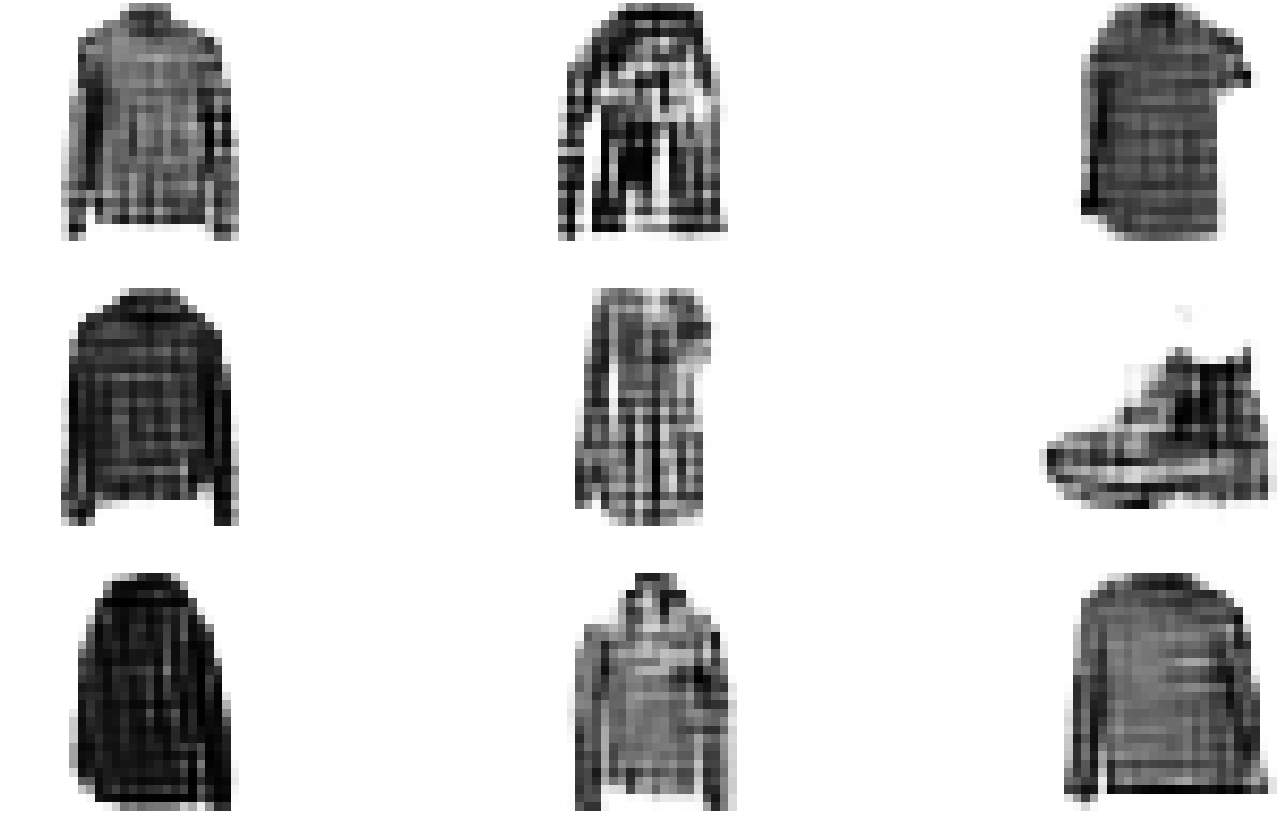}
}
{
\includegraphics[width=0.18\textwidth]{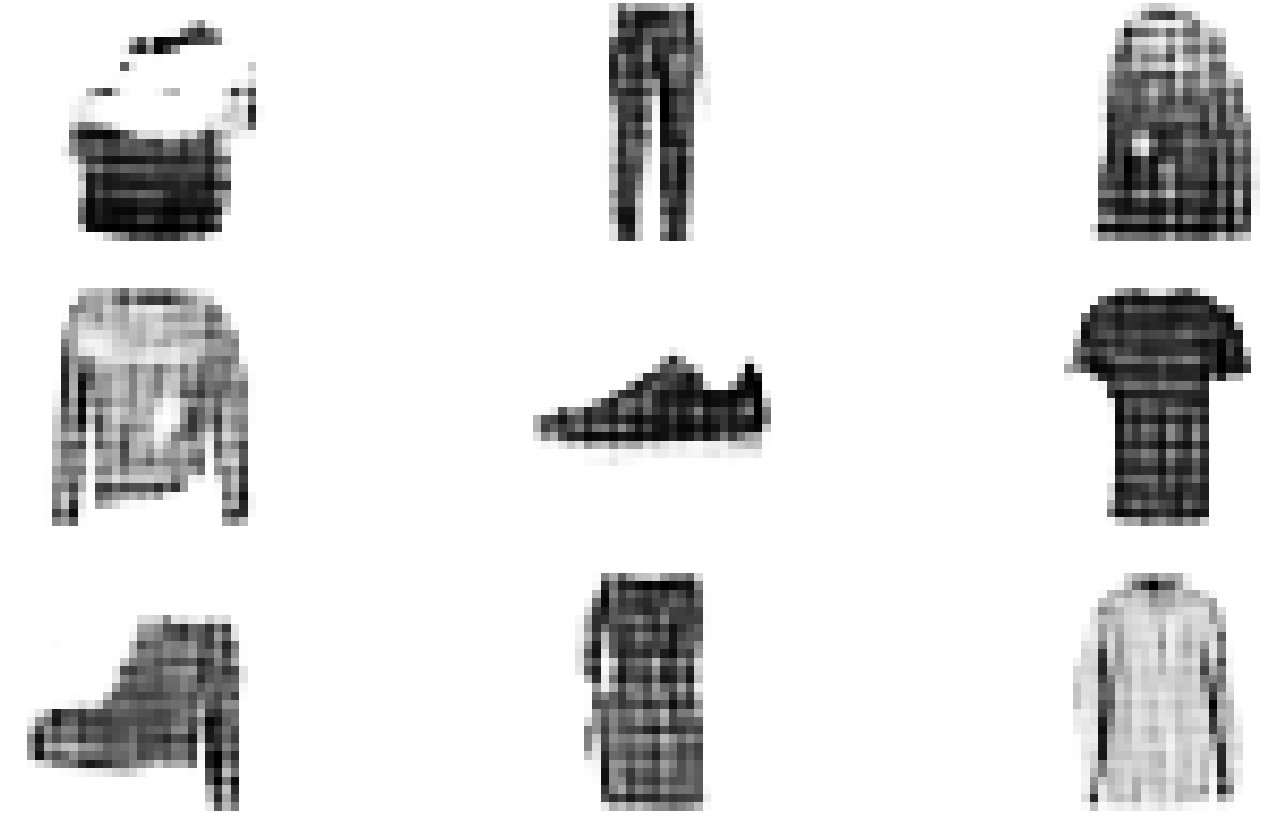}
}
{
\includegraphics[width=0.18\textwidth]{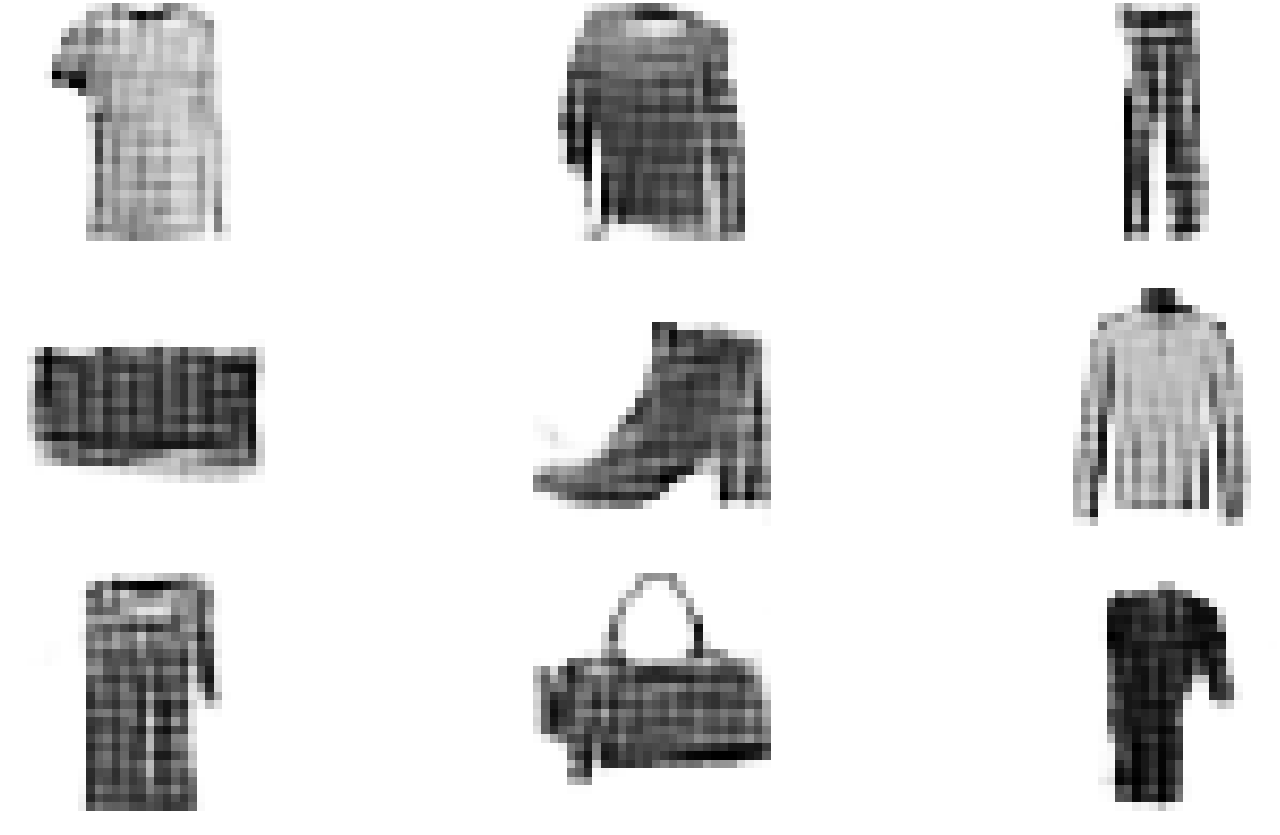}
}
\\
{
\includegraphics[width=0.18\textwidth]{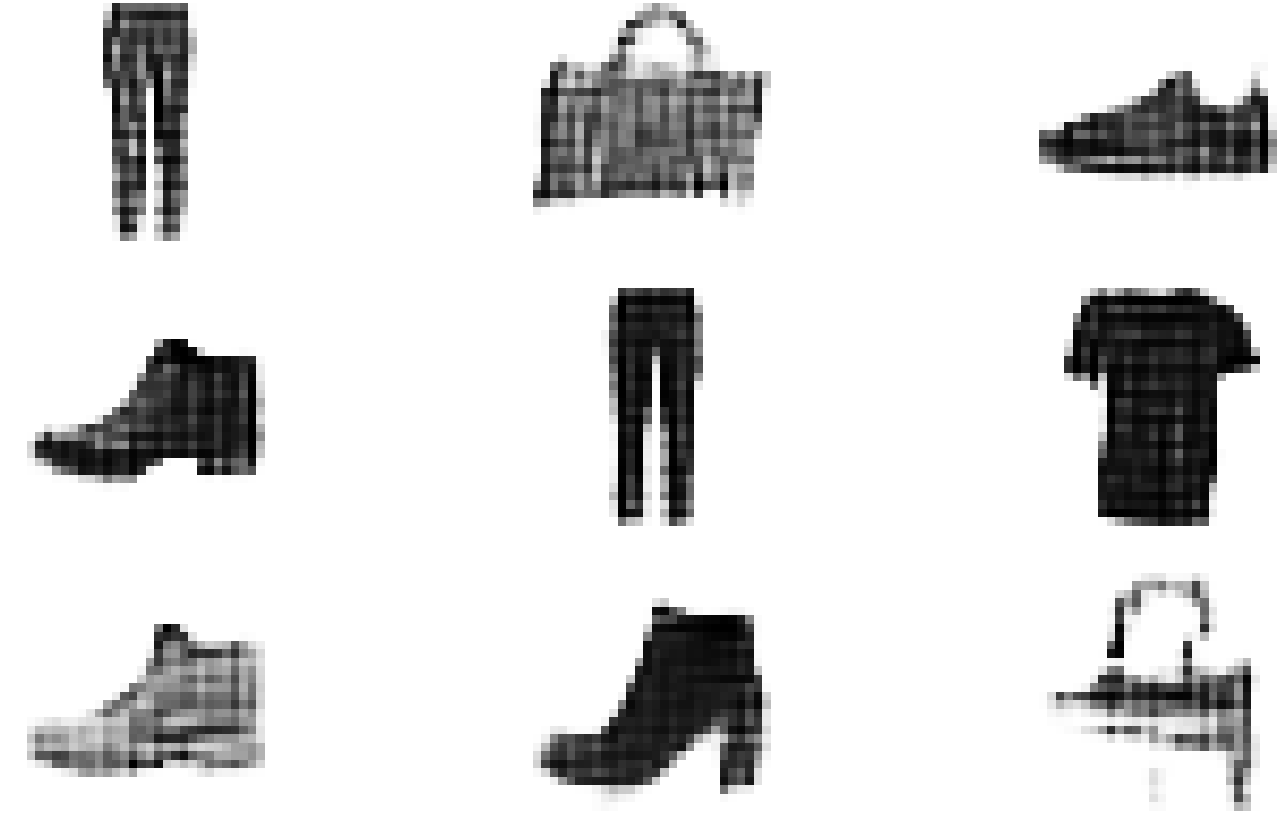}
}
{
\includegraphics[width=0.18\textwidth]{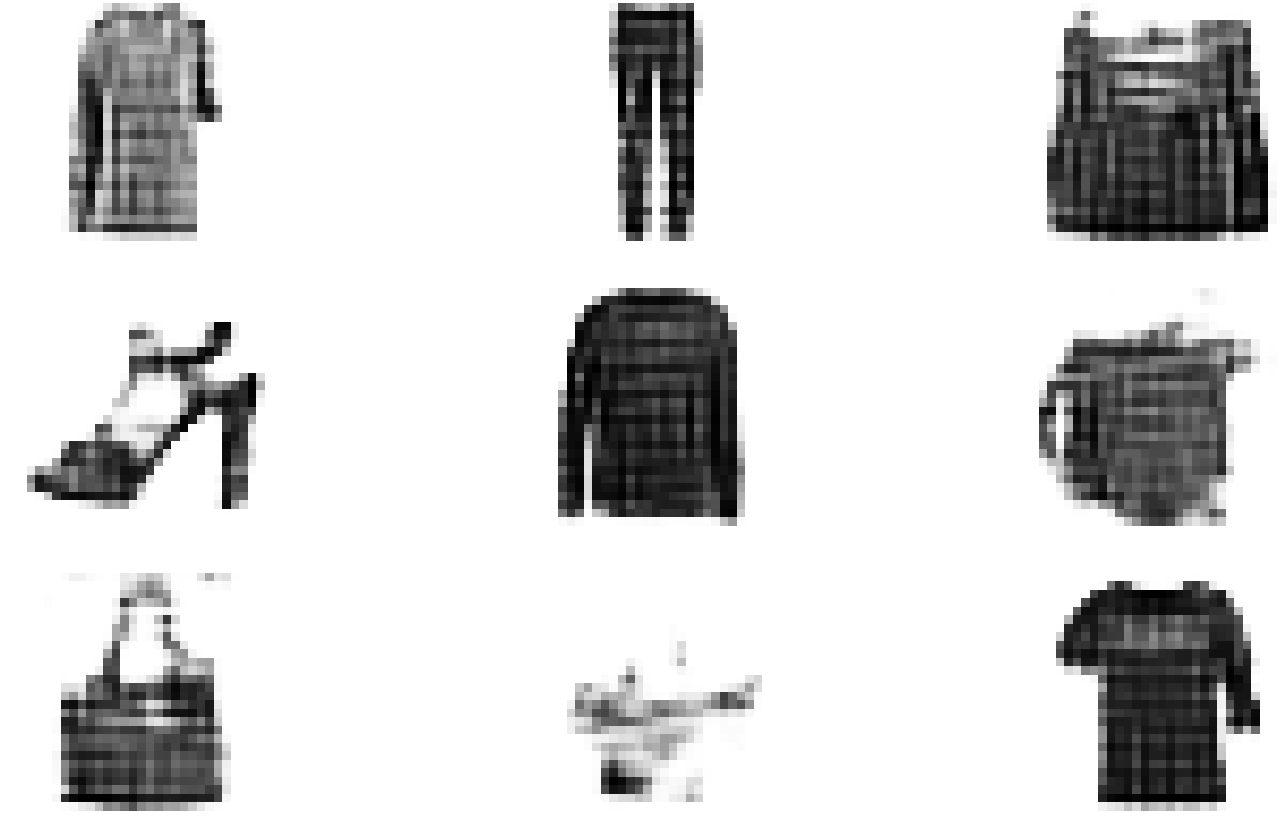}
}
{
\includegraphics[width=0.18\textwidth]{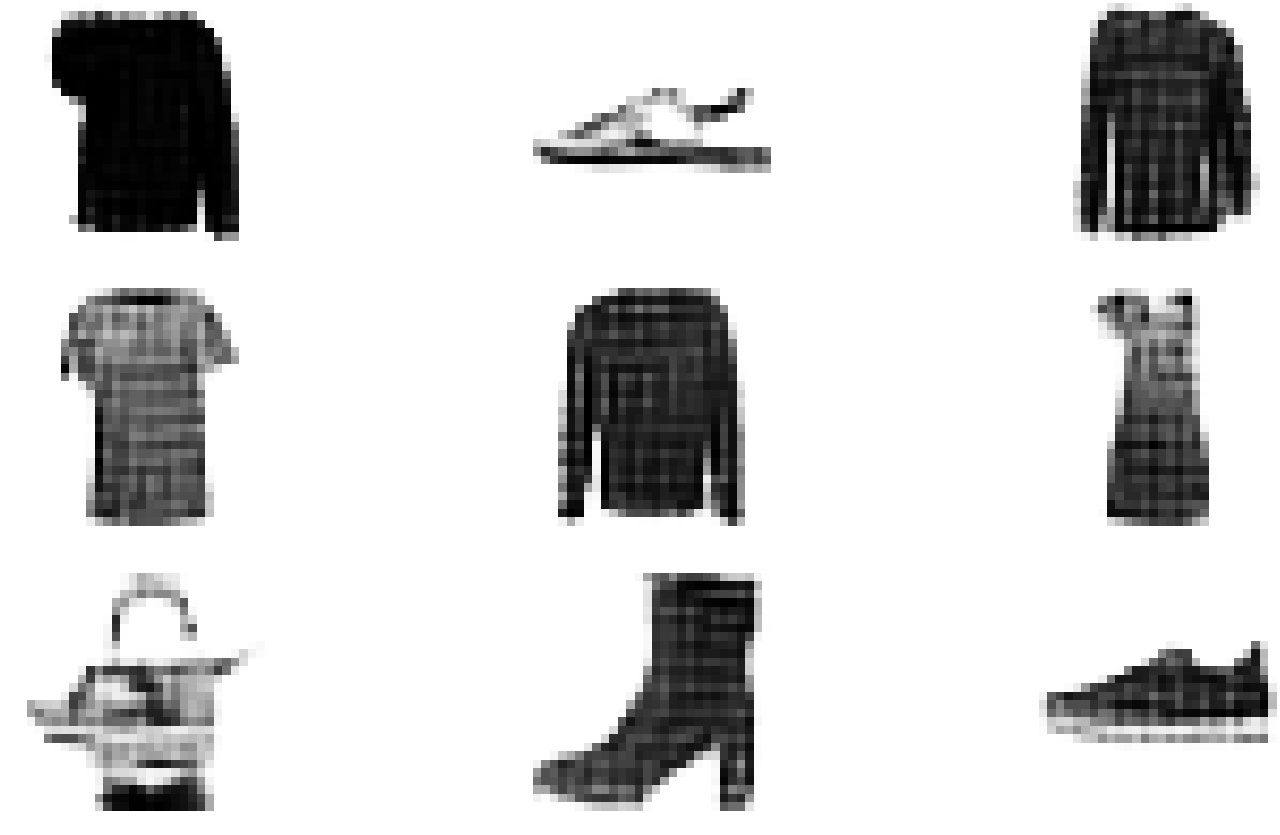}
}
{
\includegraphics[width=0.18\textwidth]{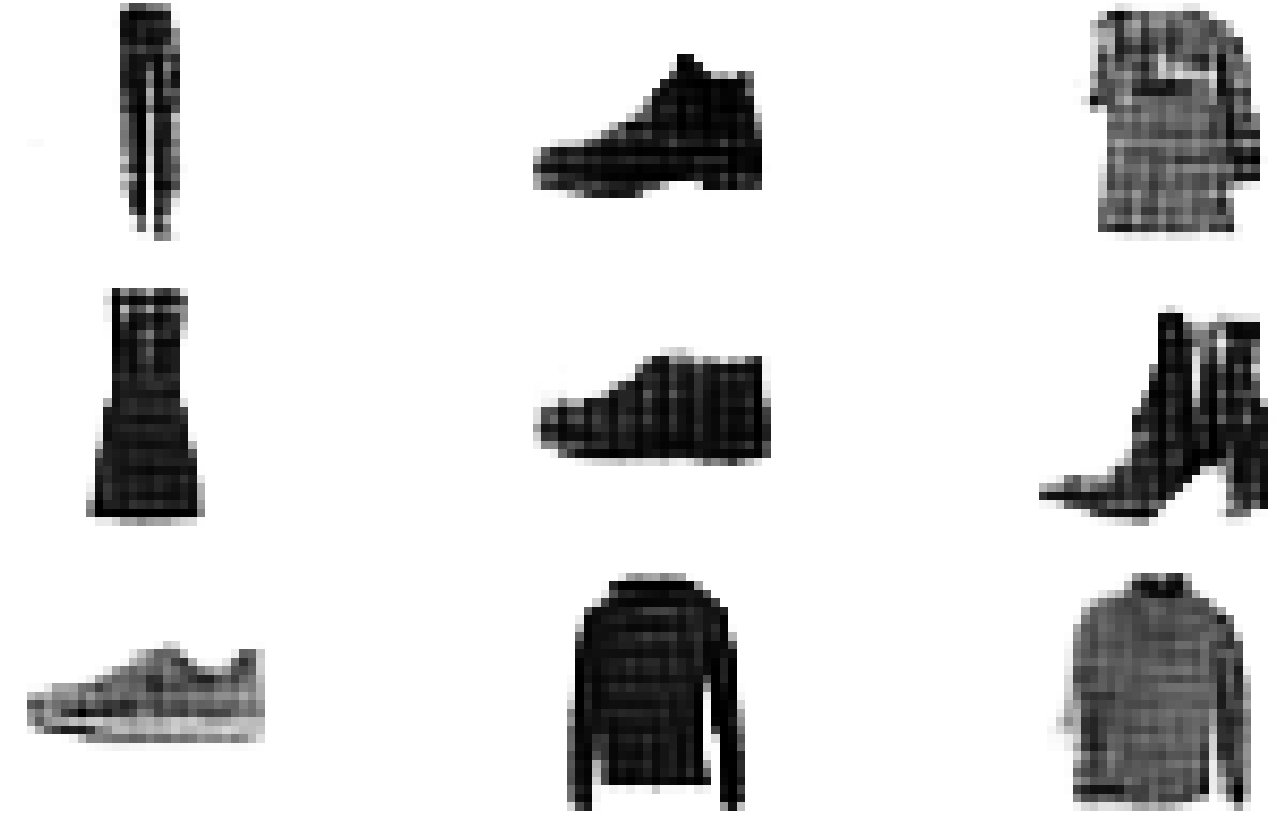}
}

\rule{0.8\textwidth}{0.3mm}\\
{
\includegraphics[width=0.18\textwidth]{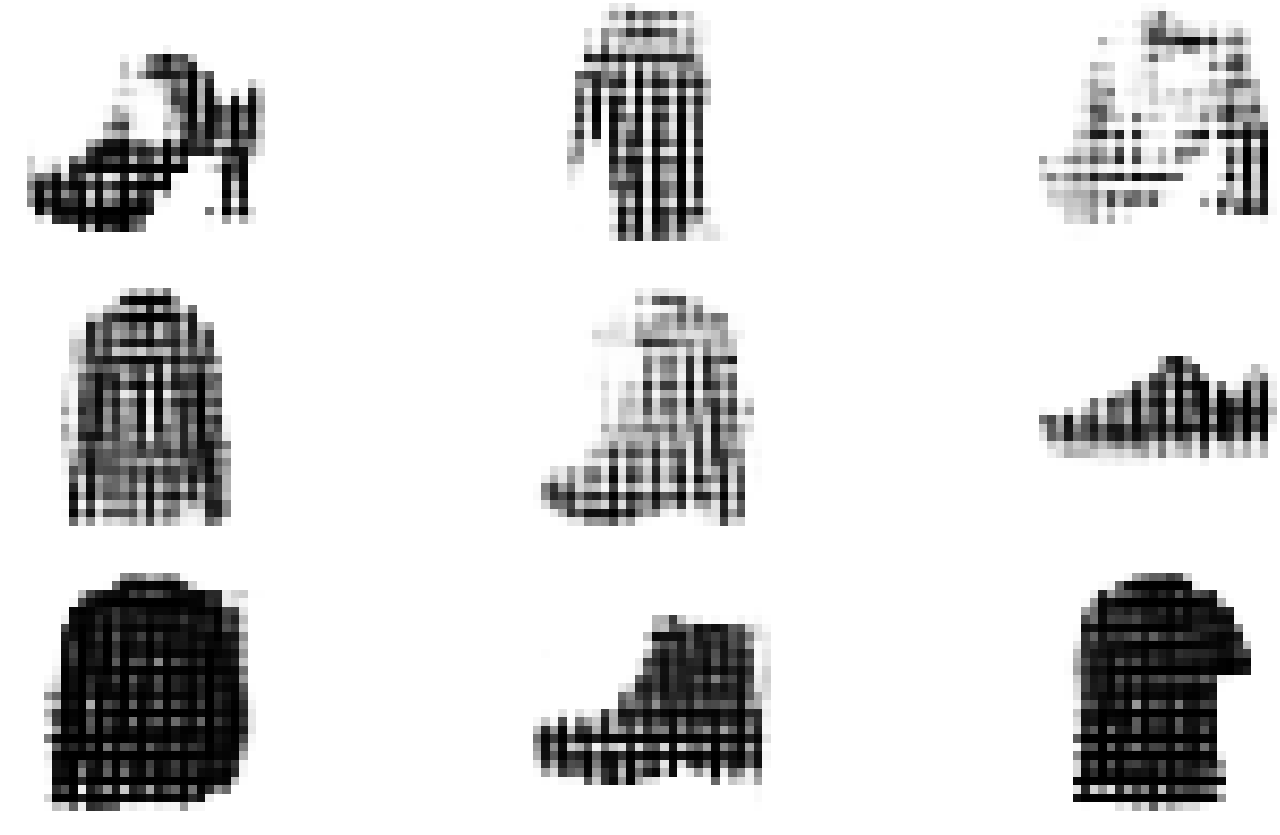}
}
{
\includegraphics[width=0.18\textwidth]{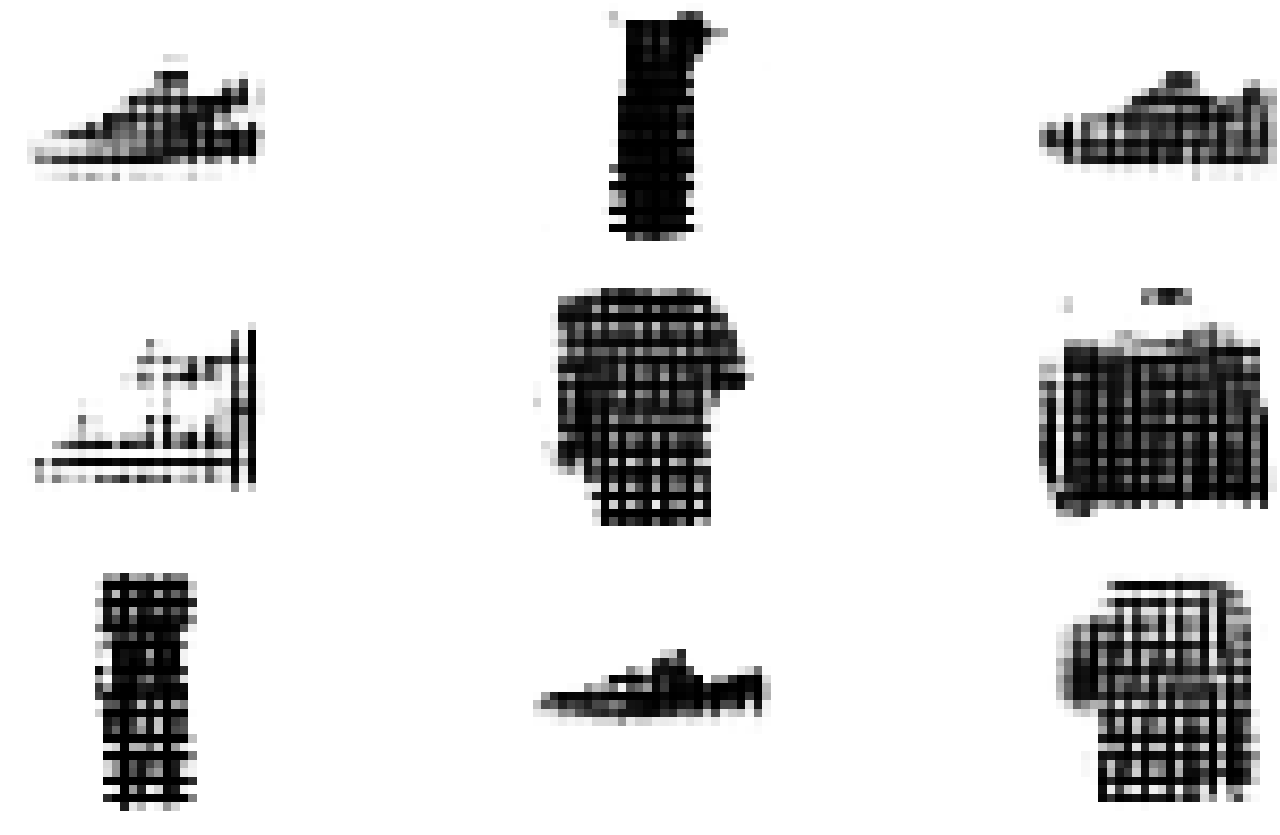}
}
{
\includegraphics[width=0.18\textwidth]{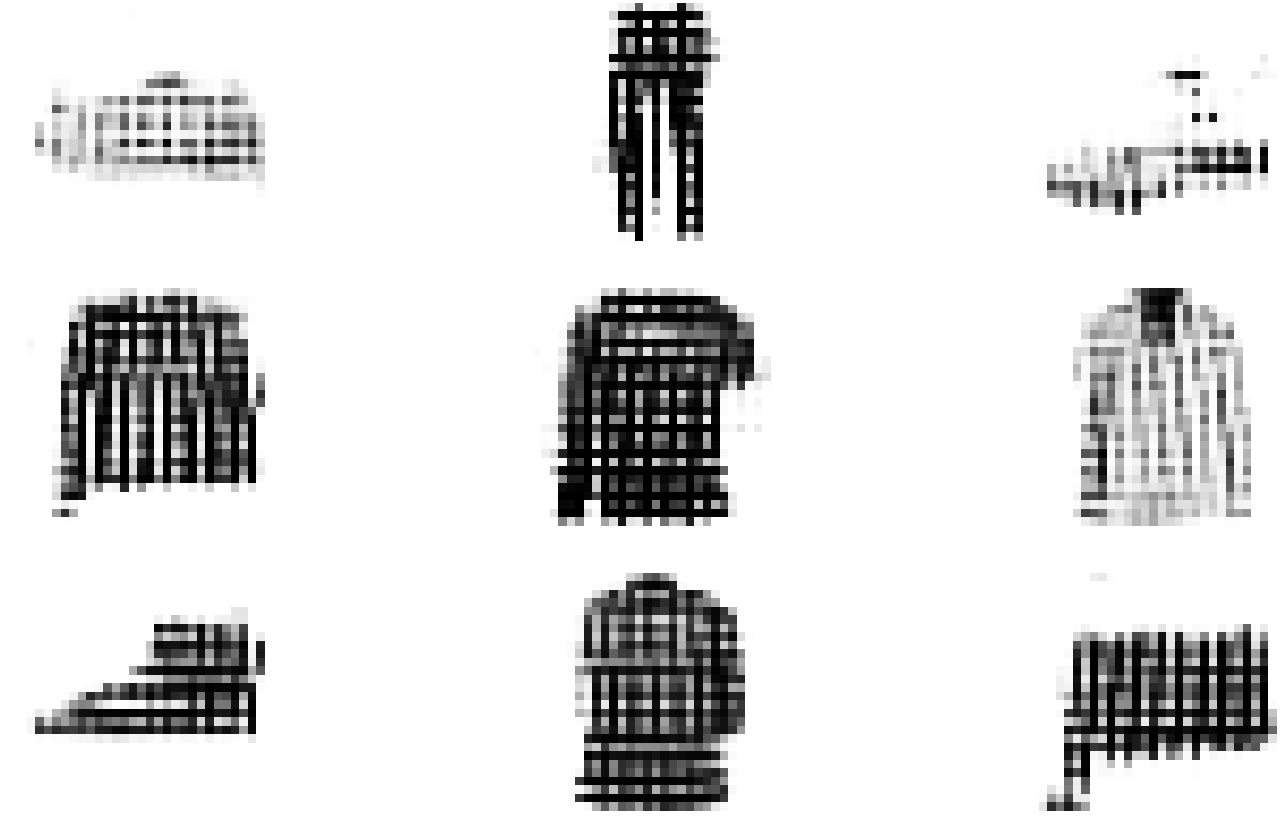}
}
{
\includegraphics[width=0.18\textwidth]{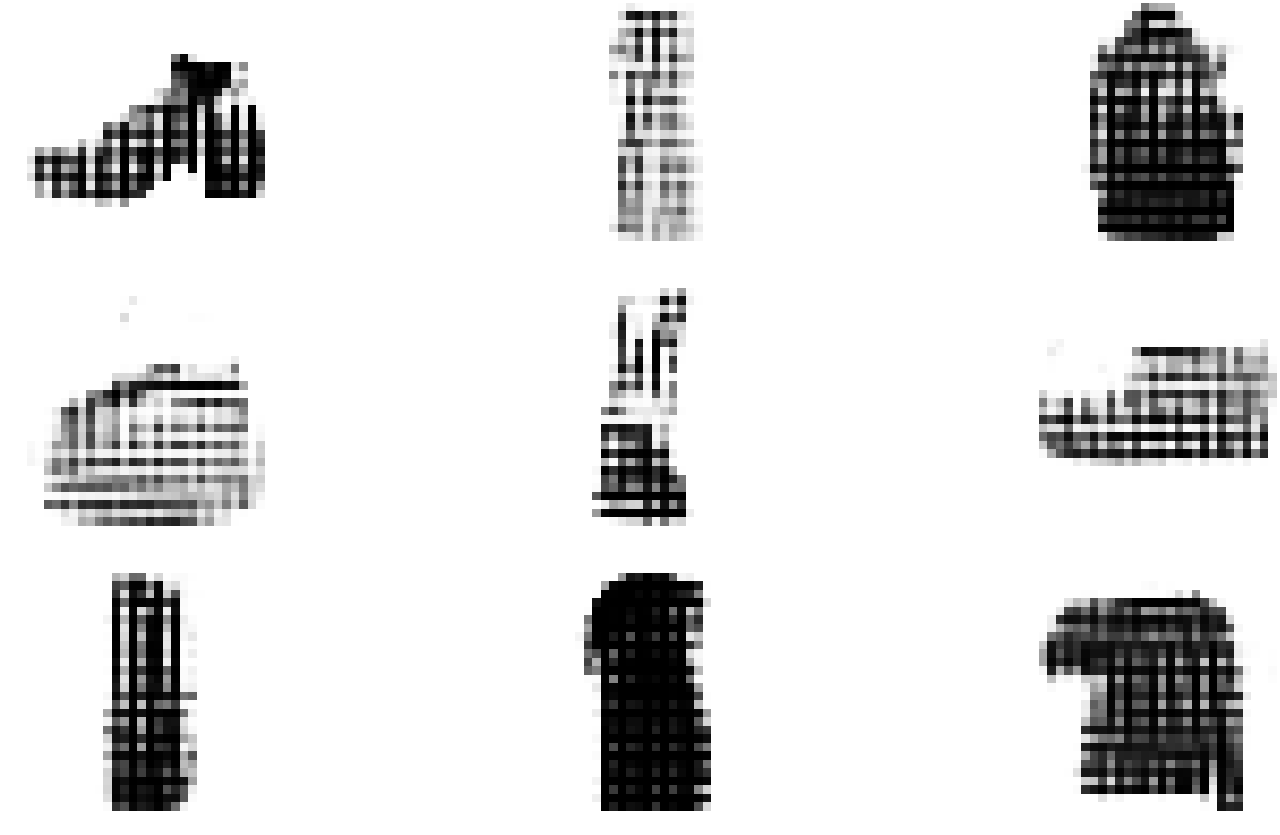}
}
\\
{
\includegraphics[width=0.18\textwidth]{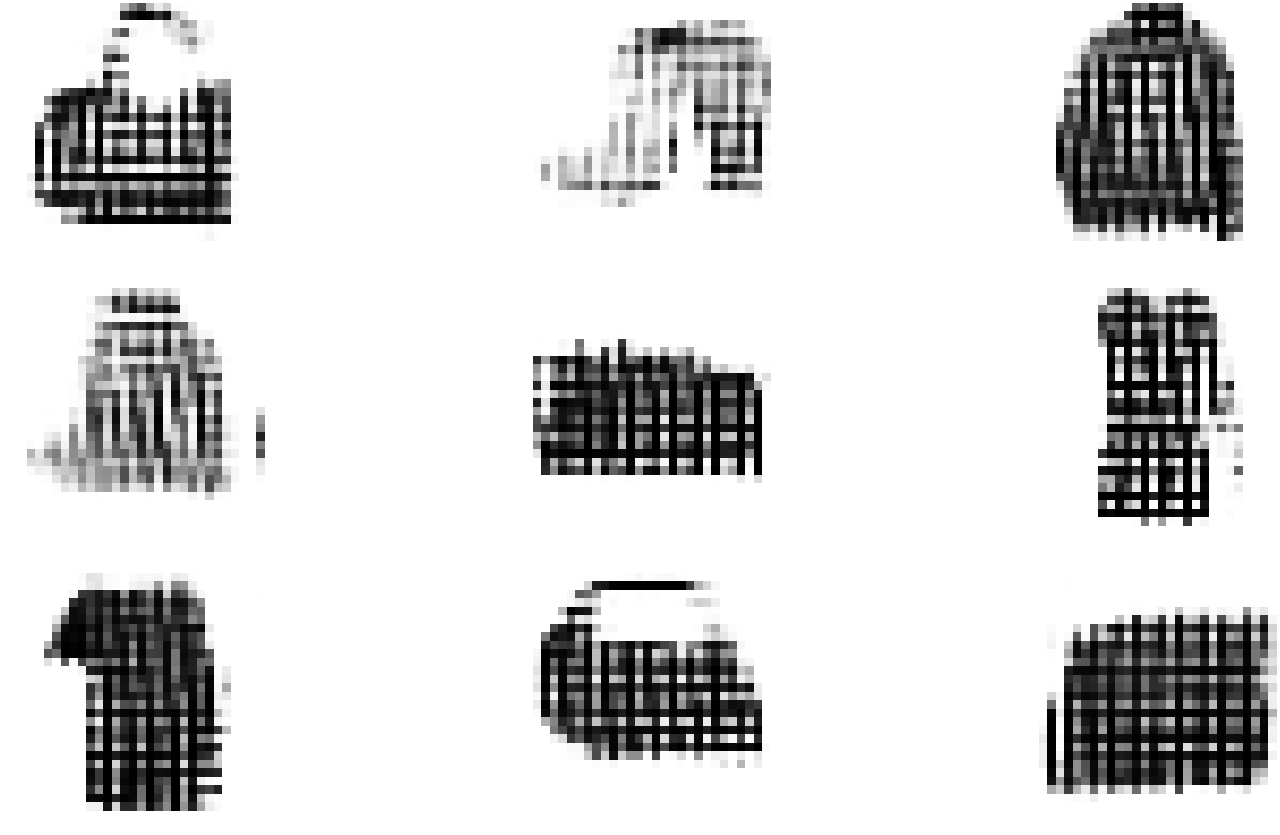}
}
{
\includegraphics[width=0.18\textwidth]{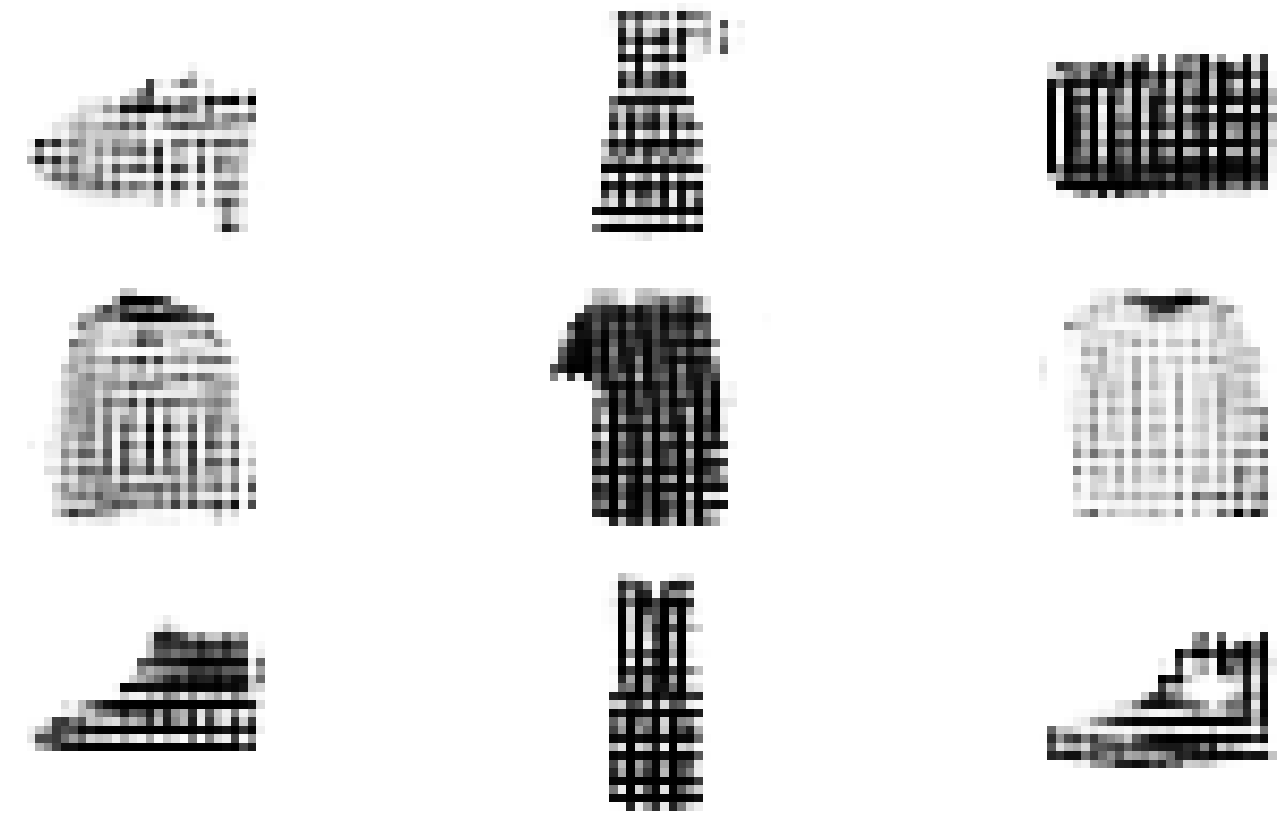}
}
{
\includegraphics[width=0.18\textwidth]{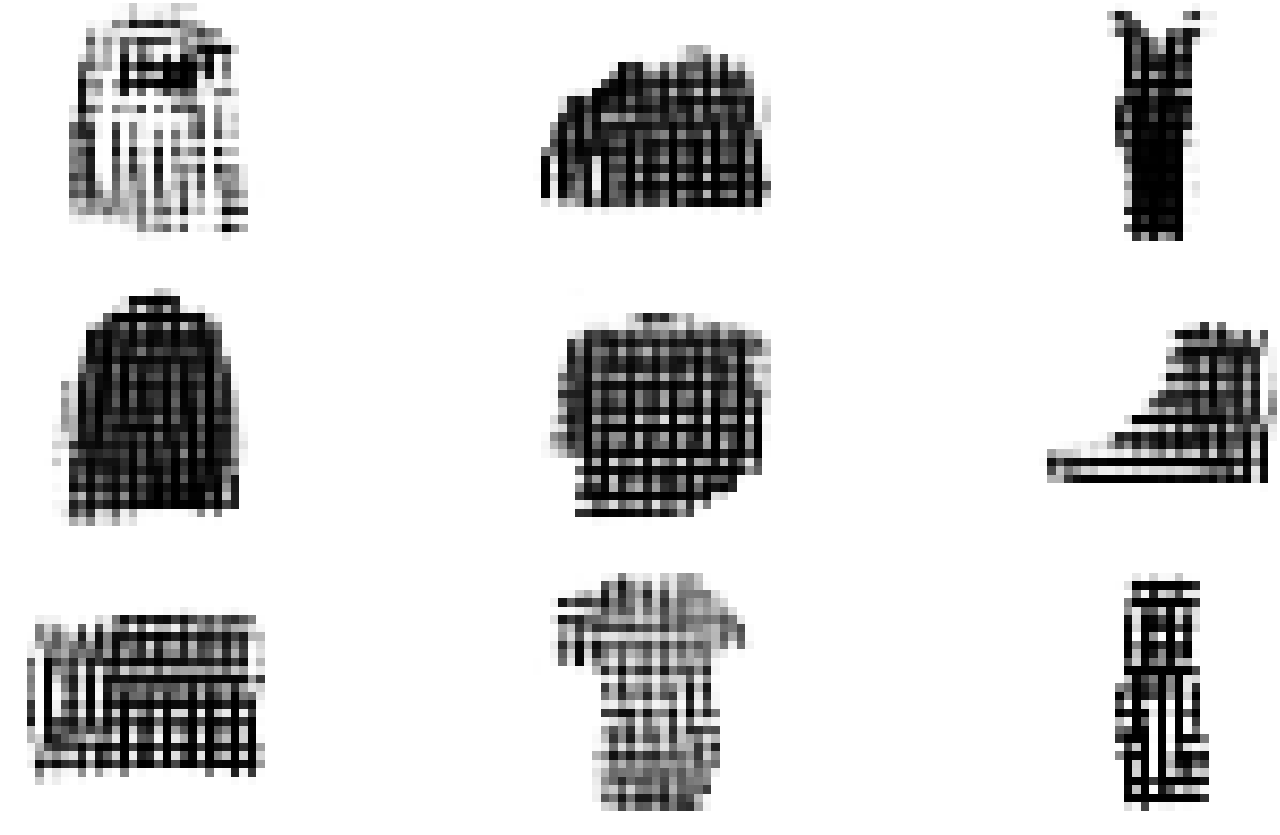}
}
{
\includegraphics[width=0.18\textwidth]{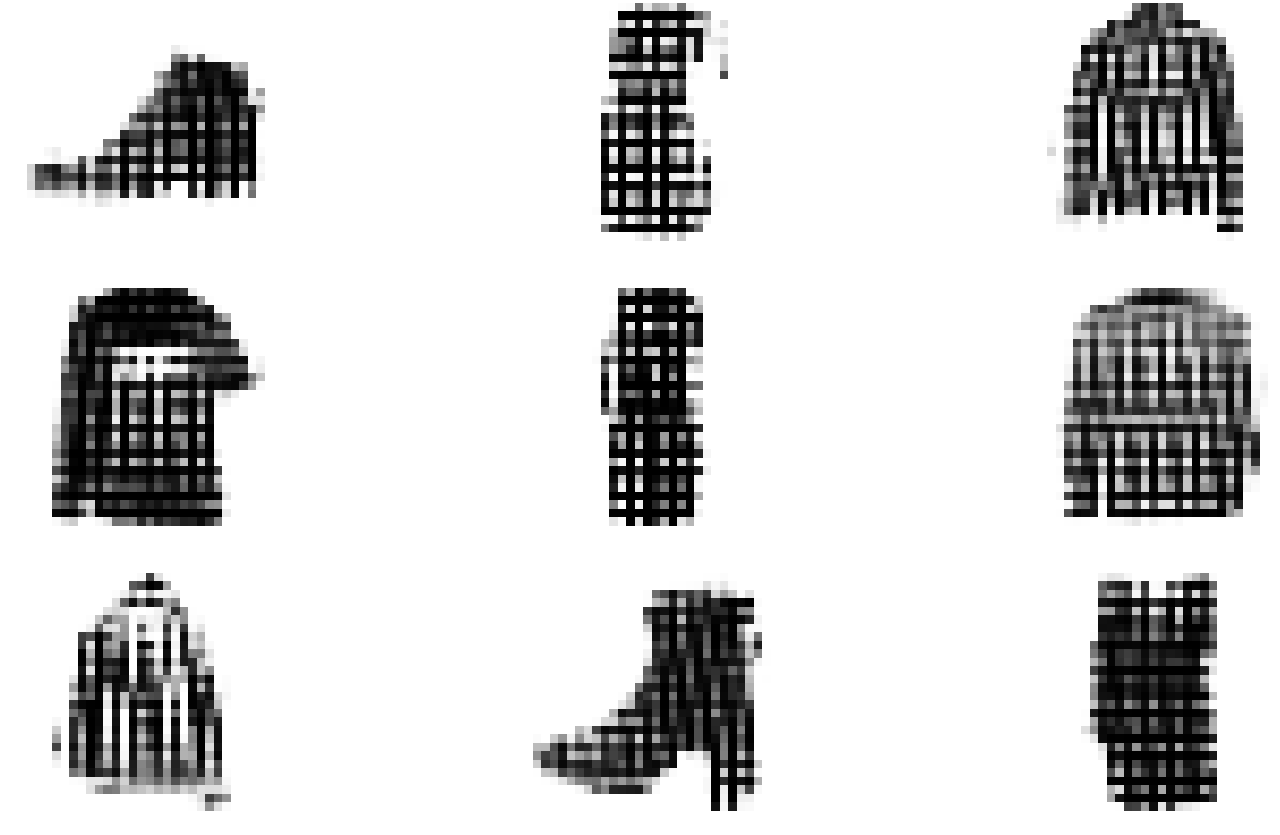}
}

\caption{Blurred and recovered images. The image in a box refers to the blurred image with convolution kernel $K_{blur}$. The rest images, divided into 3 parts by two lines, refer to the recovered images generated by the models with convolution kernels $K_{rec}^1\ (top),\ K_{rec}^2\ (medium),\ K_{rec}^3\ (bottom)$. The models are trained with 20000 images from the Fashion-MNIST dataset.}
\label{fashion_recovery}
\end{figure}

Furthermore, if the images are already blurred by the previous method with some convolution kernel $K_{blur}$, we can recover the blurred images with our model with convolution kernel $K_{rec}$. Essentially, $K_{blur}$ and $K_{rec}$ do not need to be exactly the same. Fig \ref{recovery} and Fig \ref{fashion_recovery} demonstrate the blurred and recovered images under several different pairs of $(K_{blur}, K_{rec})$. The results show that images indeed can be recovered even when $K_{rec}\neq K_{blur}$, which implies that our model can be used in practical situations when $K_{blur}$ is unknown but can be roughly estimated. The selected convolution kernels are
\begin{equation}
K_{blur},K_{rec}^1=\frac19\left(
\begin{array}{ccc}
 1 & 1 & 1 \\
 1 & 1 & 1 \\
 1 & 1 & 1 \\
\end{array}
\right), K_{rec}^2=\left(
\begin{array}{ccc}
 0.1 & 0.12 & 0.1 \\
 0.12 & 0.13 & 0.12 \\
 0.1 & 0.12 & 0.1 \\
\end{array}
\right), K_{rec}^3=\left(
\begin{array}{ccc}
 0.08 & 0.12 & 0.08 \\
 0.12 & 0.19 & 0.12 \\
 0.08 & 0.12 & 0.08 \\
\end{array}
\right).
\end{equation}

\subsection{Measurement of Sharpness}
In order to examine the effect of the sharpening and recovery, a measurement of sharpness $\chi_s$ is introduced in this section. $\chi_s$ is a function that maps an image (essentially a matrix with all elements $\in[-1,1]$) into the interval $[0,1]$. For an image $P$, the larger $\chi_s(P)$ is, the sharper the image is, according to the meaning of this function. Now we give the definition of $\chi_s$.

Let $P$ be a matrix that represents an image, where $P_{ij}\in[-1,1]\forall i,j$. Let $\Delta P$ be a matrix with exactly the same size as $P$, where $(\Delta P)_{ij}$ is the average value of the absolute differences of $P_{ij}$ and its neighbours. That is,
\begin{equation}
(\Delta P)_{ij}=\frac14
\left(
|P_{ij}-P_{i-1,j}|+|P_{ij}-P_{i+1,j}|
+|P_{ij}-P_{i,j-1}|+|P_{ij}-P_{i,j+1}|
\right)
\end{equation}
and similar for $P_{ij}$'s on the edge or corner of the image. In one sentence, $\Delta$ is the absolute average difference operator. Then, $\chi_s$ is defined as the average value of the second-order absolute average difference of $P$:
\begin{equation}
\chi_s(P)=\overline{\Delta(\Delta P)}.
\end{equation}

\begin{figure}
\centering
\includegraphics[width=0.45\textwidth]{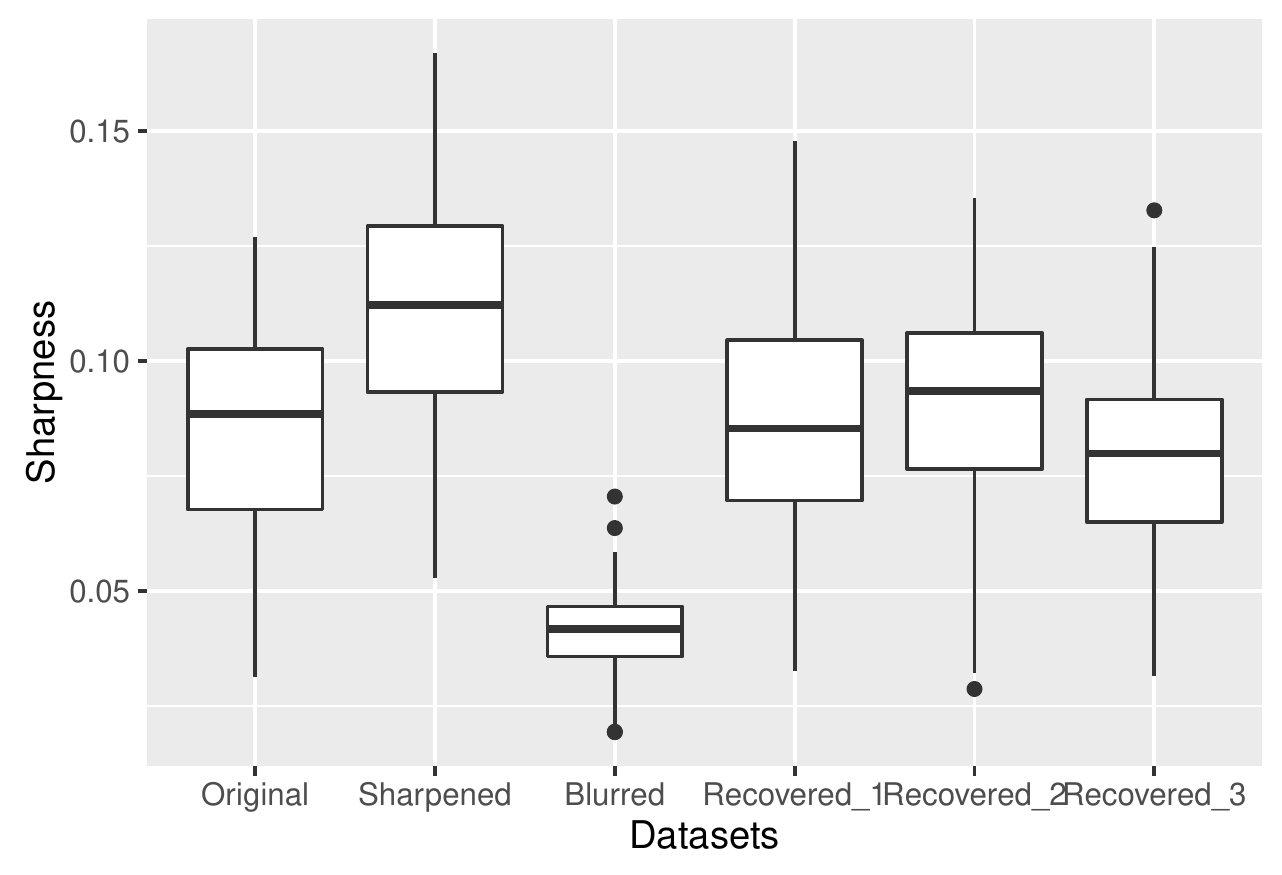}
\includegraphics[width=0.45\textwidth]{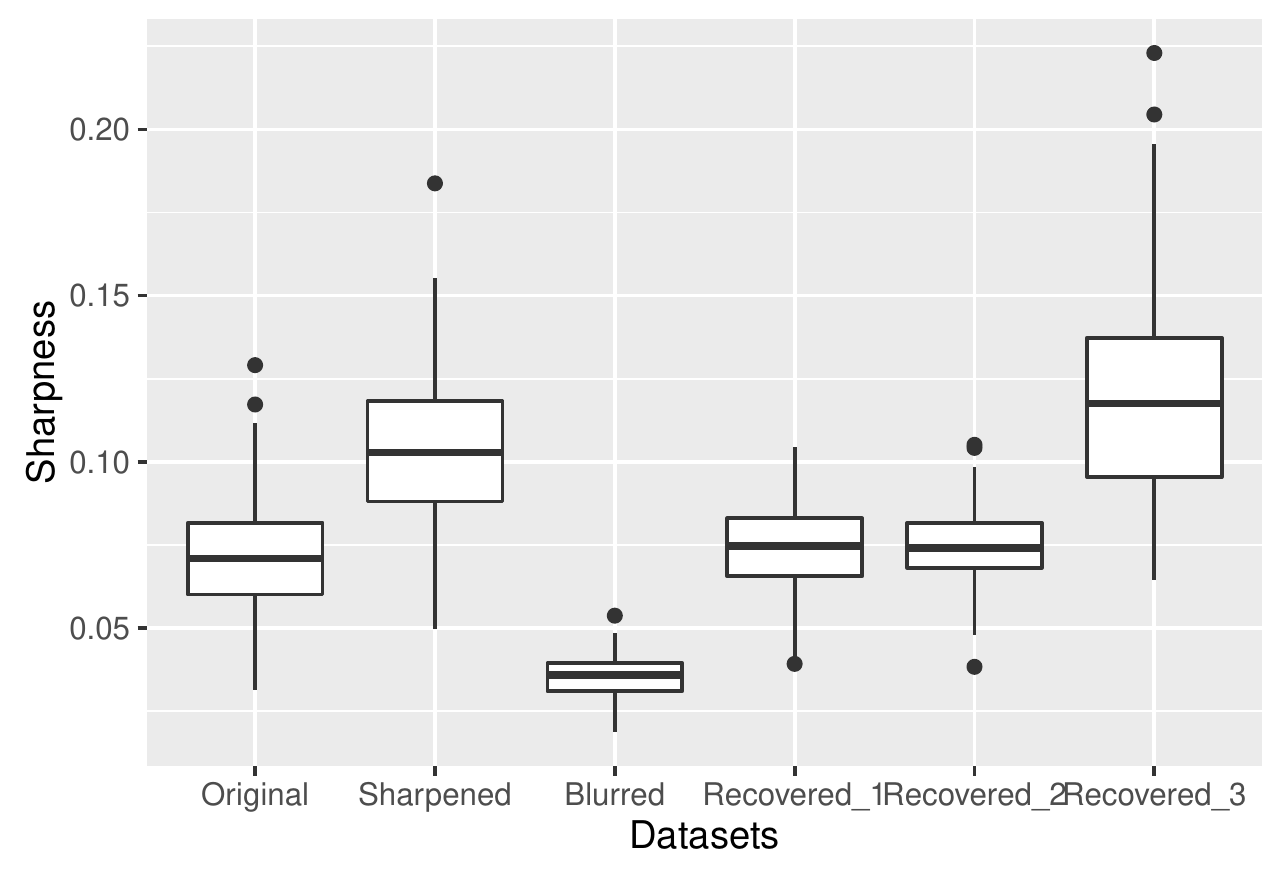}
\caption{Distribution of the sharpness of six groups of images discussed in this section with respect to the MNIST dataset (left) and the Fashion--MNIST dataset (right). The samples are randomly selected from the original datasets, sharpened images with convolution kernels $K_{sharpen}$, blurred images with convolution kernel $K_{blur}$, and recovered images with convolution kernels $K_{rec}^1, K_{rec}^2, K_{rec}^3$ (from left to right). The minimal values, first quartiles, second quartiles, third quartiles, and maximal values are illustrated in this figure.}
\label{boxplot}
\end{figure}

Using this measurement, we examined the sharpness of images from six different groups with respect to the two datasets. For each dataset, the six groups includes the original MNIST (or Fashion--MNIST) images, sharpened images, blurred images and recovered images with various convolution kernels. For each group, 108 samples were extracted randomly, and the distributions of the values of sharpness are demonstrated in Figure \ref{boxplot}. The results almost conform with our theory. For the MNIST dataset, the sharpened images have a higher value of $\chi_s$; although the blurred images have a much lower $\chi_s$, the recovered ones with all three $K_{rec}$'s tend to have almost the same $\chi_s$ as the original images. For the Fashion--MNIST dataset, despite that recovered images with convolution kernel $K_{rec}^3$, which is the farthest away from $K_{blur}$, have higher $\chi_s$, the other five groups of images still yield good results similar to those of the MNIST dataset.

\section{Conclusion}
In this paper, we presented a new architecture of Generative Adversarial Networks by adding an inverse transformation unit behind the generator. We made rigorous theoretical analysis to our model: we explicitly solved the optimal discriminator given the generator $G$, and proved the convergence of the algorithm under certain conditions. We also made several experiments. The first experiment was a general survey on models with various transformation functions. The survey illustrates that when the $T$ is not bijection, the model may still work. In the second experiment, we demonstrated that our architecture is able to generate sharpened images, and able to recover blurred images without using the same convolution kernel. In the third experiment, we defined a measurement of sharpness $\chi_s$ and compared this value with respect to different groups of images; the results also imply that our model works well for generating images with sharpening or recovering them at the same time. In the future, we plan to apply our model to a wider range of transformation functions in computer vision, such as various filters, and survey the inverse effects of them.

\newpage
\section*{Appendix}

\begin{table}[!h]
\centering
\caption{Nine $T$'s studied in the general survey on the MNIST dataset. The properties include injection, surjection, differentiability and continuity. The last column implies if a transformation function works. For convenience, if two functions share the same properties, they share the first index too.}
\begin{tabular}{|c|c|c|c|c||c|}
  \hline
  Indices & Function & $\begin{array}{l}\mbox{Injection}/\\ \mbox{Surjection}\end{array}$& Differentiability & Continuity & Effect\\
   \hline
  $T_1$ & $T(\vec{x})=\hat{I}\vec{x}$ & Yes / Yes & Yes & Yes & Yes\\
  $T_{21}$ & $T(x)=\sigma(x)$ & Yes / No\ & Yes & Yes & No\\
  $T_{22}$ & $T(x)=\arctan(x)$ & Yes / No\ & Yes & Yes & Yes\\
  $T_{23}$ & $T(x)=\tanh(x)$ & Yes / No\ & Yes & Yes & Yes\\
  $T_{31}$ & $T(x)=x+\frac14-\frac12\sigma(10x+9)$ & \ No / Yes & Yes & Yes & No\\
  $T_{32}$ & $T(x)=x+\sin(\pi x)/2$ & No\ / Yes & Yes & Yes & Yes\\
  $T_4$ & $T(x)=x^2+\sin(\pi x)$ & No\ / No\ & Yes & Yes & No\\
  $T_{51}$ & $T(x)=x\sin(1/x)$ & No\ / No\ & No when $x=0$& Not uniform & No \\
  $T_{52}$ & $T(x)=|x|$ & No\ / No\ & No when $x=0$ & Not uniform & No\\
  \hline
\end{tabular}

\label{all_T_table}
\end{table}

\begin{thebibliography}{}

\bibitem{GAN}
Ian J. Goodfellow, Jean Pouget-Abadie, Mehdi Mirza, Bing Xu, David Warde-Farley,
Sherjil Ozair, Aaron Courville, Yoshua Bengio: Generative Adversarial Nets. In: Advances in Neural Information Processing Systems, pp. 2672-2680. (2014)

\bibitem{MNIST}
LeCun, Y., Bottou, L., Bengio, Y., and Haffner, P.: Gradient-based learning applied to document recognition. In: Proceedings of the IEEE, vol. 86(11), pp. 2278-2324. (1998)

\bibitem{fashion_MNIST}
Han Xiao, Kashif Rasul, Roland Vollgraf. \textit{Fashion-MNIST: a Novel Image Dataset for Benchmarking Machine Learning Algorithms.}  arXiv preprint arXiv:1708.07747 (2017)

\bibitem{DCGAN}
Alec Radford, Luke Metz, Soumith Chintala: Unsupervised Representation Learning with Deep Convolutional Generative Adversarial Networks. arXiv preprint arXiv:1511.06434 (2015)

\bibitem{bGAN}
Masatosi Uehara, Issei Sato, Masahiro Suzuki, Kotaro Nakayama, Yutaka Matsuo: B-GAN: Unified Framework of Generative Adversarial Networks. In: \url{https://openreview.net/pdf?id=S1JG13oee} (2016)

\bibitem{fGAN}
Sebastian Nowozin, Botond Cseke, Ryota Tomioka: f-GAN: Training Generative Neural Samplers using Variational Divergence Minimization. arXiv preprint arXiv:1606.00709 (2016)

\bibitem{InfoGAN}
Xi Chen, Yan Duan, Rein Houthooft, John Schulman, Ilya Sutskever, Pieter Abbeel: InfoGAN: Interpretable Representation Learning by Information Maximizing Generative Adversarial Nets. In: Advances in Neural Information Processing Systems, pp. 2172-2180 (2016)

\bibitem{GRAN}
Daniel Jiwoong Im, Chris Dongjoo Kim, Hui Jiang, Roland Memisevic: Generating images with recurrent adversarial networks. arXiv preprint arXiv:1602.05110 (2016)

\bibitem{LAPGAN}
Emily Denton, Soumith Chintala, Arthur Szlam, Rob Fergus: Deep Generative Image Models using a Laplacian Pyramid of Adversarial Networks. In: Advances in Neural Information Processing Systems, pp. 1486-1494 (2015)


\bibitem{EnergyGAN}
Junbo Zhao, Michael Mathieu and Yann LeCun: Energy-Based Generative Adversarial Networks. arXiv preprint arXiv:1609.03126 (2016)

\bibitem{WGAN}
Martin Arjovsky, Soumith Chintala, and Leon Bottou: Wasserstein GAN. arXiv preprint arXiv:1701.07875 (2017)

\bibitem{LSGAN}
Guo-Jun Qi: Loss-Sensitive Generative Adversarial Networks
on Lipschitz Densities. arXiv preprint arXiv:1701.06264 (2017)


\bibitem{seqGAN}
Lantao Yu, Weinan Zhang, Jun Wang, Yong Yu: SeqGAN: Sequence Generative Adversarial Nets with Policy Gradient. In: Thirty-First AAAI Conference on Artificial Intelligence (2017)

\bibitem{cross-domain}
Yaniv Taigman, Adam Polyak, Lior Wolf: Unsupervised Cross-Domain Image Generation. arXiv preprint arXiv:1611.02200 (2016)


\bibitem{transfer_CNN}
Leon A. Gatys, Alexander S. Ecker, Matthias Bethge: Image Style Transfer Using Convolutional Neural Networks. In: Proceedings of the IEEE Conference on Computer Vision and Pattern Recognition, pp. 2414-2423. (2016)

\bibitem{letter_CNN}
Shumeet Baluja: Learning Typographic Style. arXiv preprint arXiv:1603.04000 (2016)

\bibitem{letter_GAN}
TJ TORRES: \\
\url{http://multithreaded.stitchfix.com/blog/2016/02/02/a-fontastic-voyage/} (2016)

\bibitem{letter_VAE}
Paul Upchurch, Noah Snavely, Kavita Bala: From A to Z: Supervised Transfer of Style and Content Using Deep Neural Network Generators. arXiv preprint arXiv:1603.02003 (2016)

\bibitem{realNVP}
Laurent Dinh, Jascha Sohl-Dickstein, Samy Bengio: Density Estimation using Real NVP. arXiv preprint arXiv: 1605.08803 (2016)

\bibitem{cycleGAN}
Jun-Yan Zhu, Taesung Park, Phillip Isola, Alexei A. Efros: Unpaired Image-to-Image Translation using Cycle-Consistent Adversarial Networks. arXiv preprint arXiv:1703.10593 (2017)


\end{thebibliography}
\end{document}